\documentclass{article}

 \usepackage[main, preprint]{neurips_2026}

\usepackage{threeparttable}
\usepackage{amsmath}
\usepackage{amssymb}
\usepackage{mathtools}
\usepackage{amsthm}

\usepackage{tocloft}
\usepackage{etoc}
\usepackage{hyperref}
\usepackage[utf8]{inputenc} 
\usepackage[T1]{fontenc}    
\usepackage{hyperref}       
\usepackage{url}            
\usepackage{booktabs}       
\usepackage{amsfonts}       
\usepackage{nicefrac}       
\usepackage{microtype}      
\usepackage{xcolor}         

\usepackage{graphicx}
\usepackage{subcaption}
\usepackage{booktabs} 
\usepackage{multirow}

\usepackage{algpseudocode}
\usepackage{algorithm}
\usepackage[textsize=tiny]{todonotes}

\hypersetup{hidelinks}

\title{Newton-PINet: A fast physics-informed neural network with Newton linearization for meta-learning nonlinear PDEs}

\author{%
Yuchen Fan\textsuperscript{1,\(\ast\)} \quad
Chang Wei\textsuperscript{1} \quad
Pao-Hsiung Chiu\textsuperscript{2} \quad
Chin Chun Ooi\textsuperscript{2} \\
[0pt]
\textbf{Heyang Wang}\textsuperscript{\textbf{1}} \quad
\textbf{Jian Cheng Wong}\textsuperscript{\textbf{2,}}\thanks{Corresponding authors: fanyuchen@tju.edu.cn (Yuchen Fan); wongj@a-star.edu.sg (Jian Cheng Wong)}
\\[2pt]
\textsuperscript{1}School of Mechanical Engineering, Tianjin University, China \\
\textsuperscript{2}Institute of Advanced Intelligence and Computing, A*STAR, Singapore}

\begin{document}

\maketitle

\begin{abstract}
  Scientific machine learning has opened new avenues for solving parameterized partial differential equations (PDEs), enabling models to learn a family of PDEs and generalize to unseen instances. In this context, data-driven operator learning methods typically require large training datasets, while physics-informed neural networks (PINNs) suffer from difficult optimization and limited generalization, especially for nonlinear PDEs. We propose Newton-PINet, a physics-informed network enhanced by Newton linearization, offering an effective meta-learning framework for nonlinear PDEs. Newton-PINet (i) employs a physics-informed multilayer network with skip connections, where the output-layer weights are solved by least squares; (ii) adopts a two-stage learning strategy that first leverages gradient-based training to learn robust representations from the available training tasks, and then performs gradient-free fine-tuning on the output layer for fast task-specific generalization; and (iii) incorporates a Newton linearization method to speed up the least-squares iteration for nonlinear PDE problems. On a challenging nonlinear reaction-diffusion benchmark, Newton-PINet achieves up to three orders of magnitude lower relative error than recent neural solvers, while using 16× fewer training tasks and over an order of magnitude less training time (about 5 minutes versus several hours). This work advances the meta-learning of PINNs toward data-efficient, fast, and generalizable physics solvers. The code and datasets are publicly available at: \href{https://github.com/fycAIwj/Newton-PINet}{\texttt{https://github.com/fycAIwj/Newton-PINet}}.
\end{abstract}

\section{Introduction and related works}
Solving partial differential equations (PDEs) is fundamental across diverse fields, including fluid dynamics, climate modeling, and biophysics ~\citep{karniadakis2021physics,calm-pde-hagnberger:2025}. Traditional numerical solvers for PDEs often incur prohibitive computational costs, particularly when repeated evaluations are required in applications. Recently, scientific machine learning has emerged as a powerful paradigm, leveraging advances in deep neural networks to deliver fast and accurate approximations of PDE solutions, thereby enabling high-fidelity simulations and real-time predictive capabilities that were previously unattainable~\citep{cai2021physicsa,cuomo2022scientific,hu2026realpdebench}. Deep learning for PDEs can be broadly categorized into: data-driven operator learning, physics-informed deep learning, and meta-learning physics-informed neural networks (PINNs).

\textbf{Data-driven operator learning:} Data-driven operator learning focuses on training neural networks to directly map input functions, such as PDE parameters or initial conditions, to their corresponding PDE solutions. Well-known examples of this approach include DeepONet~\citep{lu2021learning}, Derivative-enhanced-DeepONet~\citep{qiu2024derivative}, Transformer-DeepONet~\citep{wei2025efficient}, Fourier neural operators (FNO)~\citep{li2020fourier}, Factorized-FNO~\citep{tran2023factorized},  Decomposed-FNO~\citep{li2025d}, and Tucker-FNO~\citep{zhou2026tucker}.
Once trained, these models can generalize to new parameter settings (e.g., new initial conditions), enabling fast prediction of complex dynamical behaviors. However, they lack explicit physics constraints and interpretability, and their generalization depends on adequacy of labeled training data, which are often expensive to obtain~\citep{li2024physics}. For example, Transformer-DeepONet typically requires at least 1,000 different PDE solutions for adequate performance~\citep{wei2025efficient}.

\textbf{Physics-informed deep learning:} An alternative paradigm, physics-informed deep learning, incorporates governing PDEs and boundary conditions into the training loss function~\citep{wong2022learning,xu2025fp64,wei2025ffv}. By penalizing violation of physical laws, these models can be trained effectively even in data-sparse regimes, often requiring only initial or boundary conditions. Representative approaches include physics-informed variants of operator learning, such as physics-informed DeepONet ~\citep{wang2021learning}, physics-informed neural operators (PINO) ~\citep{li2024physics}, and HyPINO~\citep{bischof2025hypino}. Another prominent method is PINN~ \citep{raissi2019physics}, which learns to map the spatiotemporal coordinates to the PDE solution for a given system. While these methods alleviate the need for large labeled datasets, their reliance on physics-informed objectives often results in prohibitively long training time, due to highly nonconvex loss landscapes caused by the stiff PDE constraints~\citep{chiu2022can,wang2023expert,datar2024fast}, especially for nonlinear PDEs. For example, training physics-informed DeepONet on a family of Burgers' equations can take more than 9 hours~\citep{wang2021learning}, while PINNs may require over 24 hours to solve a single Kuramoto-Sivashinsky equation~\citep{wang2025gradient}.

\textbf{Meta-learning PINNs:} Adapting PINN to new PDEs or parameter settings typically requires costly retraining. Transfer learning strategies accelerate convergence by reusing knowledge from related tasks~\citep{wong2021can,wang2022respecting}. Parameterized PINNs (P$^2$INNs) further incorporate problem parameters as network inputs, enabling a single model to represent families of PDEs without labeled data while preserving physical consistency~\citep{cho2024parameterized}. However, P$^2$INNs still suffer from the optimization challenges of physics-based loss functions, and their adaptation to new tasks remains heavily reliant on gradient-based fine-tuning which is often slow to converge, especially for nonlinear PDEs. Meta-learning methods aim to enable rapid adaptation to unseen PDEs with limited data by learning parameter initializations or update rules~\citep{penwarden2023metalearning,wong2025evolutionary}. MAML-style approaches learn transferable parameter priors~\citep{liu2022novel, wong2026transferable}, while other methods leverage latent codes or meta-optimizers for task-specific adaptation~\citep{huang2022meta,cho2023hypernetwork}. To avoid gradient-based adaptation, Baldwinian-PINN introduces a gradient-free meta-learning strategy based on neuroevolution, evolving shared representations and fine-tuning the output layer via least squares~\citep{wong2026evolutionary}. While highly efficient for linear PDEs, its single-layer architecture limits nonlinear expressiveness. Moreover, its Picard (lagging-of-coefficients) iteration exhibits only linear convergence, leading to slow solution refinement for nonlinear problems~\citep{chiu2008effective,pletcher2012computational}.  

\begin{figure*}[!hbt]
    \begin{center}    
    \centerline{\includegraphics[width=1\linewidth]{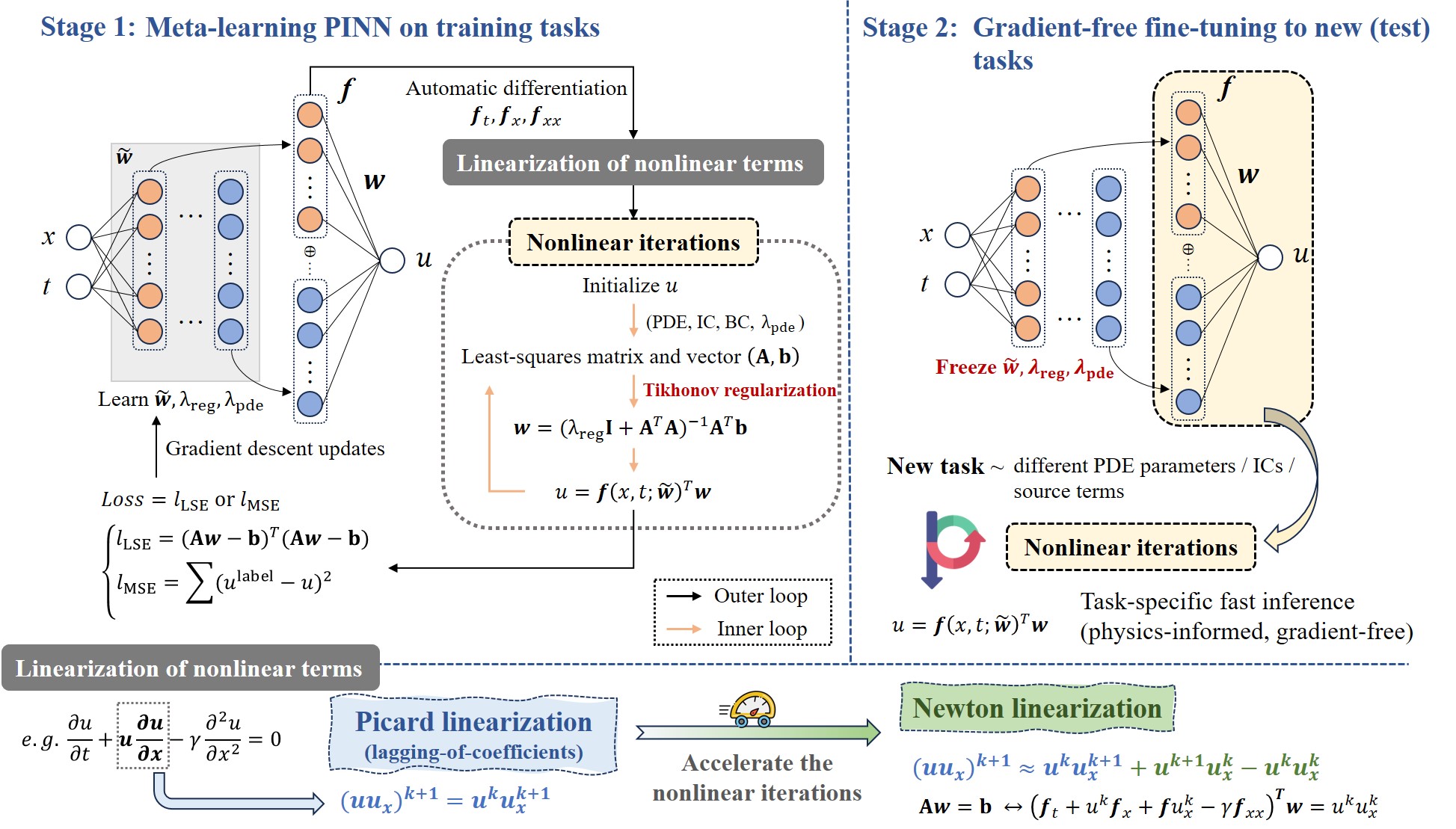}}
    \caption{\label{fig:Newton-PINet model framework}Newton-PINet model framework. Stage~1 (pre-training) meta-learns the task-agnostic parameters $\Theta=\{\boldsymbol{\tilde{w}},\lambda_\text{pde},\lambda_\text{reg}\}$ by gradient descent (outer loop), while the output-layer weights $\boldsymbol{w}$ of each training task are updated by gradient-free Tikhonov-regularized least squares (inner loop); Stage~2 (inference) freezes $\Theta$ and adapts only $\boldsymbol{w}$ to new tasks. The bottom panel compares Picard and Newton linearization; the latter accelerates the convergence of nonlinear iterations for solving nonlinear PDEs.}
    \end{center}
    
\end{figure*}

See Appendix~\ref{app:Extended review of related work} for an extended literature review. In summary, existing deep learning approaches for solving nonlinear PDEs remain constrained by substantial data requirements and/or slow, optimization-heavy adaptation. For this reason, we propose a novel meta-learning PINN framework (Figure~\ref{fig:Newton-PINet model framework}), termed \textbf{Newton-PINet}. The main contributions are as follows:

\begin{itemize}

  \item We introduce a physics-informed multilayer network with skip connections from early hidden layers to the output, where the final-layer weights are computed using least-squares approach (Tikhonov regularization). This skip-connected Tikhonov regularization PINN architecture improves the model's nonlinear representation capabilities.

  \item We employ a two-stage learning strategy, where the first stage uses a gradient-based method to meta-learn the nonlinear hidden layers' network weights and essential hyperparameters for task-specific adaptation in an unsupervised or few-shot manner. At inference (test time), task-specific adaptation is confined to the output layer, which can be rapidly updated in a gradient-free manner using Tikhonov regularization. 

  \item We integrate Newton linearization into the Tikhonov-regularized PINN and establish its equivalence to the classical regularized Gauss-Newton method through mathematical derivation, with experiments validating its near-quadratic convergence. The Newton linearization accelerates nonlinear least-squares convergence, providing a more efficient alternative to the traditional Picard approach during both meta-learning and inference. \textbf{We therefore refer to the proposed physics-informed network with Newton linearization as Newton-PINet.}
  
\end{itemize}

Results on diverse nonlinear PDE benchmarks show that, compared with recent neural solvers, Newton-PINet significantly reduces the reliance on training data while achieving fast and accurate generalization to new PDE tasks through physics-consistent fine-tuning.

\section{Preliminaries}  
For simplicity, consider a nonlinear PDE with spatial variable $x$, time $t$, and solution $u$ defined on $\Omega \times [0,T]$:  
\begin{subequations}
    \label{eq:general_pde}
    \begin{alignat}{3}
    \text{PDE:} \quad & \mathcal{N_{\theta}}[u(x,t) ] = q(x,t) 
    && \quad x \in \Omega, \; t \in [0,T], \label{eq:pde_1a}\\
    \text{IC:} \quad & u(x,t=0) = u_0(x) 
    && \quad x \in \Omega, \label{eq:pde_1b}\\
    \text{BC:} \quad & \mathcal{B}[u(x,t)] = g(x,t) 
    && \quad x \in \partial \Omega, \; t \in [0,T]. \label{eq:pde_1c}
    \end{alignat} 
    \label{eq:pde_all}
\end{subequations}
where the general differential operator $\mathcal{N}_{\theta}$ can include both linear and nonlinear combinations of the temporal and spatial derivatives and PDE parameters $\theta$, $q(x,t)$ is the source term, $u_0(x)$ is the initial condition (IC), and $g(x,t)$ is the boundary condition (BC).

\textbf{Single PDE problem:}  
Standard PINNs aim to solve a single PDE instance (\textbf{a single task}) defined by specific PDE parameters, IC, and BC. PINNs are trained by minimizing the discrepancy between Eq.~(\ref{eq:pde_all}) and the model’s prediction.

\textbf{Towards generalizable PINNs:}  
There is growing interest in models capable of generalizing across a set of tasks belonging to some underlying task-distribution $p(\mathcal{T})$, e.g., a family of PDEs spanning different PDE parameters, ICs, and BCs. For meta-learning PINNs, the goal is to learn network initializations using training tasks sampled from $p(\mathcal{T})$ that enable fast, accurate, and physics-aware predictions on unseen scenarios, i.e., any new task from the distribution $\mathcal{T}_i \sim p(\mathcal{T})$ through fine-tuning~\citep{wong2025evolutionary}.

\section{Methodology}
\label{Methodology}

\paragraph{Skip-connected Tikhonov regularization PINN:} We introduce skip connections from all hidden layers to the output layer, with the output-layer weights computed using Tikhonov regularization~\citep{golub1999tikhonov}. The skip-connected neural architecture improves expressivity, ensuring stable and accurate least-squares computation by increasing the output-layer width through stacking additional hidden layers, while still maintaining a moderate number of nodes per layer. As shown in Figure~\ref{fig:Newton-PINet model framework}, we employ the neural networks with $L+1$ layers, where the input is $\mathbf{x} = (x,t)$ (layer $0$) and the output is $u$ (layer $L$). Each hidden layer $l=1,\dots, L-1$ contains the same number of neurons ($N_n$). Sinusoidal feature embeddings are applied at the first layer~\citep{wong2022learning}, and all hidden layers use $\sin(\cdot)$ activation functions. 
The pre-final output $\boldsymbol{f}$ is constructed as the concatenation of all hidden activations (skip connections).  All trainable network parameters up to the pre-final layer are denoted $\boldsymbol{\tilde{w}}$. The output $u$ can be computed by $u(\mathbf{x}) = \boldsymbol{f}(\mathbf{x};\boldsymbol{\tilde{w}})^T \, \boldsymbol{w}$, where $\boldsymbol{w}$ is the output-layer weights. See Appendix~\ref{app:Skip-connected PINN architecture} for a detailed description of the skip-connected model architecture.

The objective of task-specific learning is to determine the best set of $\boldsymbol{w}$ such that $u(\mathbf{x})=\boldsymbol{f}(\mathbf{x};\boldsymbol{\tilde{w}})^T\boldsymbol{w}$ satisfies the PDE, IC, and BC defined in Eq.~(\ref{eq:pde_all}) for a task $\mathcal{T}_i \sim p(\mathcal{T})$. This can be formulated as a physics-based least-squares problem. Given collocation points for PDE residual, points at the initial time, and points on the boundary, together with the loss importance hyperparameter $\lambda_{\mathrm{pde}}$, the following system is obtained:

\begin{equation}
    \begin{bmatrix}
    \cdots &  \lambda_{\mathrm{pde}} \cdot \mathcal{N}_{\theta}[f_j(x^{\mathrm{pde}}_1, t^{\mathrm{pde}}_1;\boldsymbol{\tilde{w}})] & \cdots \\[-2pt]
     & \vdots  \\[-2pt]
    \cdots & f^i_j(x^{\mathrm{ic}}_1, 0;\boldsymbol{\tilde{w}}) & \cdots \\[-2pt]
    & \vdots  \\[-2pt]
    \cdots & \mathcal{B}[f_j(x^{\mathrm{bc}}_1, t^{\mathrm{bc}}_1;\boldsymbol{\tilde{w}})] & \cdots \\[-2pt]
    & \vdots  \\[-2pt]
    \end{bmatrix}
    \begin{bmatrix}
    \vdots \\[-2pt]
    w_j \\[-2pt]
    \vdots
    \end{bmatrix}
    =
    \begin{bmatrix}
    \lambda_{\mathrm{pde}} \cdot q(x^{\mathrm{pde}}_1, t^{\mathrm{pde}}_1) \\[-2pt]
    \vdots \\[-2pt]
    u_0(x^{\mathrm{ic}}_1) \\[-2pt]
    \vdots \\[-2pt]
    g(x^{\mathrm{bc}}_1, t^{\mathrm{bc}}_1) \\[-2pt]
    \vdots \\[-2pt]
    \end{bmatrix},
\label{eq:maintext-L-S matrix}
\end{equation}

which can be compactly written as
\[\mathbf{A}\boldsymbol{w} = \mathbf{b}.\]

Here $\mathbf{A} \boldsymbol{w}$ represents the left-hand side of Eq.~(\ref{eq:pde_all}) at a given set of collocation points, and $\mathbf{b}$ denotes the corresponding right-hand side of Eq.~(\ref{eq:pde_all}). The number of rows in $\mathbf{A}$ and $\mathbf{b}$ equals the total number of collocation points. The number of columns in $\mathbf{A}$ corresponds to the product of the network's pre-final feature dimension and the output dimension. The final output weights $\boldsymbol{w}$ are obtained via Tikhonov regularization: $ \arg\min_{\boldsymbol{w}} \; \left\lVert \mathbf{A}\boldsymbol{w} - \mathbf{b} \right\rVert_2^2+ \lambda_{\text{reg}} \, \boldsymbol{w}^T \boldsymbol{w}$. $\lambda_{\text{reg}}$ is the Tikhonov regularization parameter whose proper setting improves the conditioning of the problem. The system has a closed-form solution, either $\boldsymbol{w}=(\lambda_{\text{reg}} \mathbf{I} + \mathbf{A}^T\mathbf{A})^{-1} \mathbf{A}^T \mathbf{b}$ for over-determined system or $\mathbf{A}^T (\lambda_{\text{reg}} \mathbf{I} + \mathbf{A}\mathbf{A}^T)^{-1} \mathbf{b}$ for under-determined one, enabling fast, gradient-free updates. See Appendix~\ref{app:Physics-based least-squares formulation} for the details.

Note that the Tikhonov-regularized formulation provides a closed-form result in
a single computation step for any linear PDE. For nonlinear PDEs, the nonlinear terms must be linearized so that the system can be cast into a linear form. An initial guess is then required, followed by a few \textbf{nonlinear iterations} (a process that can also be interpreted as gradient-free fine-tuning) to update $\boldsymbol{w}$ toward the optimal solution. Hereafter, we use $\text{N}$ to denote the number of nonlinear iterations.
 
The proposed Newton-PINet model framework  consists of two stages (Figure~\ref{fig:Newton-PINet model framework}): (i) meta-learning on training tasks and (ii) gradient-free fine-tuning on new tasks.

\paragraph{Meta-learning on training tasks:} The meta-learning objective is to optimize the network weights $\boldsymbol{\tilde{w}}$, importance hyperparameter $\lambda_{\text{pde}}$ of PDE loss relative to IC/BC, and the regularization parameter $\lambda_{\text{reg}}$, collectively denoted $\Theta$, to learn task-agnostic representations spanning a family of PDEs and enable fast generalization to unseen tasks via fine-tuning only the output-layer weights using the Tikhonov regularization update. In our framework, the outer loop of meta-learning updates the learnable parameters via gradient descent with learning rate $\eta$, while task-specific adaptation in the inner loop, or generalization, is performed through gradient-free Tikhonov regularization applied to the output-layer weights. The meta-learning objective (outer-loop loss) can be either the physics-based least-squares error $l_\text{LSE}(\boldsymbol{w}^*) = \|\mathbf{A}\boldsymbol{w}^* - \mathbf{b}\|_2^2$, or the data-driven mean squared error $l_\text{MSE}(\boldsymbol{w}^*) = \frac{1}{n}\sum_s (u_s^{\text{label}} - \boldsymbol{f}(\mathbf{x};\boldsymbol{\tilde{w}})^T \boldsymbol{w}^*)^2$ given labeled data $\{u_s^{\text{label}}\}_{s=1}^n$ where $n$ is the total number of collocation points (PDE residual, IC, and BC). Here, Tikhonov regularization is used to compute the optimal task-specific output-layer weights $\boldsymbol{w}^*$, enabling the model to specialize to any realization of the task. $l_\text{LSE}$-based learning is termed \textbf{unsupervised learning}. $l_\text{MSE}$-based learning is termed \textbf{hybrid learning}, since it couples physics-based Tikhonov updates in the inner loop with data-driven minimization in the outer loop. See Algorithm \ref{alg:meta-learning} for the pseudocode and Appendix~\ref{app:Meta-learning PINN on training tasks} for a detailed mathematical description.

\paragraph{Gradient-free fine-tuning to new tasks:} After meta-learning, $\Theta$ is fixed. For a new task $\mathcal{T}_i \sim p(\mathcal{T})$ with different PDE parameters or IC/BC, Tikhonov regularization is used to compute the task-specific output-layer weights, enabling fast gradient-free adaptation to the new PDE instance. Since this update is independent of gradient-based backpropagation optimizers such as stochastic gradient descent (SGD), the resulting adaptation is extremely fast while remaining physics-compliant. See Algorithm~\ref{alg:inference} for the pseudocode.

\begin{figure}[h]
\centering

\begin{minipage}[t]{0.5\linewidth}
\footnotesize
\begin{algorithm}[H]
\caption{: Meta-learning on training tasks}
\label{alg:meta-learning}
\begin{algorithmic}
\Require Initialize $\Theta$; $p(\mathcal{T})$; $\eta$
\Require $\boldsymbol{\mathcal{C}}$, constructor of least-squares system $(\mathbf{A},\mathbf{b})$; see Eq.~(\ref{eq:maintext-L-S matrix})
\While{not done}
    \State Sample training tasks $\{\mathcal{T}_i\}\sim p(\mathcal{T})$
    \For{each task $\mathcal{T}_i$}
        \State Get initial guess $u^{(1)}$
        \For{$k=1$ to $\text{N}$} 
            \State $(\mathbf{A},\mathbf{b})=\boldsymbol{\mathcal{C}}(\mathcal{T}_i, \Theta,u^{k})$
            \State $\boldsymbol{w}=\!\begin{cases}
            (\lambda_\text{reg}\mathbf{I}+\mathbf{A}^T\mathbf{A})^{-1}\mathbf{A}^T\mathbf{b},
            & \text{or} \\
            \mathbf{A}^T(\lambda_\text{reg}\mathbf{I}+\mathbf{A}\mathbf{A}^T)^{-1}\mathbf{b}
            \end{cases}$
            \State $u^{(k+1)}=\boldsymbol{f}^T\boldsymbol{w}$
        \EndFor
        \State Compute loss $l_{\mathcal{T}_i}$
    \EndFor
    \State $\Theta\leftarrow\Theta - \eta\nabla_\Theta\sum_i l_{\mathcal{T}_i}$
\EndWhile
\end{algorithmic}
\end{algorithm}
\end{minipage}
\hfill
\begin{minipage}[t]{0.48\linewidth}
\small
\begin{algorithm}[H]
\caption{: Gradient-free fine-tuning to new tasks}
\label{alg:inference}
\begin{algorithmic}
\Require Freeze $\Theta$
\For{each test task $\mathcal{T}_i$}
    \State Get initial guess $u^{(1)}$
    \For{$k=1$ to $\text{N}$} 
        \State $(\mathbf{A},\mathbf{b})=\boldsymbol{\mathcal{C}}(\mathcal{T}_i,\Theta,u^{k})$
        \State $\boldsymbol{w}=\!\begin{cases}
        (\lambda_\text{reg}\mathbf{I}+\mathbf{A}^T\mathbf{A})^{-1}\mathbf{A}^T\mathbf{b}, & \text{or} \\
        \mathbf{A}^T(\lambda_\text{reg}\mathbf{I}+\mathbf{A}\mathbf{A}^T)^{-1}\mathbf{b}
        \end{cases}$
        \State $u^{(k+1)}=\boldsymbol{f}^T\boldsymbol{w}$
    \EndFor
    \State Store solution $u_{\mathcal{T}_i}=u^{(\text{N}+1)}$
\EndFor
\State \Return all task solution $\{u_{\mathcal{T}_i}\}$
\end{algorithmic}
\end{algorithm}
\end{minipage}
\end{figure}

\paragraph{Newton linearization:} Note that the Tikhonov regularization applies to linear PDEs, where a linear matrix system can be constructed and solved in a single step. For nonlinear PDEs, the nonlinear terms must first be linearized so that the system can be cast into a linear form. The previous approaches typically relied on Picard method (lagging-of-coefficients), which linearizes the nonlinear terms using the solution from the previous iteration and then performs least-squares solves for multiple nonlinear iterations. Although simple, it is only linearly convergent ~\citep{dennis1996numerical}. To address this limitation, we use the Newton linearization approach to approximate the nonlinear term by Taylor expansion around the current iterate. For a function $F$ of one state variable $u$, this gives $F(u^{k+1}) = F(u^k) + \left.\tfrac{d F}{d u}\right|^k (u^{k+1}-u^k) + \text{H.O.T}$, where $k$ is the nonlinear iteration step, and $\text{H.O.T}$ denotes the higher-order truncated terms of Taylor expansion. For a function of two variables $u$ and $v$, the expansion becomes 
$F(u^{k+1}, v^{k+1}) = F(u^k, v^k) + \left.\tfrac{\partial F}{\partial u}\right|^k (u^{k+1}-u^k) + \left.\tfrac{\partial F}{\partial v}\right|^k (v^{k+1}-v^k) + \text{H.O.T}$. Based on this principle, the nonlinear terms commonly arising in PDEs can be expressed in a Newton-linearized form amenable to Tikhonov regularization. Table~\ref{tab:newton-linearization} summarizes several representative nonlinear terms derived by our work. 

\begin{table}[h]
  \begin{center}
        \caption{Newton-linearized expressions of several nonlinear terms derived by our work. $m$ denotes the exponent. $k$ and $k+1$ refer to the current and next nonlinear iterations, respectively.}
        \label{tab:newton-linearization}
        \renewcommand{\arraystretch}{1.3} 
        \setlength{\tabcolsep}{2.5pt}
        \begin{tabular}{ll}
        Nonlinear term & \multicolumn{1}{c}{ Newton-linearized expression} \\  \toprule
        $(u^m)^{k+1}$ 
        & $m (u^k)^{m-1} u^{k+1} + (1-m)(u^k)^m$ \\ 
        $(u^m u_x)^{k+1}$ 
        & $m (u^k)^{m-1} u_x^k u^{k+1} + (u^k)^m u_x^{k+1} - m (u^k)^m u_x^k$ \\
        $[\sinh(u)]^{k+1}$ 
        & $\cosh(u^k) u^{k+1} + \sinh(u^k) - \cosh(u^k) u^k$ \\ 
        $[\exp(u)]^{k+1}$ 
        & $\exp(u^k) u^{k+1} + \exp(u^k) - \exp(u^k) u^k$ \\ 
        $[u \ln(u)]^{k+1}$ 
        & $(\ln(u^k) + 1) u^{k+1} + u^k \ln(u^k) - (\ln(u^k) + 1) u^k$ \\  \bottomrule
        \end{tabular}
  \end{center}
\vskip -0.1in
\end{table}

A detailed mathematical derivation is provided in Appendix~\ref{app:Linearization for nonlinear PDEs}. In Appendix~\ref{app:Convergence analysis}, we further establish that the Newton linearization used in the Tikhonov-regularized PINN is essentially equivalent to the classical regularized Gauss-Newton method, with experiments validating its near-quadratic convergence. 

\section{Experiment results}
\label{Experiment results}
We compare the performance of the proposed \textbf{Newton-PINet} (Newton linearization), \textbf{PINet} (Picard linearization) and purely data-driven deep neural network (DNN) on several representative classes of nonlinear PDEs:

(i) Nonlinear convection-type PDEs: $\tfrac{\partial u}{\partial t} + \beta u^{m} \tfrac{\partial u}{\partial x} - \gamma \tfrac{\partial^{2} u}{\partial x^{2}} + \delta \tfrac{\partial^{3} u}{\partial x^{3}} + \sigma \tfrac{\partial^{4} u}{\partial x^{4}} = 0$, 

where $u(x,t)$ is the state variable, $\beta, \gamma, \delta, \sigma$ are PDE parameters, and $m$ denotes the nonlinearity order. We test three 1D time-dependent problems: Burgers, generalized Korteweg-de Vries (KdV), and Kuramoto-Sivashinsky (K-S) equations. In addition, we consider a 2D lid-driven cavity (LDC) flow governed by the Navier-Stokes equations. These problems span nonlinear convection, high-order dispersion/dissipation, and viscous flow with pressure-velocity coupling, providing canonical benchmarks for assessing generalization across different nonlinearities and dimensions.  

(ii) Nonlinear forcing-type PDEs: $\frac{\partial u}{\partial t} + \alpha\nabla u - \gamma\Delta u + R(u) = q,$

where $R(u)$ denotes a nonlinear reaction operator, which may take polynomial, hyperbolic, exponential, or logarithmic forms, and $q$ represents an external source term. We consider four 1D time-dependent problems: convection-diffusion-reaction (CDR), Klein-Gordon (K-G), hyperbolic heat, and logarithmic heat equations; two 2D problems: Helmholtz and parametric diffusion-reaction equations; and one 3D problem: parametric diffusion-reaction equation. These equations represent systems where diffusion and wave propagation interact with nonlinear reaction or source terms, making them central to the modeling of chemical kinetics, heat transfer, quantum fields, and complex reaction-diffusion phenomena in physics and engineering. 

See Appendix~\ref{app:PDE_problems} for detailed descriptions of all PDE problems and meta-learning configurations (summarized in Table~\ref{tab:Experimental config summary}). See Appendix~\ref{app:Data generation and computational setup} for details on data generation, error metrics, and computational setup. Note that, unless otherwise specified, the meta-learning outer-loop loss function for the Newton-PINet is $l_\text{MSE}$ (hybrid learning). We also adopt a temporal domain decomposition strategy within our meta-learning framework to leverage temporal causality~\citep{wang2022respecting} as a means of improving accuracy for complex time-dependent PDEs. Each time block is trained to model only short-term dynamics; however, iterative composition across blocks during inference ensures seamless integration over the full temporal horizon.  Table~\ref{tab:Experimental-summary} below summarizes the performance of the models on the nonlinear PDEs considered in this study. 

\begin{table}[h]
\caption{Summary of model performance on nonlinear PDEs in this study. The reported meta-learning and fine-tuning time correspond to the Newton-PINet. ``NaN'' entries under PINet indicate numerical overflow during computation. See Appendix~\ref{app:PDE_problems} for detailed descriptions of all PDE problems and meta-learning configurations (summarized in Table~\ref{tab:Experimental config summary}). See Appendix~\ref{app:Data generation and computational setup} for details on data generation, error metrics, and computational setup.}
\label{tab:Experimental-summary}
\footnotesize 
\setlength{\tabcolsep}{3pt} 
\begin{center}
\begin{tabular}{lccccccc}
\multirow[b]{2}{*}{Problem} & \multirow[b]{2}{*}{\begin{tabular}[c]{@{}c@{}}Meta-learning \\ epoch\end{tabular}}  & \multirow[b]{2}{*}{\begin{tabular}[c]{@{}c@{}}Meta-learning \\ time cost\end{tabular}} & \multirow[b]{2}{*}{\begin{tabular}[c]{@{}c@{}}Inference time \\ per test task\end{tabular}} & \multicolumn{3}{c}{Test MSE of models}          \\ \cmidrule{5-7}
                                                                &                                                                             &                                                                                     &   & DNN      & PINet    & Newton-PINet \\ \toprule
Burgers'         &1000   & 286s                                                                                & 0.035s                                                                                 & 1.06e-4 & 1.10e-4 & 1.79e-6     \\ \midrule
Generalized KdV          &1000        & 453s                                                                                & 0.359s                                                                                 & 3.74e-2 & 1.49e-2 & 1.94e-3     \\ \midrule
Kuramoto-Sivashinsky     &3000     & 1014s                                                                               & 0.552s                                                                                 & 5.03e-1 & 106      & 3.45e-2     \\ \midrule
{\begin{tabular}[c]{@{}l@{}}Convection-diffusion- \\  reaction \end{tabular}}       &1000                                               & 203s                                                                                & 0.099s                                                                                 & 5.27e-1 & 4.65e-3 & 4.32e-5     \\ \midrule
Klein-Gordon          &1000                                               & 105s                                                                                & 0.065s                                                                                 & 1.04     & 1.88e-7 & 2.36e-7     \\ \midrule
\begin{tabular}[c]{@{}l@{}} Hyperbolic heat\end{tabular}  &2000  & 393s                                                                                & 0.069s                                                                                 & 1.06e-1 & NaN      & 1.50e-6     \\ \midrule
\begin{tabular}[c]{@{}l@{}} Logarithmic heat \end{tabular}  &2000   & 395s                                                                                & 0.067s                                                                                 & 2.99e-1 & NaN      & 2.16e-6     \\ \midrule
Helmholtz                                   &2000                  & 344s                                                                                & 0.067s                                                                                 & 5.98     & NaN      & 1.33e-3     \\ \midrule

\begin{tabular}[c]{@{}l@{}}Parametric\\ diffusion-reaction (2D)\end{tabular} &2000  & 342s                                                                                & 0.047s                                                                                & 3.43e-1 & 1.18e-8 & 1.60e-10                 \\ \midrule

\begin{tabular}[c]{@{}l@{}} Lid-driven cavity\end{tabular}  &1000  & 290s                                                                               & 0.40s                                                                                 & 1.42e-1     & 1.92e-4 & 1.67e-4 \\ \midrule

\begin{tabular}[c]{@{}l@{}}Parametric\\ diffusion-reaction (3D)\end{tabular} &2000  & 1907s                                                                                & 0.104s                                                                                & 7.82e-1  & 4.58e-5 & 1.72e-6                 \\ \bottomrule

\end{tabular}
\end{center}
\end{table}

\subsection{Learning to solve PDEs with nonlinear convection term}

\begin{figure*}[htb]
  \vskip 0.2in
    \begin{center}
        \includegraphics[width=1\linewidth]{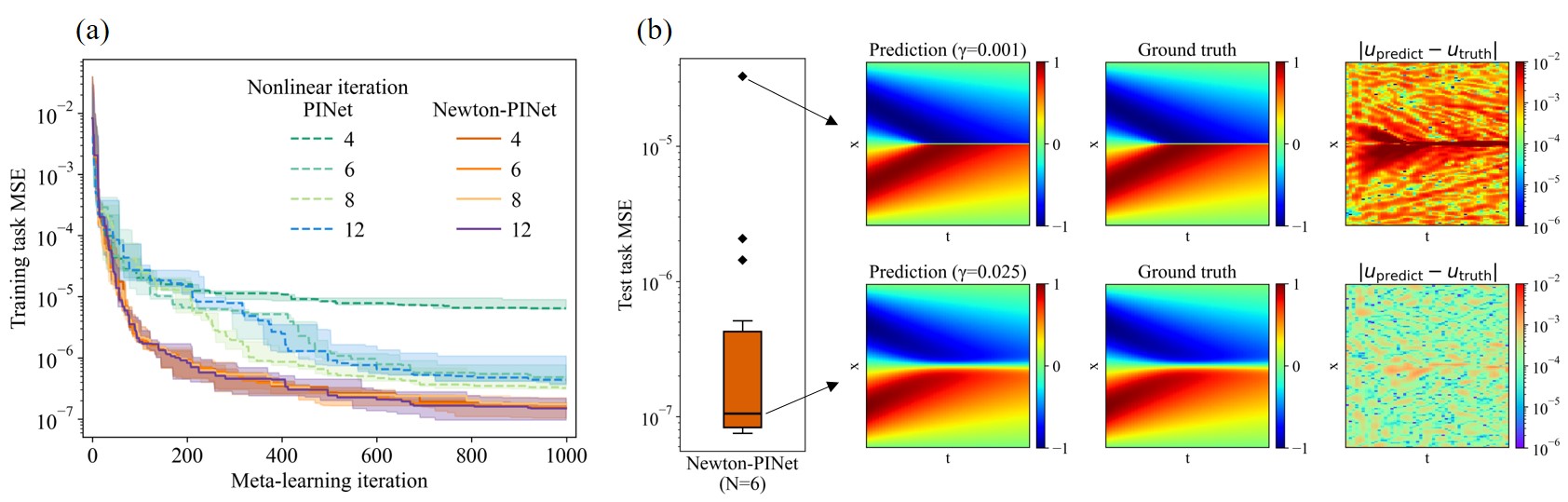}
        \caption{\label{fig:Burgers-meta-learning and test results} 
                Burgers' problem. (a) Meta-learning convergence of Newton-PINet vs. PINet under different nonlinear iterations (N) (lines and shaded areas: the median convergence path and interquartile ranges of 5 random-seed runs). (b) Boxplot of error distributions across all test tasks and predictions by Newton-PINet (N=6).}
    \end{center}
\end{figure*}

\paragraph{Burgers' problems:} $\tfrac{\partial u}{\partial t} +  u\tfrac{\partial u}{\partial x} - \gamma \tfrac{\partial^{2} u}{\partial x^{2}}  = 0$.
We first fix the initial condition as $u(x,0) = -\sin(\pi x)$, and vary the viscosity parameter $\gamma$ in the range $[0.001,0.05]$ with an increment of $0.001$ to generate 50 tasks. Among them, 16 are randomly selected for training, and the remaining 34 are used as test tasks. Under each condition, we keep the number of nonlinear iterations ($\mathrm{N}$) for inner-loop fine-tuning in meta-learning stage consistent with those in generalization stage. We then compare Newton-PINet and PINet across different conditions, with the number of nonlinear iterations ($\mathrm{N}$) ranging from 4 to 12. As shown in Figure~\ref{fig:Burgers-meta-learning and test results}(a), with an increase of $\mathrm{N}$, Newton-PINet achieves significantly faster convergence during meta-learning compared to PINet. Meanwhile, Newton-PINet achieves lower test errors than PINet. For example, when $\mathrm{N}=6$, the average MSE across all test tasks of PINet is $1.10\times10^{-4}$, whereas that of Newton-PINet is $1.79\times10^{-6}$ (see Figure~\ref{fig:Burgers-meta-learning and test results}(b) for the prediction plots). These results demonstrate that Newton linearization significantly improves the convergence of meta-learning, which in turn enhances generalization accuracy on test tasks.

\paragraph{Model comparison under varying initial conditions:} We evaluate the model under the setting $\gamma = 0.01$ in Burgers' problem, where the initial conditions are sampled from a Gaussian random field $\mathcal{N}(0,625(-\Delta+25I)^{-4})$ to construct both training and test tasks. We compare (i) \textbf{unsupervised Newton-PINet} (1 time block) and (ii) \textbf{hybrid Newton-PINet} (4 time blocks) against popular neural operator baselines in terms of training time and test error. The \textbf{unsupervised} operator considered is Physics-Only DeepONet (PO-DeepONet)~\citep{wang2021learning}. The \textbf{supervised} operators include DeepONet~\citep{lu2021learning} and Transformer-DeepONet with Trunk net enhanced by Fourier coefficients (T-DeepONet-TF)~\citep{wei2025efficient}, while the \textbf{hybrid} operator is PINO (a physics-informed FNO)~\citep{li2024physics}. Note that PINO uses a training loss that combines both data-driven and PDE-based physics losses. The results are summarized in Table~\ref{tab:Burgers-transpose}. Compared to PO-DeepONet, unsupervised Newton-PINet reduces training time from 9.25h to 291s while improving test accuracy (in a data-absent scenario). Compared to supervised and hybrid neural operators, hybrid Newton-PINet consistently achieves superior test accuracy, with significantly reduced training cost. More results under different $\gamma$ configurations are reported in Appendix Table~\ref{tab:Burgers-Compar}.

\begin{table}[h]
\centering
\footnotesize
\caption{Comparison of training cost and performance at $\gamma=0.01$ on Burgers' problem.}
\label{tab:Burgers-transpose}
\setlength{\tabcolsep}{4pt}
\begin{tabular}{lcc|cc cc}
\toprule
& \multicolumn{2}{c|}{Unsupervised} 
& \multicolumn{2}{c}{Supervised} 
& \multicolumn{2}{c}{Hybrid} \\ \cmidrule(lr){2-3} \cmidrule(lr){4-5} \cmidrule(lr){6-7}

& PO-DeepONet & \textbf{Newton-PINet} 
& DeepONet & T-DeepONet-TF 
& PINO & \textbf{Newton-PINet} \\ \midrule

No. training tasks 
& 1000 & 50 
& 1000 & 1000 
& 1000 & 50 \\

Test relative $L^2$ error 
& 1.38e-2 & \textbf{3.51e-3}
& 1.17e-2 & 2.08e-3
& 3.80e-3 & \textbf{5.46e-4} \\

Training time 
& 9.25h & \textbf{291s}
& 2800s & 3041s 
& 1200s & \textbf{133s} \\

\bottomrule
\end{tabular}
\vskip -0.1in
\end{table}

\paragraph{Generalized KdV, K-S, and LDC problems:} We further evaluate Newton-PINet on more complex nonlinear PDEs, including the generalized KdV equations with higher-order nonlinear terms $u \tfrac{\partial u}{\partial x}$, $u^{2} \tfrac{\partial u}{\partial x}$, $u^{3} \tfrac{\partial u}{\partial x}$, the K-S equations with nonlinear term $u \tfrac{\partial u}{\partial x}$, and the LDC problem governed by the Navier-Stokes equations. These PDEs pose highly challenging benchmarks for PINNs. The state-of-the-art PirateNets+SOAP~\citep{wang2025gradient} requires over 24 hours of training to solve a single K-S problem and must be retrained for each new problem. In contrast, our \textbf{Newton-PINet} completes the meta-learning stage in \textit{under half an hour}, while single-task adaptation takes \textit{less than one second}, demonstrating a clear advantage in computational efficiency. The results show that Newton-PINet achieves test errors of $\mathrm{MSE}=1.94\times 10^{-3}$ on the generalized KdV, $\mathrm{MSE}=3.45\times 10^{-2}$ on the K-S, and $\mathrm{MSE}=1.67\times 10^{-4}$ on the LDC problem, outperforming PINet and DNN under the same configuration. Figures~\ref{fig:KDV-Visualization plot}, ~\ref{fig:KS-Visualization plot}, ~\ref{fig:LDC-Re25-100-500-900} show the prediction results of these problems. This demonstrates that our Newton linearization can improve the model performance across a variety of nonlinear convective PDEs. Note that the regularization parameter ($\lambda_{\mathrm{reg}}$) is meta-learned rather than manually tuned, and is shown to be robust across diverse test tasks (Appendix~\ref{app:Robustness of the meta-learned Tikhonov regularization parameter}). The initial guess ($u_{\mathrm{guess}}$) may affect convergence and Appendix~\ref{app:Effect of initial guess in nonlinear iterations} provides simple strategies to mitigate this sensitivity. In addition, since the Tikhonov solution depends on $\mathbf{A}$, Appendix~\ref{app:Effect of the number of columns in A on model inference time} analyzes its impact on inference time, showing that larger $\mathbf{A}$ increases the computational cost.

\subsection{Learning to solve PDEs with nonlinear forcing term}\vspace{-5pt} 

\begin{figure*}[htb]
\vskip 0.2in
\begin{center}
    \includegraphics[width=1\linewidth]{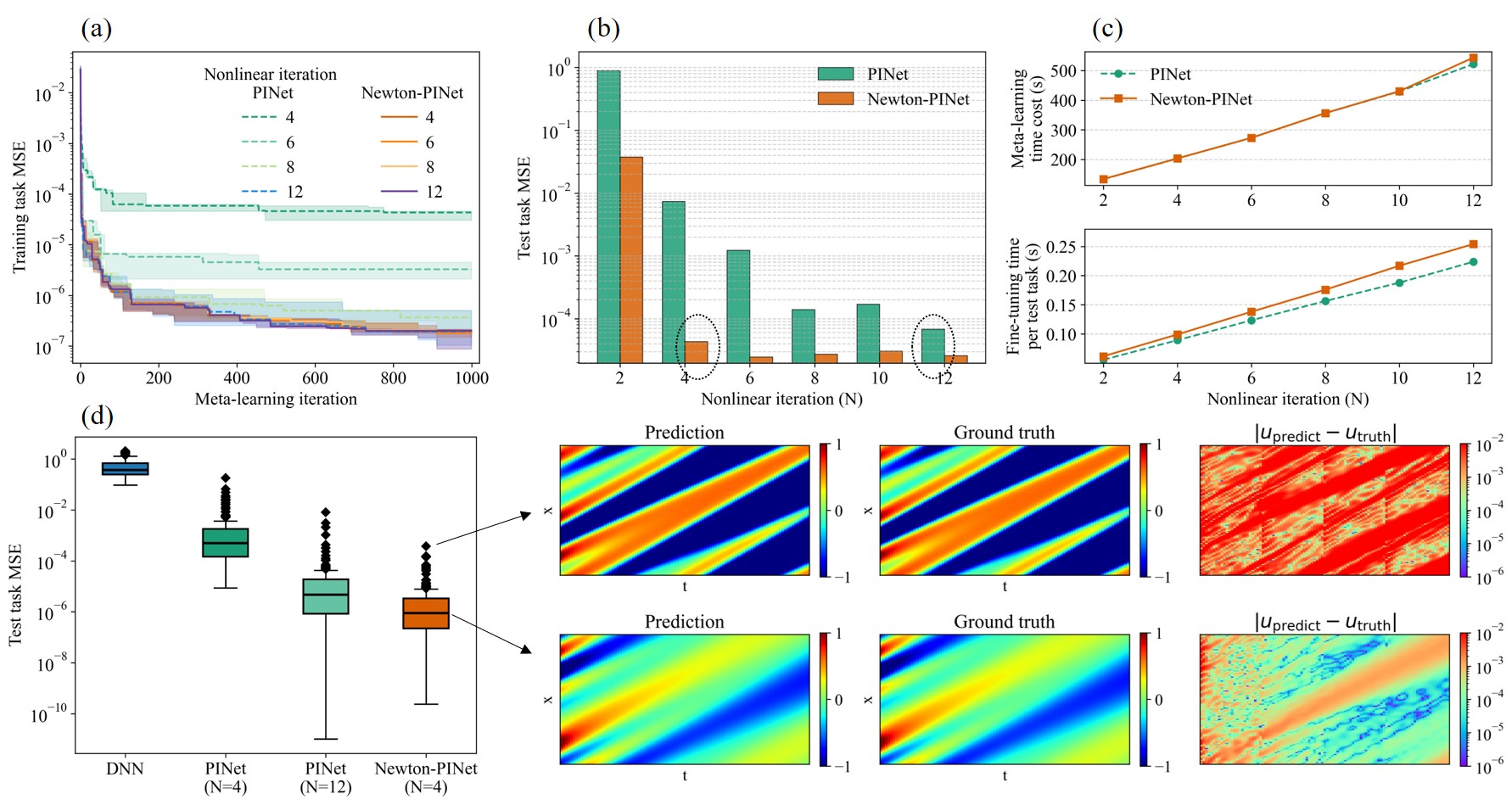}
    \caption{\label{fig:CDR-meta-learning and test results} Convection-diffusion-reaction (CDR) problem. (a) Meta-learning convergence of Newton-PINet vs. PINet under different nonlinear iterations (N). (b,c) Test task MSE and runtime aggregated from 5 runs. In (b), the two circled regions correspond to Newton-PINet (N=4) and PINet (N=12), respectively. (d) Error distribution boxplot across all test tasks for different models; the prediction fields correspond to Newton-PINet (N=4).}
\end{center}
\end{figure*}

\paragraph{Convection-diffusion-reaction (CDR) problem:} $\frac{\partial u}{\partial t} + \alpha \frac{\partial u}{\partial x} - \gamma \frac{\partial^2 u}{\partial x^2} - \rho_1 u + \rho_2 u^2 + \rho_3 u^3 = 0$. We generate 40 training and 160 test tasks by varying PDE parameters and initial conditions: $\alpha = 1$, $\gamma \in \{0.005, 0.01, 0.05\}$, $\rho_1, \rho_2 \in \{0, 1, 3, 5\}$, $\rho_3 = 5$, and $u(x,t=0) = \sum_{j=1}^{J} A_j \sin\left(l_j x + \phi_j\right)$, where $J=5$, $A_{j} \in [0.1,0.5]$, $l_{j} \in \{1,2,3,4\}$, and $\phi_{j} \in [-\pi,\pi]$. We adopt a 4-time-block decomposition during meta-training and inference to improve accuracy. Figure~\ref{fig:CDR-meta-learning and test results} (a) and (b) show the meta-learning process and test results. It can be seen that PINet converges well during meta-learning, and its test error decreases as nonlinear iteration (N) increases. In contrast, Newton-PINet converges much faster and achieves lower test error. As shown in Figure~\ref{fig:CDR-meta-learning and test results} (c) and (d), for a comparable error level, Newton-PINet ($\mathrm{N}=4$) requires less than half the meta-training and fine-tuning time of PINet ($\mathrm{N}=12$). These results indicate that Newton linearization can substantially accelerate nonlinear iterations without significantly introducing additional time cost in either meta-learning or inference. More prediction results under various test conditions are provided in Appendix Figure~\ref{fig:CDR-Visualization plot}. Note that different time-blocking strategies can affect model inference performance, as discussed in Appendix~\ref{app:different time-blocking strategies}. As shown in Figure~\ref{fig:CDR-meta-learning and test results}, the predictions exhibit drift at temporal block boundaries. We can mitigate this drift by increasing the initial condition weight, as detailed in Appendix~\ref{app:Effect of initial condition weight parameter on test errors}.

\paragraph{Additional nonlinear forcing-type PDEs:} We evaluate Newton-PINet on several other challenging nonlinear forcing-type PDEs: Klein-Gordon, hyperbolic/logarithmic heat, Helmholtz, and parametric diffusion-reaction equations. The results show that Newton-PINet outperforms the DNN and PINet baselines on nearly all problems. Notably, for problems with hyperbolic or logarithmic nonlinearities, the Picard approach used by PINet often fails to converge or leads to unstable training, whereas Newton-PINet remains stable and accurate. Additional results for these problems are provided in Appendix~\ref{app:Summary_of_model_performance} and Figure~\ref{fig:KG-Heat-DR-Visualization plot}. We further demonstrate the model's applicability to different boundary condition types in Appendix~\ref{app:Effect of boundary condition types}.

\paragraph{Model generalization to unseen boundary conditions and irregular geometries:} We verify that the model, pretrained only on samples with \emph{Neumann} boundary conditions, can effectively generalize to unseen tasks with \emph{Dirichlet} and mixed boundary conditions (see Appendix~\ref{app:Summary_of_model_performance} and Table~\ref{tab:heat_hyperbolic_BC_generalization}). Furthermore, the model, pretrained only on samples with regular geometry, can effectively generalize to test cases with complex and varying irregular geometries, as detailed in Appendix~\ref{app:Summary_of_model_performance}, Tables~\ref{tab:heat_hyperbolic_shape_param} and~\ref{tab:heat_hyperbolic_shape_param_3d}, Figs.~\ref{fig:Hyperbolic Heat problem across varying domain shapes} and~\ref{fig:3D parametric diffusion-reaction test}.

\paragraph{Model comparison on a nonlinear reaction-diffusion problem:} 
We further evaluate our method on a challenging 1D nonlinear reaction-diffusion problem 
$\frac{\partial u}{\partial t} - \gamma \frac{\partial^2 u}{\partial x^2} - \rho u(1-u)= 0$~\citep{krishnapriyan2021characterizing}. In this benchmark, the initial condition is fixed, while the PDE parameters are sampled over $\gamma \in [1,5]$ and $\rho \in [-5,5]$. The benchmark setup and baseline results are taken from the physics-informed neural solver (PI-neural-solver)~\citep{boudec2025learning}, which is trained on 800 labeled tasks and evaluated on 200 unseen tasks. In contrast, our \textbf{Newton-PINet}, under a hybrid training scheme, uses only 50 labeled tasks, i.e., \emph{16$\times$ fewer tasks}, and is evaluated on the same test set. For a fair comparison, both models are trained and evaluated on an NVIDIA A6000 GPU. As shown in Table~\ref{tab:maintext-reaction_diffusion_model_compare}, Newton-PINet achieves a test MSE that is approximately three orders of magnitude lower than that of the PI-neural-solver, while requiring only 301 seconds (5 minutes) of total meta-training wall-clock time, compared to several hours for the baseline. These results highlight the \textit{high data efficiency and computational efficiency} of Newton-PINet. 
More detailed comparisons with additional baselines, as well as model performance under different GPU configurations, are provided in Appendix~\ref{app:Model comparison under varying PDE parameters on a nonlinear reaction-diffusion problem} and~\ref{app:Model comparison under different GPU configurations}. Model prediction results are presented in Appendix Figure~\ref{fig:ICLR_DR-Visualization plot}.

\begin{table*}[htbp]
\caption{Model comparison on a 1D nonlinear reaction-diffusion problem.}
\label{tab:maintext-reaction_diffusion_model_compare}
\begin{center}
\setlength{\tabcolsep}{4.5pt} 
\begin{tabular}{llc ccc}
                              &\multicolumn{1}{l}{Model}                     & \begin{tabular}[c]{@{}c@{}}No. training tasks\end{tabular} & \begin{tabular}[c]{@{}c@{}}Test relative MSE\end{tabular} & \begin{tabular}[c]{@{}c@{}}Training time\end{tabular} & \begin{tabular}[c]{@{}c@{}}Inference time\end{tabular} \\ \toprule

                              &  \begin{tabular}[l]{@{}l@{}}PI-neural-solver\\ \citep{boudec2025learning}\end{tabular}    &  \begin{tabular}[c]{@{}c@{}}800 \end{tabular}                                                              & 2.91e-4                 & 4h30m                          & 0.284s               \\ \midrule
                              & \textbf{Newton-PINet} (Ours) & \begin{tabular}[c]{@{}c@{}}50 \end{tabular}                       & 1.72e-7                     & 301s                  & 0.080s                  \\  \bottomrule
\end{tabular}

\end{center}

\end{table*}

\paragraph{Comparison with other meta-learning PINN and neural operator methods:}
We further compare the performance of Newton-PINet with state-of-the-art gradient-based meta-learning PINNs \citep{penwarden2023metalearning} and gradient-free Baldwinian-PINN \citep{wong2026evolutionary}. Detailed results are provided in Appendix~\ref{app:Model comparison with other meta-learning PINN methods} and Table~\ref{tab:Baldwin-PINN-Meta-PINN-compare}. On a parametric diffusion-reaction problem, Newton-PINet achieves approximately a \textit{179$\times$ improvement in generalization accuracy} while requiring \textit{7000$\times$ less fine-tuning time} compared with gradient-based meta-learning PINNs. Moreover, relative to Baldwinian-PINN, Newton-PINet reduces test MSE by 2$\sim$5 orders of magnitude across benchmark problems and shortens the task-specific fine-tuning time by roughly one order of magnitude. These results highlight Newton-PINet's superior generalization performance and computational speed. We next evaluate the performance of our method and state-of-the-art neural operators on the newly introduced PDE benchmarks, including the generalized KdV, K-S, CDR, and LDC problems. The results show that Newton-PINet achieves higher generalization accuracy than neural operators under identical training configurations. Detailed results are provided in Appendix~\ref{app:Model comparison with neural operators on the newly introduced datasets} and Table~\ref{tab:new-data model_comparison}.
Furthermore, Appendix~\ref{app:Runtime comparison with classical numerical solvers} compares the inference time of Newton-PINet with the runtime of classical numerical solvers, demonstrating substantial speedups achieved by our model while maintaining comparable accuracy.

In Appendix~\ref{app:Ablation_studies_on_model_architecture},
we demonstrate that Newton-PINet benefits significantly from skip connections, which enhance robustness to depth, width, and mesh resolution, thereby reducing the need for extensive hyperparameter tuning. In Appendix~\ref{app:Practical scaling discussion of Newton-PINet}, we discuss Newton-PINet’s scalability. Briefly, since the features are generated by a neural representation, each row of $\mathbf{A}$ is dense rather than sparse in the numerical sense. The computational cost primarily depends on the expressiveness of pre-final layer features. By meta-learning a compact yet informative representation, the least-squares solve remains efficient even for large collocation sets.

\section{Limitations \& Outlook}

Current experiments focus on 1D+time, 2D, and 3D cases. For more complex 2D/3D+time or multi-physics systems, larger, denser, and more ill-conditioned $\mathbf{A}$ matrix, combined with GPU memory limitations, hinders efficient solution. Future work will focus on developing low-rank strategies and scalable solvers for efficiently solving dense neural feature-based least-squares problems, thereby enhancing overall model performance. Additionally, for long-time prediction of chaotic systems such as K-S, we currently require 15 temporal blocks for accurate inference. Enabling single-shot prediction or reducing dependence on temporal decomposition remains an open challenge.

\section{Conclusions}

Newton-PINet has been validated on 1D, 2D, and 3D PDEs, as well as across diverse boundary conditions and domain geometries, covering a broad range of challenging nonlinear PDE operators. Newton-PINet demonstrates superior computational efficiency compared with state-of-the-art neural operators and meta-learning PINN methods. The main strengths of our model can be summarized as follows. \textbf{(1) Computational efficiency}: task adaptation in both meta-learning and inference is performed via least-squares updates of only the output layer. This gradient-free fine-tuning is faster than typical methods such as SGD, requiring just one step for linear PDEs and a few iterations for nonlinear PDEs. The Newton linearization further accelerates nonlinear solves. \textbf{(2) Data efficiency}: because the least-squares fine-tuning for new-task generalization is physics-informed, the meta-learning stage of our model requires very few labeled training samples and can converge in only a few epochs. \textbf{(3) Accuracy}: Tikhonov regularization stabilizes the least-squares solve, reducing ill-conditioning and improving generalization accuracy. Altogether, Newton-PINet achieves high generalization accuracy while requiring an order of magnitude fewer training tasks than state-of-the-art baselines. Task-specific inference on new tasks is also orders of magnitude faster compared to gradient-based meta-learning PINNs. These advantages make Newton-PINet a practical and scalable framework for learning large families of nonlinear PDEs in few-shot and real-time scenarios.

\newpage

\begin{ack}
This research was in part supported by the National Research Foundation, Singapore under its AI-for-Science (AI4S) Catalytic Grant (Award No.: NRF-AI4SCT\_2025-0007), and in part supported by the National Research Foundation, Singapore through the AI Singapore Programme, under the project ``AI-based urban cooling technology development'' (Award No.: AISG3-TC-2024-014-SGKR). The authors declare no competing interests.
\end{ack}



{\small
\bibliographystyle{plainnat}
\bibliography{example_paper}

@article{cai2021physicsa,
  title={Physics-informed neural networks (PINNs) for fluid mechanics: A review},
  author={Cai, Shengze and Mao, Zhiping and Wang, Zhicheng and Yin, Minglang and Karniadakis, George Em},
  journal={Acta Mechanica Sinica},
  volume={37},
  number={12},
  pages={1727--1738},
  year={2021},
  publisher={Springer}
}

@article{xu2025fp64,
  title={FP64 is All You Need: Rethinking Failure Modes in Physics-Informed Neural Networks},
  author={Xu, Chenhui and Liu, Dancheng and Nassereldine, Amir and Xiong, Jinjun},
  journal={Advances in Neural Information Processing Systems},
  year={2025}
}

@article{datar2024fast,
  title={Fast training of accurate physics-informed neural networks without gradient descent},
  author={Datar, Chinmay and Kapoor, Taniya and Chandra, Abhishek and Sun, Qing and Bolager, Erik Lien and Burak, Iryna and Veselovska, Anna and Fornasier, Massimo and Dietrich, Felix},
  journal={arXiv preprint arXiv:2405.20836},
  year={2024}
}

@inproceedings{calm-pde-hagnberger:2025,
      title={{CALM-PDE}: Continuous and {A}daptive {C}onvolutions for {L}atent {S}pace {M}odeling of {T}ime-dependent {PDE}s}, 
      author={Jan Hagnberger and Daniel Musekamp and Mathias Niepert},
      booktitle={Advances in Neural Information Processing Systems},
      volume={38},
      year={2025}
}

@inproceedings{hu2026realpdebench,
      title={RealPDEBench: A Benchmark for Complex Physical Systems with Real-World Data}, 
      author={Peiyan Hu and Haodong Feng and Hongyuan Liu and Tongtong Yan and Wenhao Deng and Tianrun Gao and Rong Zheng and Haoren Zheng and Chenglei Yu and Chuanrui Wang and Kaiwen Li and Zhi-Ming Ma and Dezhi Zhou and Xingcai Lu and Dixia Fan and Tailin Wu},
      booktitle={The Fourteenth International Conference on Learning Representations},
      year={2026},
      url={https://openreview.net/forum?id=y3oHMcoItR},
      note={Oral Presentation}
}

@inproceedings{zhou2026tucker,
  title={Tucker-FNO: Tensor Tucker-Fourier Neural Operator and its Universal Approximation Theory},
  author={Zhou, Guancheng and Zeng, Zelin and Luo, Yisi and Xie, Qi and Meng, Deyu},
  booktitle={The Fourteenth International Conference on Learning Representations},
  year={2026}
}

@article{bischof2025hypino,
  title={HyPINO: Multi-Physics Neural Operators via HyperPINNs and the Method of Manufactured Solutions},
  author={Bischof, Rafael and Piovar{\v{c}}i, Michal and Kraus, Michael A and Mishra, Siddhartha and Bickel, Bernd},
  journal={arXiv preprint arXiv:2509.05117},
  year={2025}
}

@inproceedings{tran2023factorized,
  title     = {Factorized Fourier Neural Operators},
  author    = {Alasdair Tran and Alexander Mathews and Lexing Xie and Cheng Soon Ong},
  booktitle = {The Eleventh International Conference on Learning Representations},
  year      = {2023},
  url       = {https://openreview.net/forum?id=tmIiMPl4IPa}
}

@article{li2024physics,
  title={Physics-informed neural operator for learning partial differential equations},
  author={Li, Zongyi and Zheng, Hongkai and Kovachki, Nikola and Jin, David and Chen, Haoxuan and Liu, Burigede and Azizzadenesheli, Kamyar and Anandkumar, Anima},
  journal={ACM/IMS Journal of Data Science},
  volume={1},
  number={3},
  pages={1--27},
  year={2024},
  publisher={ACM New York, NY}
}

@article{li2025d,
  title={D-FNO: A decomposed Fourier neural operator for large-scale parametric partial differential equations},
  author={Li, Kangjie and Ye, Wenjing},
  journal={Computer Methods in Applied Mechanics and Engineering},
  volume={436},
  pages={117732},
  year={2025},
  publisher={Elsevier}
}

@article{qiu2024derivative,
  title={Derivative-enhanced deep operator network},
  author={Qiu, Yuan and Bridges, Nolan and Chen, Peng},
  journal={Advances in Neural Information Processing Systems},
  volume={37},
  pages={20945--20981},
  year={2024}
}

@inproceedings{kim2021dpm,
  title={DPM: A novel training method for physics-informed neural networks in extrapolation},
  author={Kim, Jungeun and Lee, Kookjin and Lee, Dongeun and Jhin, Sheo Yon and Park, Noseong},
  booktitle={Proceedings of the AAAI conference on artificial intelligence},
  volume={35},
  number={9},
  pages={8146--8154},
  year={2021}
}

@article{dong2021local,
  title={Local extreme learning machines and domain decomposition for solving linear and nonlinear partial differential equations},
  author={Dong, Suchuan and Li, Zongwei},
  journal={Computer Methods in Applied Mechanics and Engineering},
  volume={387},
  pages={114129},
  year={2021},
  publisher={Elsevier}
}

@article{golub1999tikhonov,
  title={Tikhonov regularization and total least squares},
  author={Golub, Gene H and Hansen, Per Christian and O'Leary, Dianne P},
  journal={SIAM journal on matrix analysis and applications},
  volume={21},
  number={1},
  pages={185--194},
  year={1999},
  publisher={SIAM}
}

@article{wei2025ffv,
  title={FFV-PINN: A fast physics-informed neural network with simplified finite volume discretization and residual correction},
  author={Wei, Chang and Fan, Yuchen and Wong, Jian Cheng and Ooi, Chin Chun and Wang, Heyang and Chiu, Pao-Hsiung},
  journal={Computer Methods in Applied Mechanics and Engineering},
  volume={444},
  pages={118139},
  year={2025},
  publisher={Elsevier}
}

@article{karniadakis2021physics,
  title={Physics-informed machine learning},
  author={Karniadakis, George Em and Kevrekidis, Ioannis G and Lu, Lu and Perdikaris, Paris and Wang, Sifan and Yang, Liu},
  journal={Nature Reviews Physics},
  volume={3},
  number={6},
  pages={422--440},
  year={2021},
  publisher={Nature Publishing Group UK London}
}

@article{raissi2019physics,
  title={Physics-informed neural networks: A deep learning framework for solving forward and inverse problems involving nonlinear partial differential equations},
  author={Raissi, Maziar and Perdikaris, Paris and Karniadakis, George E},
  journal={Journal of Computational physics},
  volume={378},
  pages={686--707},
  year={2019},
  publisher={Elsevier}
}

@article{cuomo2022scientific,
  title={Scientific machine learning through physics--informed neural networks: Where we are and what’s next},
  author={Cuomo, Salvatore and Di Cola, Vincenzo Schiano and Giampaolo, Fabio and Rozza, Gianluigi and Raissi, Maziar and Piccialli, Francesco},
  journal={Journal of Scientific Computing},
  volume={92},
  number={3},
  pages={88},
  year={2022},
  publisher={Springer}
}

@article{lu2021learning,
  title={Learning nonlinear operators via DeepONet based on the universal approximation theorem of operators},
  author={Lu, Lu and Jin, Pengzhan and Pang, Guofei and Zhang, Zhongqiang and Karniadakis, George Em},
  journal={Nature machine intelligence},
  volume={3},
  number={3},
  pages={218--229},
  year={2021},
  publisher={Nature Publishing Group UK London}
}

@article{li2020fourier,
  title={Fourier neural operator for parametric partial differential equations},
  author={Li, Zongyi and Kovachki, Nikola and Azizzadenesheli, Kamyar and Liu, Burigede and Bhattacharya, Kaushik and Stuart, Andrew and Anandkumar, Anima},
  journal={arXiv preprint arXiv:2010.08895},
  year={2020}
}

@article{wang2021learning,
  title={Learning the solution operator of parametric partial differential equations with physics-informed DeepONets},
  author={Wang, Sifan and Wang, Hanwen and Perdikaris, Paris},
  journal={Science advances},
  volume={7},
  number={40},
  pages={eabi8605},
  year={2021},
  publisher={American Association for the Advancement of Science}
}

@article{wang2024piratenets,
  title={Piratenets: Physics-informed deep learning with residual adaptive networks},
  author={Wang, Sifan and Li, Bowen and Chen, Yuhan and Perdikaris, Paris},
  journal={Journal of Machine Learning Research},
  volume={25},
  number={402},
  pages={1--51},
  year={2024}
}

@article{raghu2019rapid,
  title={Rapid learning or feature reuse? towards understanding the effectiveness of maml},
  author={Raghu, Aniruddh and Raghu, Maithra and Bengio, Samy and Vinyals, Oriol},
  journal={arXiv preprint arXiv:1909.09157},
  year={2019}
}

@inproceedings{cyr2020robust,
  title={Robust training and initialization of deep neural networks: An adaptive basis viewpoint},
  author={Cyr, Eric C and Gulian, Mamikon A and Patel, Ravi G and Perego, Mauro and Trask, Nathaniel A},
  booktitle={Mathematical and Scientific Machine Learning},
  pages={512--536},
  year={2020},
  organization={PMLR}
}

@article{wong2022learning,
  title={Learning in sinusoidal spaces with physics-informed neural networks},
  author={Wong, Jian Cheng and Ooi, Chin Chun and Gupta, Abhishek and Ong, Yew-Soon},
  journal={IEEE Transactions on Artificial Intelligence},
  volume={5},
  number={3},
  pages={985--1000},
  year={2022},
  publisher={IEEE}
}

@article{krishnapriyan2021characterizing,
  title={Characterizing possible failure modes in physics-informed neural networks},
  author={Krishnapriyan, Aditi and Gholami, Amir and Zhe, Shandian and Kirby, Robert and Mahoney, Michael W},
  journal={Advances in neural information processing systems},
  volume={34},
  pages={26548--26560},
  year={2021}
}

@article{han2022predicting,
  title={Predicting physics in mesh-reduced space with temporal attention},
  author={Han, Xu and Gao, Han and Pfaff, Tobias and Wang, Jian-Xun and Liu, Li-Ping},
  journal={arXiv preprint arXiv:2201.09113},
  year={2022}
}

@article{hou2026learning,
  title={Learning Neural Operators from Partial Observations via Latent Autoregressive Modeling},
  author={Hou, Jingren and Wang, Hong and Xu, Pengyu and Gao, Chang and Liu, Huafeng and Jing, Liping},
  journal={arXiv preprint arXiv:2601.15547},
  year={2026}
}

@article{wang2022respecting,
  title={Respecting causality is all you need for training physics-informed neural networks},
  author={Wang, Sifan and Sankaran, Shyam and Perdikaris, Paris},
  journal={arXiv preprint arXiv:2203.07404},
  year={2022}
}

@article{wang2023expert,
  title={An expert's guide to training physics-informed neural networks},
  author={Wang, Sifan and Sankaran, Shyam and Wang, Hanwen and Perdikaris, Paris},
  journal={arXiv preprint arXiv:2308.08468},
  year={2023}
}

@article{chiu2022can,
  title={CAN-PINN: A fast physics-informed neural network based on coupled-automatic--numerical differentiation method},
  author={Chiu, Pao-Hsiung and Wong, Jian Cheng and Ooi, Chinchun and Dao, My Ha and Ong, Yew-Soon},
  journal={Computer Methods in Applied Mechanics and Engineering},
  volume={395},
  pages={114909},
  year={2022},
  publisher={Elsevier}
}

@article{wong2021can,
  title={Can transfer neuroevolution tractably solve your differential equations?},
  author={Wong, Jian Cheng and Gupta, Abhishek and Ong, Yew-Soon},
  journal={IEEE Computational Intelligence Magazine},
  volume={16},
  number={2},
  pages={14--30},
  year={2021},
  publisher={IEEE}
}

@article{penwarden2023metalearning,
  title={A metalearning approach for physics-informed neural networks (PINNs): Application to parameterized PDEs},
  author={Penwarden, Michael and Zhe, Shandian and Narayan, Akil and Kirby, Robert M},
  journal={Journal of Computational Physics},
  volume={477},
  pages={111912},
  year={2023},
  publisher={Elsevier}
}

@article{liu2022novel,
  title={A novel meta-learning initialization method for physics-informed neural networks},
  author={Liu, Xu and Zhang, Xiaoya and Peng, Wei and Zhou, Weien and Yao, Wen},
  journal={Neural Computing and Applications},
  volume={34},
  number={17},
  pages={14511--14534},
  year={2022},
  publisher={Springer}
}

@article{cho2023hypernetwork,
  title={Hypernetwork-based meta-learning for low-rank physics-informed neural networks},
  author={Cho, Woojin and Lee, Kookjin and Rim, Donsub and Park, Noseong},
  journal={Advances in Neural Information Processing Systems},
  volume={36},
  pages={11219--11231},
  year={2023}
}

@article{cho2024parameterized,
  title={Parameterized physics-informed neural networks for parameterized PDEs},
  author={Cho, Woojin and Jo, Minju and Lim, Haksoo and Lee, Kookjin and Lee, Dongeun and Hong, Sanghyun and Park, Noseong},
  journal={arXiv preprint arXiv:2408.09446},
  year={2024}
}

@article{huang2022meta,
  title={Meta-auto-decoder for solving parametric partial differential equations},
  author={Huang, Xiang and Ye, Zhanhong and Liu, Hongsheng and Ji, Shi and Wang, Zidong and Yang, Kang and Li, Yang and Wang, Min and Chu, Haotian and Yu, Fan and others},
  journal={Advances in Neural Information Processing Systems},
  volume={35},
  pages={23426--23438},
  year={2022}
}

@article{boudec2025learning,
  title={Learning a neural solver for parametric PDE to enhance physics-informed methods},
  author={Boudec, Lise Le and De B{\'e}zenac, Emmanuel and Serrano, Louis and Regueiro-Espino, Ramon Daniel and Yin, Yuan and Gallinari, Patrick},
  booktitle = {The Thirteenth International Conference on Learning Representations},
  year      = {2025},
  url       = {https://openreview.net/forum?id=jqVj8vCQsT}
}

@article{kumar2025synergistic,
  title={Synergistic learning with multi-task DeepONet for efficient PDE problem solving},
  author={Kumar, Varun and Goswami, Somdatta and Kontolati, Katiana and Shields, Michael D and Karniadakis, George Em},
  journal={Neural Networks},
  volume={184},
  pages={107113},
  year={2025},
  publisher={Elsevier}
}

@article{wei2025efficient,
  title={Efficient Transformer-Inspired Variants of Physics-Informed Deep Operator Networks},
  author={Wei, Zhi-Feng and Chen, Wenqian and Stinis, Panos},
  journal={arXiv preprint arXiv:2509.01679},
  year={2025}
}

@incollection{goswami2023physics,
  title={Physics-informed deep neural operator networks},
  author={Goswami, Somdatta and Bora, Aniruddha and Yu, Yue and Karniadakis, George Em},
  booktitle={Machine learning in modeling and simulation: methods and applications},
  pages={219--254},
  year={2023},
  publisher={Springer}
}

@article{bradbury2018jax,
  title={JAX: composable transformations of Python+ NumPy programs},
  author={Bradbury, James and Frostig, Roy and Hawkins, Peter and Johnson, Matthew James and Leary, Chris and Maclaurin, Dougal and Necula, George and Paszke, Adam and VanderPlas, Jake and Wanderman-Milne, Skye and others},
  year={2018}
}

@article{wong2025evolutionary,
  title={Evolutionary optimization of physics-informed neural networks: Evo-pinn frontiers and opportunities},
  author={Wong, Jian Cheng and Gupta, Abhishek and Ooi, Chin Chun and Chiu, Pao-Hsiung and Liu, Jiao and Ong, Yew-Soon},
  journal={IEEE Computational Intelligence Magazine},
  volume={21},
  number={1},
  pages={16--36},
  year={2026},
  publisher={IEEE}
}

@article{wang2025gradient,
  title={Gradient alignment in physics-informed neural networks: A second-order optimization perspective},
  author={Wang, Sifan and Bhartari, Ananyae Kumar and Li, Bowen and Perdikaris, Paris},
  journal={arXiv preprint arXiv:2502.00604},
  year={2025}
}

@book{pletcher2012computational,
  title={Computational fluid mechanics and heat transfer},
  author={Pletcher, Richard H and Tannehill, John C and Anderson, Dale},
  year={2012},
  publisher={CRC press}
}

@article{chiu2008effective,
  title={An effective explicit pressure gradient scheme implemented in the two-level non-staggered grids for incompressible Navier--Stokes equations},
  author={Chiu, Pao-Hsiung and Sheu, Tony WH and Lin, Reui-Kuo},
  journal={Journal of Computational Physics},
  volume={227},
  number={8},
  pages={4018--4037},
  year={2008},
  publisher={Elsevier}
}

@article{sheu2005high,
  title={On a high-order Newton linearization method for solving the incompressible Navier--Stokes equations},
  author={Sheu, Tony WH and Lin, RK},
  journal={International journal for numerical methods in engineering},
  volume={62},
  number={11},
  pages={1559--1578},
  year={2005},
  publisher={Wiley Online Library}
}

@misc{driscoll2014chebfun,
  title={Chebfun guide},
  author={Driscoll, Tobin A and Hale, Nicholas and Trefethen, Lloyd N},
  year={2014},
  publisher={Pafnuty Publications, Oxford}
}

@article{cox2002exponential,
  title={Exponential time differencing for stiff systems},
  author={Cox, Steven M and Matthews, Paul C},
  journal={Journal of Computational Physics},
  volume={176},
  number={2},
  pages={430--455},
  year={2002},
  publisher={Elsevier}
}

@article{wong2026evolutionary,
  title={Evolutionary optimization of physics-informed neural networks: Advancing generalizability by the Baldwin effect},
  author={Wong, Jian Cheng and Ooi, Chin Chun and Gupta, Abhishek and Chiu, Pao-Hsiung and Low, Joshua Shao Zheng and Dao, My Ha and Ong, Yew-Soon},
  journal={IEEE Transactions on Evolutionary Computation},
  year={2026},
  publisher={IEEE}
}

@INPROCEEDINGS{wong2026transferable,
  author={Wong, Jian Cheng and Lai, Isaac Yin Chung and Chiu, Pao-Hsiung and Ooi, Chin Chun and Gupta, Abhishek and Ong, Yew-Soon},
  booktitle={2026 International Joint Conference on Neural Networks (IJCNN)}, 
  title={Transferable Physics-Informed Representations via Closed-Form Head Adaptation}, 
  year={2026},
  volume={},
  number={},
  pages={1-8},
  doi={10.1109/IJCNN69854.2026.11729888}}

@book{dennis1996numerical,
  title={Numerical methods for unconstrained optimization and nonlinear equations},
  author={Dennis Jr, John E and Schnabel, Robert B},
  year={1996},
  publisher={SIAM}
}

@article{sheu2004newton,
  title={Newton linearization of the incompressible Navier--Stokes equations},
  author={Sheu, Tony WH and Lin, RK},
  journal={International Journal for Numerical Methods in Fluids},
  volume={44},
  number={3},
  pages={297--312},
  year={2004},
  publisher={Wiley Online Library}
}

@article{chiu2021development,
  title={Development of an improved divergence-free-condition compensated coupled framework to solve flow problems with time-varying geometries},
  author={Chiu, Pao-Hsiung and Poh, Hee Joo},
  journal={International Journal for Numerical Methods in Fluids},
  volume={93},
  number={1},
  pages={44--70},
  year={2021},
  publisher={Wiley Online Library}
}
}

\newpage
\appendix

\section*{Appendix}

In this Appendix, we provide comprehensive materials to facilitate a deeper understanding of our study. 

\begin{itemize}

    \item Appendix~\ref{app:Extended review of related work} provides an extended review of the literature.
    
    \item Appendix~\ref{app:newton-pinet} presents a detailed description of the Newton-PINet model, including the meta-learning and gradient-free fine-tuning framework, and the Newton linearization method. 

    \item Appendix~\ref{app:Convergence analysis} provides theoretical and experimental evidence showing that Newton linearization used in the Tikhonov-regularized PINN achieves near-quadratic convergence.

    \item Appendix~\ref{app:Datasets_and_experiment} describes the data generation, computational configurations, and additional experimental results. 

    \item Appendix~\ref{app:summary of model comparations} provides a comprehensive comparison of model performance on public datasets and newly introduced nonlinear benchmark problems.
    
    \item Appendix~\ref{app:Ablation_studies} provides ablation studies covering meta-learning hyperparameters, factors influencing inference cost, model architecture design, and related aspects.

    \item Appendix~\ref{app:Practical scaling discussion of Newton-PINet} presents an analysis of the computational complexity of the model and discusses its limitations and failure modes.

\end{itemize}

\section{Extended review of related work}
\label{app:Extended review of related work}

In this section, we provide an extended review of the literature and clarify its relation to our work.

\begin{itemize}

\item 
Existing PINN architectures, such as DPM and PirateNets~\citep{kim2021dpm,wang2024piratenets}, employ standard residual connections (e.g., $x + f(x)$) to improve gradient flow and stabilize training. In contrast, our approach adopts a skip-connected design that concatenates features from all hidden layers and feeds them into the output layer, followed by a least-squares solve. This design aims to construct a richer basis space, where each hidden layer progressively contributes additional nonlinear features, thereby enhancing representation capacity and improving the conditioning and accuracy of the physics-informed Tikhonov solution, rather than merely stabilizing training.

\item 
Least-squares Gradient Descent (LSGD)~\citep{cyr2020robust} combines least-squares formulations with gradient-based optimization, focusing on data-driven single-task learning with linear structures. In contrast, our method extends least-squares formulations to meta-learning for nonlinear PDEs through a physics-informed Tikhonov framework. Specifically, the system is constructed from physical constraints (PDE residuals, ICs, and BCs), and Newton linearization transforms nonlinear residuals into local linear least-squares problems. This enables efficient gradient-free inner-loop adaptation with inherent enforcement of physical laws, which is absent in LSGD.

\item 
Methods such as Almost No Inner Loop (ANIL)~\citep{raghu2019rapid} improve efficiency by restricting adaptation to the last layer. While sharing this idea, these approaches rely on gradient-based updates. In contrast, our method performs closed-form, physics-constrained least-squares updates via Tikhonov regularization. By eliminating inner-loop gradient descent and exploiting the algebraic structure of PDEs, our approach enables fast and stable adaptation to new tasks, fundamentally differing from conventional data-driven meta-learning methods.

\end{itemize}

Solving complex nonlinear PDEs via meta-learning presents fundamental challenges that require entirely different algorithmic considerations from traditional data-driven methods like LSGD or ANIL. By synergistically integrating basis-enriching skip-connections, Newton linearization, and physics-constrained Tikhonov least-squares, our work proposes a distinct physics-informed meta-learning framework tailored to the algebraic structure of physical systems, which is firmly backed by both theoretical and empirical analyses.

\section{Newton-PINet model}
\label{app:newton-pinet}

This study presents a skip-connected Tikhonov regularization PINN model for meta-learning nonlinear PDEs, enhanced with Newton linearization, to enable efficient few-shot learning and fast generalization, hereafter referred to as Newton-PINet.

\subsection{Skip-connected PINN architecture}
\label{app:Skip-connected PINN architecture}

Building upon the conventional feed-forward MLP architecture of PINNs, we introduce skip connections from all hidden layers to the output layer, with the output-layer weights computed using Tikhonov regularization. 

The skip-connected neural architecture enables us to provide better expressivity to ensure stable and accurate least-squares computation by increasing the width of the output layer through the stacking of additional hidden layers, even while maintaining a moderate number of nodes per layer. In addition, the concatenation mechanism can be interpreted as expanding a richer basis space: each added hidden layer introduces new nonlinear features, analogous to incorporating higher-order terms in a polynomial or Chebyshev-type basis, leading to a more expressive representation with reduced truncation effects. A complementary viewpoint is that skip connections also help stabilize optimization—much like in ResNets—by mitigating vanishing-gradient issues in deeper architectures. 

This architecture resembles extreme learning machines (ELMs) in its stacked structure \citep{dong2021local}. However, in traditional ELMs, hidden-layer weights are randomly initialized, which makes training deeper architectures increasingly difficult and can compromise the accuracy of the solution. In contrast, our framework employs a meta-learning strategy to update the hidden-layer weights instead of relying on random initialization. As a result, the network depth can be increased appropriately to improve nonlinear representation capacity while maintaining stable and accurate generalization.

We employ the Tikhonov-regularized PINN with $L+1$ layers, where the input is $\mathbf{x} \in \mathbb{R}^{D_{in}}$ (layer $0$) and the output is $\mathbf{u} \in \mathbb{R}^{D_o}$ (layer $L$). Here, $D_{in}$ and $D_o$ denote the dimensionality of the input and output variables, respectively. Each hidden layer $l=1,\dots,L-1$ contains the same number of neurons ($N_n$). Sinusoidal feature embeddings are applied at the first layer~\citep{wong2022learning}, and all hidden layers use $\sin(\cdot)$ activation functions. Skip connections from all hidden layers to the output layer are employed to allow each hidden layer to contribute nonlinearly to the final representation without additional weights in the concatenation.

The transformation and activation in the first hidden layer are defined as
\begin{equation}
    z^1_j = \sum_{d=1}^{D_{in}} W^1_{j,d} x_d + b^1_j, 
    \quad 
    f^1_j(\mathbf{x}) = \sin(2\pi z^1_j), 
    \quad j=1,\dots,N_n,
\end{equation}
where $j$ and $d$ denote neurons in the current and previous layers respectively, $W^1 \in \mathbb{R}^{N_n \times D_{in}}$, and $b^1 \in \mathbb{R}^{N_n}$. We incorporate Sinusoidal feature embeddings in the first layer to enhance the representation of high-frequency components.

For subsequent hidden layers $l=2,\dots,L-1$, we have
\begin{equation}
    z^l_j = \sum_{d=1}^{N_n} W^l_{j,d} f^{l-1}_d + b^l_j, 
    \quad 
    f^l_j(\mathbf{x}) = \sin(z^l_j), 
    \quad j=1,\dots,N_n,
\end{equation}
where $W^l \in \mathbb{R}^{N_n \times N_n}$ and $b^l \in \mathbb{R}^{N_n}$.  

All trainable network parameters up to the pre-final layer are denoted as $\boldsymbol{\tilde{w}}=[W^l,b^l], \; l=1,\dots, L-1$.

The pre-final output is constructed as the concatenation of all hidden activations (skip connections):
\begin{equation}
    \boldsymbol{f}(\mathbf{x};\boldsymbol{\tilde{w}})^T = 
    \big[ f^1_1, \dots, f^1_{N_n},\;
           f^2_1, \dots, f^2_{N_n},\;
           \dots,\;
           f^{L-1}_1, \dots, f^{L-1}_{N_n} \big],
\end{equation}
which is then connected to the output layer with $D_o$ dimension, so that $\boldsymbol{f}(\mathbf{x};\boldsymbol{\tilde{w}})^T \in \mathbb{R}^{ D_o\times{N_n}(L-1)}$. Then, we need to flatten it, resulting in $\boldsymbol{f}(\mathbf{x};\boldsymbol{\tilde{w}})^T \in \mathbb{R}^{1 \times D_o{N_n}(L-1)}$.

The output $u$ can be computed by
\begin{equation}
    u(\mathbf{x}) = \sum_{j} w_j f_j
    = \boldsymbol{f}(\mathbf{x};\boldsymbol{\tilde{w}})^T\boldsymbol{w},
\end{equation}
where $\boldsymbol{w}^T = [\ldots w_j \ldots] \in \mathbb{R}^{1 \times D_oN_n(L-1)}$ denotes the output-layer weights.

\subsection{Physics-based least-squares formulation}
\label{app:Physics-based least-squares formulation}

The objective of task-specific learning is to determine the best set of $\boldsymbol{w}$ such that $u(\mathbf{x})=\boldsymbol{f}(\mathbf{x};\boldsymbol{\tilde{w}})^T\boldsymbol{w}$ satisfies the PDE, IC, and BC, for a task ($\mathcal{T}_i$):
\begin{subequations}
    \label{eq:general_pde}
    \begin{alignat}{3}
    \text{PDE:} \quad & \mathcal{N_{\theta}}[u(x,t) ] = q(x,t) 
    && \quad x \in \Omega, \; t \in [0,T], \label{eq:pde_1a}\\
    \text{IC:} \quad & u(x,t=0) = u_0(x) 
    && \quad x \in \Omega, \label{eq:pde_1b}\\
    \text{BC:} \quad & \mathcal{B}[u(x,t)] = g(x,t) 
    && \quad x \in \partial \Omega, \; t \in [0,T], \label{eq:pde_1c}
    \end{alignat} \label{eq:pde__all}
\end{subequations}
where $\mathcal{N}_{\theta}, q, u_0, g$ denote the differential operator with PDE parameters $\theta$, the source term, IC, and BC, respectively.

This leads to a physics-based least-squares formulation. Let $\mathbf{x} = (x, t)$. Given collocation points for the PDE residual, $(x_n^{\mathrm{pde}}, t_n^{\mathrm{pde}}),~n=1,\dots,n_{\mathrm{pde}}$; points at the initial time, $(x_n^{\mathrm{ic}}, 0),~n=1,\dots,n_{\mathrm{ic}}$; and points on the boundary, $(x_n^{\mathrm{bc}}, t_n^{\mathrm{bc}}),~n=1,\dots,n_{\mathrm{bc}}$, where $n_{\mathrm{pde}}$, $n_{\mathrm{ic}}$, and $n_{\mathrm{bc}}$ denote the numbers of the corresponding points, respectively. The loss importance hyperparameter $\lambda_{\mathrm{pde}}$ is further introduced, and the following system is obtained:

\begin{equation}
    \begin{bmatrix}
    \cdots &  \lambda_{\mathrm{pde}} \cdot \mathcal{N}_{\theta}[f_j(x^{\mathrm{pde}}_1, t^{\mathrm{pde}}_1;\boldsymbol{\tilde{w}})] & \cdots \\[-2pt]
     & \vdots  \\[-2pt]
    \cdots & \lambda_{\mathrm{pde}} \cdot\mathcal{N}_{\theta}[f_j(x^{\mathrm{pde}}_{n_{\mathrm{pde}}}, t^{\mathrm{pde}}_{n_{\mathrm{pde}}};\boldsymbol{\tilde{w}})] & \cdots \\[4pt]
    \cdots & f^i_j(x^{\mathrm{ic}}_1, 0;\boldsymbol{\tilde{w}}) & \cdots \\[-2pt]
    & \vdots  \\[-2pt]
    \cdots & f_j(x^{\mathrm{ic}}_{n_{\mathrm{ic}}}, 0;\boldsymbol{\tilde{w}}) & \cdots \\[4pt]
    \cdots & \mathcal{B}[f_j(x^{\mathrm{bc}}_1, t^{\mathrm{bc}}_1;\boldsymbol{\tilde{w}})] & \cdots \\[-2pt]
    & \vdots  \\[-2pt]
    \cdots & \mathcal{B}[f_j(x^{\mathrm{bc}}_{n_{\mathrm{bc}}}, t^{\mathrm{bc}}_{n_{\mathrm{bc}}};\boldsymbol{\tilde{w}})] & \cdots
    \end{bmatrix}
    \begin{bmatrix}
    \vdots \\[-2pt]
    w_j \\[-2pt]
    \vdots
    \end{bmatrix}
    =
    \begin{bmatrix}
    \lambda_{\mathrm{pde}} \cdot q(x^{\mathrm{pde}}_1, t^{\mathrm{pde}}_1) \\[-2pt]
    \vdots \\[-2pt]
    \lambda_{\mathrm{pde}} \cdot q(x^{\mathrm{pde}}_{n_{\mathrm{pde}}}, t^{\mathrm{pde}}_{n_{\mathrm{pde}}}) \\[4pt]
    u_0(x^{\mathrm{ic}}_1) \\[-2pt]
    \vdots \\[-2pt]
    u_0(x^{\mathrm{ic}}_{n_{\mathrm{ic}}}) \\[4pt]
    g(x^{\mathrm{bc}}_1, t^{\mathrm{bc}}_1) \\[-2pt]
    \vdots \\[-2pt]
    g(x^{\mathrm{bc}}_{n_{\mathrm{bc}}}, t^{\mathrm{bc}}_{n_{\mathrm{bc}}})
    \end{bmatrix},
\label{eq:L-S matrix}
\end{equation}

which can be compactly written as
\[\mathbf{A}\boldsymbol{w} = \mathbf{b}.\]

Here $\mathbf{A} \boldsymbol{w}$ represents the left-hand side of Eq.~(\ref{eq:pde__all}) at a given set of collocation points, $\mathbf{b}$ denotes the corresponding right-hand side of Eq.~(\ref{eq:pde__all}). 

Therefore, using the PDE, IC, and BC, the physics-informed matrix $\mathbf{A}$ and vector $\mathbf{b}$ can be assembled.
Concretely, the pre-final representation $\boldsymbol{f}$ is obtained via forward propagation, and its derivatives with respect to $(x,t)$ in $\mathcal{N}_{\theta}$ are computed through automatic differentiation (AD). 

For a single task, i.e., a single PDE instance, the number of rows in $\mathbf{A}$ and $\mathbf{b}$ equals the total number of collocation points, including those for the PDE residual, IC, and BC. The number of columns in $\mathbf{A}$ corresponds to the product of the pre-final feature dimension $N_n(L-1)$ and the output dimension $D_o$. This results in $\mathbf{A}\in \mathbb{R}^{(n_{\text{pde}}+n_\text{ic}+n_\text{bc}) \times D_oN_n(L-1)}$ and $\mathbf{b}\in \mathbb{R}^{(n_{\text{pde}}+n_\text{ic}+n_\text{bc}) \times 1}$. 

The final output weights $\boldsymbol{w}$ are obtained via Tikhonov regularization:
\begin{equation}
    \arg\min_{\boldsymbol{w}} \; 
    \left\lVert \mathbf{A}\boldsymbol{w} - \mathbf{b} \right\rVert_2^2
    + \lambda_{\text{reg}} \, \boldsymbol{w}^T \boldsymbol{w},
\end{equation}
where $\lambda_{\text{reg}} \ge 0$ denotes the regularization parameter.
Compared with ordinary least squares (OLS), which minimizes only the residual norm $\|\mathbf{A}\boldsymbol{w} - \mathbf{b}\|_2^2$,  
Tikhonov regularization introduces an additional penalty term $\lambda_{\text{reg}} \, \boldsymbol{w}^T \boldsymbol{w}$. This term suppresses large weights, stabilizes the solution, and improves the conditioning of the problem. 

Note that for linear PDE, the closed-form Tikhonov-regularized solution of Eq.~(\ref{eq:L-S matrix}) can be obtained in a single step:
\begin{equation}
    \boldsymbol{w}=
    \begin{cases}
    (\lambda_{\text{reg}} \mathbf{I} + \mathbf{A}^{T}\mathbf{A})^{-1}\mathbf{A}^{T}\mathbf{b}, 
    & \text{if Eq.~(\ref{eq:L-S matrix}) is over-determined}, \\[6pt]
    \mathbf{A}^{T}(\lambda_{\text{reg}} \mathbf{I} + \mathbf{A}\mathbf{A}^{T})^{-1}\mathbf{b}, 
    & \text{if Eq.~(\ref{eq:L-S matrix}) is under-determined}.
    \end{cases}
    \label{eq:Tikhonov-regularized solution}
\end{equation}
For nonlinear PDEs, the nonlinear terms must first be linearized so that the system can be cast in the form of Eq.~(\ref{eq:L-S matrix}). For example, in the Burgers' equation: $ u_t+uu_x-\gamma u_{xx}=0,$
we can compute
\[u = \boldsymbol{f}^T \boldsymbol{w}, \quad
u_t = \frac{\partial u}{\partial t} = \boldsymbol{f}_t^T \boldsymbol{w}, \quad
u_x = \frac{\partial u}{\partial x} = \boldsymbol{f}_x^T \boldsymbol{w}, \quad
u_{xx} = \frac{\partial^2 u}{\partial x^2} = \boldsymbol{f}_{xx}^T \boldsymbol{w},
\]
where the pre-final representation $\boldsymbol{f}$ and its derivatives can be obtained via forward propagation and AD. However, the nonlinear term $u u_x$ cannot be directly expressed as a combination of $\boldsymbol{f}$ (or $\boldsymbol{f}_x$) and $\boldsymbol{w}$, which motivates the need for linearization. Under Newton linearization, the nonlinear term can be approximated as
\[
(uu_x)^{k+1} \approx u^{k+1} u_x^k + u^k u_x^{k+1} - u^k u_x^k,
\]
where \(k\) and \(k+1\) denote the current and next nonlinear iteration step, respectively. Given initial guesses for $u$ and $u_x$ (e.g., setting $u$ and $u_x$ at all time steps equal to the initial field), the corresponding entries in $\mathbf{A}$ and $\mathbf{b}$ are updated as:
\begin{align}
    \mathcal N_\theta \!\left[ f^{k+1}_j(\mathbf{x}^\text{{pde}}_n;\boldsymbol{\tilde{w}}) \right] 
    &= \frac{\partial f^{k+1}_j (\mathbf{x}^\text{pde}_n;\boldsymbol{\tilde{w}})}{\partial t} 
    + u^{k}(\mathbf{x}^\text{pde}_n)\frac{\partial f^{k+1}_j (\mathbf{x}^\text{pde}_n;\boldsymbol{\tilde{w}})}{\partial x} \nonumber \\
    &\quad + f^{k+1}_j (\mathbf{x}^\text{pde}_n;\boldsymbol{\tilde{w}})\frac{\partial{u}^{k}(\mathbf{x}^\text{pde}_n)}{\partial x} - \gamma \frac{\partial^2 f^{k+1}_j (\mathbf{x}^\text{pde}_n;\boldsymbol{\tilde{w}})}{\partial x^2},
\end{align}
\begin{align}
    q(\mathbf{x}^\text{pde}_n) = u^{k}(\mathbf{x}^\text{pde}_n)\frac{\partial{u}^{k}(\mathbf{x}^\text{pde}_n)}{\partial x}.
\end{align}
This leads to an iterative nonlinear solve, where Eq.~(\ref{eq:Tikhonov-regularized solution}) is repeatedly solved to update $\boldsymbol{w}$ toward the optimal solution.

\paragraph{How to make Tikhonov solver tractable?}

The task adaptation is formulated as a Tikhonov-regularized least-squares problem, whose tractability mainly depends on the size of $\mathbf{A}$, the Tikhonov regularization parameter ($\lambda_{\text{reg}}$), the PDE-loss importance hyperparameter ($\lambda_{\text{pde}}$), and the choice of the initial field during nonlinear iterations.

\begin{itemize}

    \item The matrix $\mathbf{A}$ is constructed from the network features at each sampling point of the PDE residuals/ICs/BCs, which are generated by a multilayer neural representation with skip connections. The distribution of these features directly determines the conditioning of the solve. In addition, when the system requires a large number of sampling points, the number of rows in $\mathbf{A}$ increases, making it more difficult to solve. Since the Tikhonov update only involves $\mathbf{A}^T\mathbf{A} \in \mathbb{R}^{D_o N_n(L-1)\times D_o N_n(L-1)}$, the computational cost mainly depends on the network feature dimension. Thus, the practical scaling for large collocation sets may derive more from the fact that one can still control the cost of the solve through meta-learning a better (yet minimal) set of pre-final layer features.

    \item To avoid laborious hyperparameter tuning, both $\lambda_{\text{reg}}$ and $\lambda_{\text{pde}}$ are meta-learned jointly with the network parameters.

    \item For nonlinear problems, an inappropriate initial field may slow convergence. Therefore, we use the initial condition as $u_{\mathrm{guess}}$ (setting $u_{\mathrm{guess}}$ at all time steps equal to the initial field) for time-dependent problems and employ temporal domain decomposition for more complex dynamics, ensuring that each block starts closer to the true solution, which improves the stability of the least-squares updates.
\end{itemize}

The specific discussion on the scalability of the model is provided in Appendix~\ref{app:Practical scaling discussion of Newton-PINet}.

\subsection{Meta-learning PINN on training tasks}
\label{app:Meta-learning PINN on training tasks}

The trainable hyperparameters during meta-learning include network weights ($\boldsymbol{\tilde{w}}$), PDE loss importance hyperparameter ($\lambda_{\text{pde}}$), and regularization parameter ($\lambda_{\text{reg}}$), collectively denoted as $\Theta$. These components are crucial for achieving optimal performance in task-specific physics-informed learning. 
 
The meta-learning objective is to optimize $\Theta$ to learn task-agnostic representations spanning a family of PDEs, enabling fast generalization to unseen tasks via fine-tuning only the output-layer weights using the Tikhonov regularization update. In our framework, the outer loop of meta-learning updates the learnable parameters via gradient descent, while task-specific adaptation in the inner loop, or generalization, is performed through gradient-free Tikhonov regularization applied to the output-layer weights. See Algorithm~\ref{alg:meta-learning} for the pseudocode workflow.

We consider a distribution over tasks $p(\mathcal{T})$ that we want our model to be able to adapt to. The meta-objective (loss) may consist solely of physics-based least-squares error ($l_\text{LSE}$, unsupervised), data-driven mean squared error ($l_\text{MSE}$, few-shot training), or a combination of both:
\begin{equation}
    \min_{\Theta} \, 
    \mathbb{E}_{\mathcal{T}_i \sim p(\mathcal{T})} 
    \mathbb{E}_{\boldsymbol{\tilde{w}} \sim p_\theta(\boldsymbol{\tilde{w}})} 
    \big[l_\text{LSE}(\boldsymbol{w}^*) + l_\text{MSE}(\boldsymbol{w}^*) \big].
\end{equation}
Here, Tikhonov regularization is used to obtain the optimal output-layer weights $\boldsymbol{w}^*$ that allow the model to specialize to any realization of task $\mathcal{T}_i \sim p(\mathcal{T})$ for the given network’s $\boldsymbol{\tilde{w}} \sim p_\theta(\boldsymbol{\tilde{w}})$.

$l_\text{LSE}$ and $l_\text{MSE}$ are
\begin{equation}
    l_\text{LSE}(\boldsymbol{w}^*) =  (\mathbf{A}\boldsymbol{w^*} - \mathbf{b})^{T}
    ( \mathbf{A} \boldsymbol{w^*} - \mathbf{b}),
\label{eq:loss_lse}
\end{equation}
\begin{equation}
    l_\text{MSE}(\boldsymbol{w}^*) = \frac{1}{n} \sum_{s=1}^{n} 
    ( u_s^{\text{label}} - \sum_{j} w^*_j \, f_j(x_s,t_s;\boldsymbol{\tilde{w}}) )^2,
\label{eq:loss_mse}
\end{equation}
given labeled data $\{u_s^{\text{label}}\}_{s=1}^n$ ($n$ denotes the total number of collocation points including PDE residual, IC, and BC constraints for a task).

We refer to $l_\text{LSE}$-based learning as \textbf{unsupervised learning}, and to $l_\text{MSE}$-based learning as \textbf{hybrid learning} (a combination of physics-based Tikhonov updates and data-driven loss). Unless otherwise stated, for meta-learning loss (outer-loop), our model does not consider the combined loss formulation $l_\text{LSE} + l_\text{MSE}$, as in some cases the combined objective exhibits poorer generalization performance compared with using $l_\text{MSE}$ alone.

In practice, we observe that for simple nonlinear PDEs (e.g., Burgers' equation), training with $l_\text{LSE}$ alone already leads to strong generalization. However, for more complex PDEs, incorporating a small amount of labeled data through the $l_\text{MSE}$ term becomes essential for both stability and accuracy.

\subsection{Gradient-free fine-tuning to new tasks} 
\label{app:Gradient-free fine-tuning to new tasks}

After meta-learning, the network weights ($\boldsymbol{\tilde{w}}$), the PDE-loss importance hyperparameter ($\lambda_{\text{pde}}$), and the regularization parameter ($\lambda_{\text{reg}}$) are fixed. For a new task with different PDE parameters or IC/BC conditions, Tikhonov regularization is used to compute the output-layer weights ($\boldsymbol{w}$), enabling fast gradient-free adaptation to the new PDE instance. See Algorithm~\ref{alg:inference} for the pseudocode workflow. Since this update is independent of gradient-based backpropagation optimizers such as stochastic gradient descent (SGD), the resulting adaptation is extremely fast while remaining physics-compliant.

\subsection{Linearization for nonlinear PDEs}
\label{app:Linearization for nonlinear PDEs}

It should be noted that the Tikhonov regularization is directly applicable to linear PDEs, where a linear matrix system can be constructed and solved in a single step. For nonlinear PDEs, the nonlinear terms must first be linearized so that the system can be cast into a linear form. The previous approaches typically relied on Picard method (lagging-of-coefficients), which linearizes the nonlinear terms using the solution from the previous iteration and then performs least-squares solves for multiple nonlinear iterations. 

\textbf{\textit{Picard iterations.}} Take the Burgers' equation as an example: $ u_t+uu_x-\gamma u_{xx}=0$. The Picard method approximately linearizes the nonlinear term as: $
(uu_x)^{k+1}=u^ku_x^{k+1},$ where \(k\) and \(k+1\) denote the current and next iteration steps, respectively. An initial guess for $u$ is provided, e.g., setting $u$ at all time steps equal to the initial field. Under this formulation, the entries of the least-squares matrix $\mathbf{A}$ and vector $\mathbf{b}$ in Eq.~(\ref{eq:L-S matrix}) corresponding to the $(k+1)$-th iteration, $n$-th PDE collocation point, and the $j$-th output-layer neuron are given by:
\begin{align}
    \mathcal N_\theta \!\left[ f^{k+1}_j (\mathbf{x}^\text{pde}_n;\boldsymbol{\tilde{w}}) \right] 
    &= \frac{\partial f^{k+1}_j (\mathbf{x}^\text{pde}_n;\boldsymbol{\tilde{w}})}{\partial t} 
    +  u^{k}(\mathbf{x}^\text{pde}_n)\frac{\partial f^{k+1}_j (\mathbf{x}^\text{pde}_n;\boldsymbol{\tilde{w}})}{\partial x} \nonumber \\
    &\quad - \gamma\frac{\partial^2 f^{k+1}_j (\mathbf{x}^\text{pde}_n;\boldsymbol{\tilde{w}})}{\partial x^2}
    \label{eq:burgers-lagg-A}
\end{align}
\begin{align} 
    q(\mathbf{x}^{\text{pde}}_n) = 0,
    \label{eq:burgers-lagg-b}
\end{align}
where $f_j$ denotes the pre-final output of the neural network. Note that $u^{k}$ here is the known solution, either the initial guess or the solution at the $k$-th iteration.

The final output weights $\boldsymbol{w}^T = [\ldots, w_j, \ldots]$ are obtained via Tikhonov regularization-based nonlinear iterations. At each iteration, the solution is updated by $u(\mathbf{x}) = \sum_{j} w_j f_j$, which then serves as the new guess for the next iteration. This procedure is repeated until $\boldsymbol{w}$ satisfies a prescribed convergence criterion or the maximum number of iterations is reached.

\textit{\textbf{Newton linearization iterations.}} The Picard approach suffers from slow convergence of nonlinear PDE solutions due to its linear convergence speed. To overcome these limitations, we use the Newton linearization approach. 

Let \(F(u^{k+1})\) denote the nonlinear term as a function of the state variable \(u\). In Newton linearization, a first-order Taylor expansion is performed around the current iterate \(u^k\), yielding
\begin{eqnarray} F(u^{k+1}) &= F(u^k)+\left. \frac{d F}{d u} \right|^k (u^{k+1} - u^k)+\text{H.O.T} \label{eq:taylor expansion-one order}, \end{eqnarray} where ``$\text{H.O.T}$'' denotes the higher-order truncated terms.

According to Eq.~(\ref{eq:taylor expansion-one order}), we can derive a general expression for the nonlinear terms $u^m$ with $m$ denoting the exponent that commonly appear in PDEs, as follows:
\begin{eqnarray}
    (u^m)^{k+1} &\approx& (u^m)^{k} + m(u^{m-1})^{k}(u^{k+1}-u^{k})  \nonumber\\
                &=& m (u^{k})^{m-1} u^{k+1} + (1-m)(u^{k})^{m}
    \label{eq:u^m expansion}
\end{eqnarray}

The corresponding expressions for $m=2$ and $m=3$ are as follows:
\begin{eqnarray}
    (u^2)^{k+1} &\approx& 2 u^{k+1} u^k - u^k u^k 
\end{eqnarray}
\begin{eqnarray}
    (u^3)^{k+1} &\approx& 3u^{k+1} u^k u^k - 2u^k u^k u^k 
\end{eqnarray}

For nonlinear terms involving multiple state variables, i.e., $F(u^{k+1}, v^{k+1})$, the corresponding Taylor expansion is given by:
\begin{eqnarray}
    F(u^{k+1}, v^{k+1}) = F(u^k, v^k) + \left.\tfrac{\partial F}{\partial u}\right|^k (u^{k+1}-u^k) + \left.\tfrac{\partial F}{\partial v}\right|^k (v^{k+1}-v^k) + \text{H.O.T}
    \label{eq:taylor expansion-two order}
\end{eqnarray}

Based on Eq.~(\ref{eq:taylor expansion-two order}), the general Newton linearization form for nonlinear convective terms of arbitrary order can be expressed as:
\begin{eqnarray}
    (u^m u_x)^{k+1} &\approx& (u^m) ^k u_x^k + u_x^k\left[(u^m) ^{k+1}-(u^m) ^k \right]+ (u^m) ^k(u_x ^{k+1}-u_x^k)\nonumber
    \\&=&  (u^m) ^{k+1} u_x^k + (u^m)^k u_x^{k+1} - (u^m)^k u_x^k  \nonumber\\
    &=& \left[ m (u^k)^{m-1} u^{k+1} + (1 - m)(u^k)^m \right] u_x^k + (u^k)^m u_x^{k+1} - (u^k)^m u_x^k \nonumber \\
    &=& m (u^k)^{m-1} u_x^k u^{k+1} + (1 - m)(u^k)^m u_x^k + (u^k)^m u_x^{k+1} - (u^k)^m u_x^k \nonumber \\
    &=& m (u^k)^{m-1} u_x^k u^{k+1} + (u^k)^m\, u_x^{k+1} - m (u^k)^m u_x^k
\end{eqnarray}
Accordingly, the corresponding expressions for the nonlinear convective terms with $m=1$, $m=2$, and $m=3$ are given by:
\begin{eqnarray}
    (uu_x)^{k+1} &\approx&  u^{k+1} u_x^k + u^k u_x^{k+1} - u^k u_x^k
    \label{eq:uux newton expansion}
\end{eqnarray}
\begin{eqnarray}
    (u^2 u_x)^{k+1} &\approx&  2 u^k u_x^k u^{k+1} + (u^k)^2\, u_x^{k+1} - 2 (u^k)^2 u_x^k
\end{eqnarray}
\begin{eqnarray}
    (u^3 u_x)^{k+1} &\approx& 3 (u^k)^{2} u_x^k u^{k+1} + (u^k)^3\, u_x^{k+1} - 3 (u^k)^3 u_x^k
\end{eqnarray}
Therefore, the nonlinear term in the Burgers' equation under Newton linearization can be expressed as 
$(uu_x)^{k+1} \approx \underline{u^{k+1} u_x^k} + u^k u_x^{k+1} - \underline{u^k u_x^k}$,
where the underlined terms originate from higher-order corrections that refine the classical Picard linearization: $(uu_x)^{k+1}=u^ku_x^{k+1}$.
Given initial guesses for $u$ and $u_x$ (e.g., setting $u$ and $u_x$ at all time steps equal to the initial field), the corresponding entries in \(\mathbf{A}\) and \(\mathbf{b}\) are updated as:  
\begin{align}
    \mathcal N_\theta \!\left[ f^{k+1}_j(\mathbf{x}^\text{pde}_n;\,\boldsymbol{\tilde{w}}) \right] 
    &= \frac{\partial f^{k+1}_j (\mathbf{x}^\text{pde}_n;\,\boldsymbol{\tilde{w}})}{\partial t} 
    +  u^{k}(\mathbf{x}^\text{pde}_n)\frac{\partial f^{k+1}_j (\mathbf{x}^\text{pde}_n;\,\boldsymbol{\tilde{w}})}{\partial x} \nonumber \\
    &\quad + f^{k+1}_j (\mathbf{x}^\text{pde}_n;\,\boldsymbol{\tilde{w}})\frac{\partial{u}^{k}(\mathbf{x}^\text{pde}_n)}{\partial x} - \gamma \frac{\partial^2 f^{k+1}_j (\mathbf{x}^\text{pde}_n;\,\boldsymbol{\tilde{w}})}{\partial x^2} 
\end{align}
\begin{align}
    q(\mathbf{x}^\text{pde}_n) = u^{k}(\mathbf{x}^\text{pde}_n)\frac{\partial{u}^{k}(\mathbf{x}^\text{pde}_n)}{\partial x}.
\end{align}
The Newton linearization method substantially accelerates the convergence of Tikhonov regularization-based nonlinear iterations due to its near-quadratic convergence property (see Appendix~\ref{app:Convergence analysis} for a proof).
Consequently, it reduces the time complexity of meta-learning and enhances the speed of fine-tuning and generalization on new tasks.

Similarly, the Newton-linearized expressions for other nonlinear terms, such as hyperbolic, exponential, and logarithmic forms, are given as follows:
\begin{eqnarray}
    [\sinh(u)]^{k+1} &\approx& \sinh(u^k) + \cosh(u^k)(u^{k+1} - u^k) \nonumber\\
    &=& \cosh(u^k) u^{k+1} + \sinh(u^k) - \cosh(u^k) u^k
\end{eqnarray}

\begin{eqnarray}
    [\cosh(u)]^{k+1} &\approx& \cosh(u^k) + \sinh(u^k)(u^{k+1} - u^k) \nonumber\\
    &=& \sinh(u^k) u^{k+1} + \cosh(u^k) - \sinh(u^k) u^k
\end{eqnarray}

\begin{eqnarray}
    [\tanh(u)]^{k+1} &\approx& \tanh(u^k) + \big(1 - \tanh^2(u^k)\big)(u^{k+1} - u^k) \nonumber\\
    &=& \big(1 - \tanh^2(u^k)\big) u^{k+1} + \tanh(u^k) \nonumber\\
    &\quad& - \big(1 - \tanh^2(u^k)\big) u^k
\end{eqnarray}

\begin{eqnarray}
    [\exp(u)]^{k+1} &\approx& \exp(u^k) + \exp(u^k)(u^{k+1} - u^k) \nonumber\\
    &=& \exp(u^k) u^{k+1} + \exp(u^k) - \exp(u^k) u^k
\end{eqnarray}

\begin{eqnarray}
    [u \ln(u)]^{k+1} &\approx& u^k \ln(u^k) + (\ln(u^k) + 1)(u^{k+1} - u^k) \nonumber\\
    &=& (\ln(u^k) + 1) u^{k+1} + u^k \ln(u^k) - (\ln(u^k) + 1)u^k
\end{eqnarray}

\section{Convergence analysis of the Newton-PINet}
\label{app:Convergence analysis}

Newton linearization has long been employed in numerical PDE solvers~\citep{sheu2004newton,sheu2005high,chiu2008effective}, e.g., to linearize the convective term in the Navier-Stokes equations. We integrate Newton linearization into the Tikhonov-regularized PINN, resulting in Newton-PINet, which significantly improves the performance of meta-learning nonlinear PDEs. This appendix provides theoretical analysis and experimental validation showing that the Newton-PINet optimization procedure is essentially equivalent to the classical regularized Gauss-Newton method and achieves near-quadratic convergence.

\subsection{Nonlinear system formulation}

The output $u$ and its spatial derivative $u_x$ at the $n$-th collocation point in the Tikhonov-regularized PINN are given by
\begin{equation}
    u(\mathbf{x}_n) = \sum_j f_j(\mathbf{x}_n) w_j,
    \quad  
    u_x(\mathbf{x}_n) = \sum_j f_{x,j}(\mathbf{x}_n) w_j 
    \quad  \Rightarrow \quad 
    u = \boldsymbol{\Phi} \boldsymbol{w}, 
    \quad 
    u_x = \boldsymbol{\Phi}_x \boldsymbol{w},
\end{equation}
where the pre-final features $\{f_j\}$ are generated by hidden-layer transformations followed by smooth activation functions, and are thus inherently bounded and continuously differentiable. Here, $\boldsymbol{w}^T = [\ldots, w_j, \ldots]$ denotes the output-layer weights, $\Phi_{nj} = f_j(\mathbf{x}_n)$, and $(\Phi_x)_{nj} = f_{x,j}(\mathbf{x}_n)$.

To provide a simple illustration, we consider a nonlinear equation $F = uu_x$ defined on the domain $x \in [0,1]$, subject to the boundary conditions $u(0) = 0$ and $u(1) = 1$. The PDE residuals evaluated at the collocation points $\{x_n\}$ are then given by:
\begin{equation}
    r_n(\boldsymbol{w}) = u(x_n)\, u_x(x_n)
    = (\boldsymbol{\Phi} \boldsymbol{w})_n \, (\boldsymbol{\Phi}_x \boldsymbol{w})_n.
\end{equation}
The residuals corresponding to the boundary conditions are
\begin{equation}
    r_0(\boldsymbol{w}) = u(0) - 0 = \boldsymbol{\Phi}(0,:)\boldsymbol{w},
    \qquad
    r_1(\boldsymbol{w}) = u(1) - 1 = \boldsymbol{\Phi}(1,:)\boldsymbol{w} - 1.
\end{equation}
We consider the nonlinear system of equations in a general form: $\mathbf{A}(\boldsymbol{w})\,\boldsymbol{w} = \mathbf{b}$.

By stacking all the residuals, we obtain the unified nonlinear system:
\begin{equation}
\begin{aligned}
F(\boldsymbol{w})
&= \mathbf{A}(\boldsymbol{w})\,\boldsymbol{w} - \mathbf{b} \\
&=
\begin{bmatrix}
u \odot u_x \\
\boldsymbol{\Phi}(0,:)\boldsymbol{w} \\
\boldsymbol{\Phi}(1,:)\boldsymbol{w} - 1
\end{bmatrix} =
\begin{bmatrix}
(\boldsymbol{\Phi}\boldsymbol{w}) \odot (\boldsymbol{\Phi}_x\boldsymbol{w}) \\
\boldsymbol{\Phi}(0,:)\boldsymbol{w} \\
\boldsymbol{\Phi}(1,:)\boldsymbol{w} - 1
\end{bmatrix}
= \mathbf{0},
\end{aligned}
\end{equation}
where $\odot$ denotes the elementwise product.

\subsection{Classical Gauss-Newton method}

In our experiments, the matrix $\mathbf{A}$ is predominantly overdetermined. For example, in the Burgers' equation, $\mathbf{A}\in\mathbb{R}^{6810\times512}$. In this setting, the Newton-PINet optimization can be characterized as the Gauss-Newton method~\citep{dennis1996numerical}. Specifically, the goal of the Gauss-Newton method is to minimize a nonlinear least-squares objective function:
\begin{equation}
\mathcal{L}(\boldsymbol{w}) = \frac{1}{2} \|F(\boldsymbol{w})\|_2^2 = \frac{1}{2} \sum_{i=1}^m F_i(\boldsymbol{w})^2.
\end{equation}
The Gauss-Newton method provides an iterative scheme to minimize this objective by approximating the true Hessian matrix $\nabla^2 \mathcal{L}(\boldsymbol{w})$. While the exact Hessian is given by $\nabla^2 \mathcal{L}(\boldsymbol{w}) = J_F(\boldsymbol{w})^T J_F(\boldsymbol{w}) + \sum_i F_i(\boldsymbol{w}) \nabla^2 F_i(\boldsymbol{w})$, the Gauss-Newton method neglects the second-order term $\sum_i F_i \nabla^2 F_i$, assuming it is small. This yields the approximated Hessian $H_{\mathrm{GN}} \approx J_F(\boldsymbol{w})^T J_F(\boldsymbol{w})$.

Setting the gradient to zero yields the Gauss-Newton update step $\Delta \boldsymbol{w} = \boldsymbol{w}^{k+1} - \boldsymbol{w}^k$, which is determined by solving the normal equations:
\begin{equation}
\big(J_F(\boldsymbol{w}^k)^T J_F(\boldsymbol{w}^k)\big) (\boldsymbol{w}^{k+1} - \boldsymbol{w}^k) = -J_F(\boldsymbol{w}^k)^T F(\boldsymbol{w}^k).
\end{equation}
This update is mathematically equivalent to solving the following linearized overdetermined system in the least-squares sense:
\begin{equation}
\boxed{J_F(\boldsymbol{w}^k)\,\boldsymbol{w}^{k+1} = J_F(\boldsymbol{w}^k)\,\boldsymbol{w}^k - F(\boldsymbol{w}^k).}
\label{eq:gauss_newton_linearized}
\end{equation}

\subsection{Our Newton-PINet: Newton linearization used in Tikhonov-regularized PINN}

In our Newton-PINet framework, the nonlinear PDE operators are first linearized at the continuous level (using Newton expansion), and the resulting discretized system is written in the linear form $\mathbf{A}(\boldsymbol{w}^k)\,\boldsymbol{w}^{k+1} = \mathbf{b}^k$. This linear system is then solved iteratively using linear least squares.

Using the Newton expansion for $uu_x$ (as shown in Eq.~(\ref{eq:uux newton expansion})) as an illustrative example, the linearization of $(uu_x)^{k+1}$ is:
\begin{equation}
(uu_x)^{k+1} \approx u^{k+1} u_x^k + u^k u_x^{k+1} - u^k u_x^k,
\end{equation}
where $k$ and $k+1$ denote the current and next iteration steps, respectively. Defining $u^k=\boldsymbol{\Phi} \boldsymbol{w}^k$ and $u_x^k=\boldsymbol{\Phi}_x \boldsymbol{w}^k$, we have:
\begin{equation}
\begin{aligned}
(uu_x)^{k+1} 
&= (\boldsymbol{\Phi} \boldsymbol{w})^{(k+1)} \odot (\boldsymbol{\Phi}_x \boldsymbol{w})^{(k+1)} \\ 
&\approx (\boldsymbol{\Phi}_x \boldsymbol{w}^{k}) \odot  (\boldsymbol{\Phi} \boldsymbol{w}^{(k+1)}) +(\boldsymbol{\Phi} \boldsymbol{w}^{k}) \odot (\boldsymbol{\Phi}_x \boldsymbol{w}^{(k+1)}) - (\boldsymbol{\Phi} \boldsymbol{w}^{k}) \odot (\boldsymbol{\Phi}_x \boldsymbol{w}^{k}).
\end{aligned}
\end{equation}
Through matrix manipulations, this becomes:
\[
(\boldsymbol{\Phi}_x \boldsymbol{w}^{k}) \odot  (\boldsymbol{\Phi} \boldsymbol{w}^{(k+1)})
= \operatorname{diag}(\boldsymbol{\Phi}_x \boldsymbol{w}^k)\,(\boldsymbol{\Phi} \boldsymbol{w}^{(k+1)})
= \big(\operatorname{diag}(\boldsymbol{\Phi}_x \boldsymbol{w}^k) \boldsymbol{\Phi} \big) \boldsymbol{w}^{(k+1)},
\]
\[
(\boldsymbol{\Phi} \boldsymbol{w}^{k}) \odot  (\boldsymbol{\Phi}_x \boldsymbol{w}^{(k+1)})
= \operatorname{diag}(\boldsymbol{\Phi} \boldsymbol{w}^k)\,(\boldsymbol{\Phi}_x \boldsymbol{w}^{(k+1)})
= \big(\operatorname{diag}(\boldsymbol{\Phi} \boldsymbol{w}^k) \boldsymbol{\Phi}_x \big) \boldsymbol{w}^{(k+1)}.
\]
Consequently, the constructed matrix $\mathbf{A}(\boldsymbol{w}^k)$ and the right-hand side vector $\mathbf{b}^k$ are explicitly given by:
\begin{equation}
\mathbf{A}(\boldsymbol{w}^k) 
= \begin{bmatrix}
\big(\operatorname{diag}(\boldsymbol{\Phi}_x \boldsymbol{w}^k) \boldsymbol{\Phi} \big) + \big(\operatorname{diag}(\boldsymbol{\Phi} \boldsymbol{w}^k) \boldsymbol{\Phi}_x \big)  \\
\boldsymbol{\Phi}(0,:)  \\ \boldsymbol{\Phi}(1,:) 
\end{bmatrix}, \quad
\mathbf{b}^k = \begin{bmatrix}
(\boldsymbol{\Phi} \boldsymbol{w}^{k}) \odot (\boldsymbol{\Phi}_x \boldsymbol{w}^{k})  \\
0 \\ 1
\end{bmatrix}.
\end{equation}
It can be verified that our $\mathbf{A}(\boldsymbol{w}^k)$ and $\mathbf{b}^k$ correspond exactly to the classical Gauss-Newton update:

\[
\boxed{J_F(\boldsymbol{w}^k)\,\boldsymbol{w}^{k+1} = J_F(\boldsymbol{w}^k)\boldsymbol{w}^k - F(\boldsymbol{w}^k),}
\]

i.e.,
\begin{equation}
\mathbf{A}(\boldsymbol{w}^k) = J_{F}(\boldsymbol{w}^k),
\end{equation}

\begin{equation}
\begin{split}
    \mathbf{b}^{k} &= J_F(\boldsymbol{w}^{k})\,\boldsymbol{w}^{k} - F(\boldsymbol{w}^{k}) \\
    &= 
    \begin{bmatrix}
        (\boldsymbol{\Phi}\boldsymbol{w}^{k}) \odot (\boldsymbol{\Phi}_x\boldsymbol{w}^{k}) 
        + (\boldsymbol{\Phi}\boldsymbol{w}^{k}) \odot (\boldsymbol{\Phi}_x\boldsymbol{w}^{k})
        - (\boldsymbol{\Phi}\boldsymbol{w}^{k}) \odot (\boldsymbol{\Phi}_x\boldsymbol{w}^{k}) \\
        0 \\
        1
    \end{bmatrix} \\
    &=
    \begin{bmatrix}
        (\boldsymbol{\Phi}\boldsymbol{w}^{k}) \odot (\boldsymbol{\Phi}_x\boldsymbol{w}^{k}) \\
        0 \\
        1
    \end{bmatrix}.
\end{split}
\end{equation}

Furthermore, by integrating Tikhonov regularization $\lambda_{\mathrm{reg}} \|\boldsymbol{w}\|^2$ into our linear solvers to improve conditioning, the objective function is effectively penalized. The corresponding normal equations solved in our algorithm become:
\begin{equation}
\big(J_F(\boldsymbol{w}^k)^T J_F(\boldsymbol{w}^k) + \lambda_{\mathrm{reg}} \mathbf{I} \big) \boldsymbol{w}^{k+1} = J_F(\boldsymbol{w}^k)^T \mathbf{b}^k,
\end{equation}
which categorizes our Newton-PINet approach as a \textbf{regularized Gauss-Newton method}.

The convergence of the classical regularized Gauss-Newton method depends on both the magnitude of the residuals and the strength of the regularization. Near-quadratic convergence can be achieved when the residuals are sufficiently small and the regularization term is sufficiently weak relative to the Gauss-Newton Hessian approximation. Newton-PINet exhibits near-quadratic convergence, similar to the regularized Gauss-Newton method, only when the meta-learned neural representation is sufficiently accurate, the initial guess is not too far from the solution, and the regularization coefficient is not too large.

\subsection{PINet: Picard linearization used in Tikhonov-regularized PINN}

As a baseline model, we employ Picard linearization in the PINet framework. For the Picard linearization, also referred to as the “lagging-of-coefficients’’ approach, one factor in the nonlinear term $uu_x$ is kept from the previous iteration $k$. Specifically, the linearization of $(uu_x)^{k+1}$ is given by
$(uu_x)^{k+1} \approx u^{k}\,u_x^{k+1}$, which can be written in matrix form as
\begin{equation}
\begin{aligned}
(uu_x)^{k+1} 
&\approx (\boldsymbol{\Phi} \boldsymbol{w})^{k} \odot (\boldsymbol{\Phi}_x \boldsymbol{w})^{(k+1)} \\
&= \big(\operatorname{diag}(\boldsymbol{\Phi} \boldsymbol{w}^k) \boldsymbol{\Phi}_x \big) \boldsymbol{w}^{(k+1)}.
\end{aligned}
\end{equation}

Consequently, in the linear system $\mathbf{A}(\boldsymbol{w}^k)\,\boldsymbol{w}^{k+1} = \mathbf{b}^k$, the corresponding matrix and right-hand side are

\begin{equation}
\mathbf{A}(\boldsymbol{w}^k) 
= \begin{bmatrix}
\big(\operatorname{diag}(\boldsymbol{\Phi} \boldsymbol{w}^k) \boldsymbol{\Phi}_x \big) \\
\boldsymbol{\Phi}(0,:)  \\ \boldsymbol{\Phi}(1,:) 
\end{bmatrix}, 
\quad
\mathbf{b}^k = \begin{bmatrix}
0  \\
0 \\ 1
\end{bmatrix}.
\end{equation}

Since the Picard method only updates one factor at each iteration while keeping the other fixed, the convergence rate is generally linear rather than quadratic.

\subsection{Experimental evaluation of convergence}

The convergence behavior of Newton-PINet is further verified experimentally. The residual norm $|\mathbf{A}\boldsymbol{w}- \mathbf{b}|$ and the empirical order of convergence (EOC), $\mathrm{EOC}=\frac{\log\!\left(\|e_{k+1}\|/\|e_k\|\right)}
{\log\!\left(\|e_k\|/\|e_{k-1}\|\right)}$, are evaluated for both Newton and Picard iterations on the nonlinear Burgers problem. As shown in Table~\ref{tab:convergence_comparison}, using the initialization $u_{\mathrm{guess}}(t)=u_{t=0}$ for all time steps, Newton-PINet achieves an observed EOC of $1.7$--$2.0$, a smaller final residual ($2.98\times10^{-3}$), and a substantially lower MSE ($4.59\times10^{-7}$) than Picard (EOC $\approx1.2$), indicating near-quadratic convergence when the initialization is sufficiently close to the solution. Table~\ref{tab:initial_guess_impact} further shows that poorer initializations, i.e., larger $\ell_2$ distances from the ground-truth solution $u_{\text{gt}}$, lead to a lower convergence order of Newton-PINet. As discussed in Appendix~\ref{app:Effect of initial guess in nonlinear iterations}, this sensitivity can be mitigated by initializing all time steps with the initial field and employing temporal block decomposition for more challenging nonlinear time-dependent problems such as the K-S equation. 

\begin{table}[htbp]
    \centering
    \caption{Convergence process comparison between Newton and Picard iterations}
    \label{tab:convergence_comparison}
    \small
    \begin{tabular}{c ccc ccc}
    \toprule
    & \multicolumn{3}{c}{Newton} & \multicolumn{3}{c}{Picard} \\
    \cmidrule(lr){2-4} \cmidrule(lr){5-7}
    Iter. & $\|\mathbf{A}\boldsymbol{w}- \mathbf{b}\|$ & EOC & MSE & $\|\mathbf{A}\boldsymbol{w}- \mathbf{b}\|$ & EOC & MSE \\
    \midrule
    1 & $1.27 \times 10^{-2}$ & --   & $4.91 \times 10^{-2}$ & $8.64 \times 10^{-3}$ & --   & $2.65 \times 10^{-2}$ \\
    2 & $5.62 \times 10^{-2}$ & --   & $3.28 \times 10^{-3}$ & $6.30 \times 10^{-3}$ & --   & $4.48 \times 10^{-3}$ \\
    3 & $7.26 \times 10^{-3}$ & \textcolor{red}{1.98} & $2.71 \times 10^{-5}$ & $6.01 \times 10^{-3}$ & \textcolor{red}{1.26} & $5.15 \times 10^{-4}$ \\
    4 & $2.98 \times 10^{-3}$ & \textcolor{red}{1.73} & $4.69 \times 10^{-7}$ & $4.54 \times 10^{-3}$ & \textcolor{red}{1.26} & $5.20 \times 10^{-5}$ \\
    5 & $2.98 \times 10^{-3}$ & \textcolor{red}{1.85} & $4.59 \times 10^{-7}$ & $4.08 \times 10^{-3}$ & \textcolor{red}{1.16} & $1.50 \times 10^{-5}$ \\
    \bottomrule
    \end{tabular}
\end{table}

\begin{table}[htbp]
    \centering
    \caption{Impact of initial guess on the convergence}
    \label{tab:initial_guess_impact}
    \small
    \setlength{\tabcolsep}{3pt}
    \begin{tabular}{l c ccc ccc}
    \toprule
    & & \multicolumn{3}{c}{Newton} & \multicolumn{3}{c}{Picard} \\
    \cmidrule(lr){3-5} \cmidrule(lr){6-8}
    Initial guess & $\|u_0-u_{\text{gt}}\|_2$ & $\|\mathbf{A}\boldsymbol{w}-\mathbf{b}\|$ & EOC & MSE & $\|\mathbf{A}\boldsymbol{w}-\mathbf{b}\|$ & EOC & MSE \\
    \midrule
    $u(t)=u_{t=0}$ & 28.59 & $2.98\times10^{-3}$ & 1.85 & $4.59\times10^{-7}$ & $4.08\times10^{-3}$ & 1.16 & $1.50\times10^{-5}$ \\
    Zero field & 48.93 & $2.98\times10^{-3}$ & 1.63 & $4.62\times10^{-7}$ & $4.60\times10^{-3}$ & 1.26 & $5.19\times10^{-5}$ \\
    Constant field & 94.73 & 1.33 & 0.53 & $1.53\times10^{-1}$ & $4.07\times10^{-1}$ & 0.87 & $6.01\times10^{-2}$ \\
    \bottomrule
    \end{tabular}
\end{table}

The empirical results (Table~\ref{tab:lambda_reg_impact} below) show that increasing $\lambda_{\mathrm{reg}}$ degrades the observed empirical order of convergence (EOC), transitioning Newton-PINet from near-quadratic convergence to linear/sublinear convergence. When $\lambda_{\mathrm{reg}}$ is sufficiently small, it mainly serves to improve numerical conditioning. However, as $\lambda_{\mathrm{reg}}\mathbf{I}$ begins to dominate, the regularized update $(\mathbf{A}^T\mathbf{A}+\lambda_{\mathrm{reg}}\mathbf{I})^{-1}\mathbf{A}^T\mathbf{b}$ approaches $\frac{1}{\lambda_{\mathrm{reg}}}\mathbf{A}^T\mathbf{b},$ transforming the Newton update into a heavily damped gradient-descent step and thereby degrading the convergence order.

\begin{table}[htbp]
    \centering
    \caption{Impact of $\lambda_{\mathrm{reg}}$ on convergence order (EOC) and MSE}
    \label{tab:lambda_reg_impact}
    \small
    \begin{tabular}{c c cc cc}
    \toprule
    & & \multicolumn{2}{c}{Newton} & \multicolumn{2}{c}{Picard} \\
    \cmidrule(lr){3-4} \cmidrule(lr){5-6}
    Scale & $\lambda_{\mathrm{reg}}$ & EOC & MSE & EOC & MSE \\
    \midrule
    $1\times$      & $7.58 \times 10^{-4}$ & 1.85 & $4.59 \times 10^{-7}$ & 1.16 & $1.50 \times 10^{-5}$ \\
    $10\times$     & $7.58 \times 10^{-3}$ & 1.80 & $8.90 \times 10^{-7}$ & 1.16 & $1.45 \times 10^{-5}$ \\
    $10^{2}\times$    & $7.58 \times 10^{-2}$ & 1.55 & $1.27 \times 10^{-5}$ & 1.15 & $3.22 \times 10^{-5}$ \\
    $10^{3}\times$   & $7.58 \times 10^{-1}$ & 0.93 & $3.19 \times 10^{-4}$ & 1.08 & $4.52 \times 10^{-4}$ \\
    $10^{4}\times$  & 7.58                  & 0.85 & $6.64 \times 10^{-3}$ & 0.96 & $5.99 \times 10^{-3}$ \\
    $10^{5}\times$ & $7.58 \times 10^{1}$  & 0.86 & $4.64 \times 10^{-2}$ & 0.89 & $3.91 \times 10^{-2}$ \\
    \bottomrule
    \end{tabular}
\end{table}

\section{Problem description and experimental results}  
\label{app:Datasets_and_experiment}
\subsection{Description of nonlinear PDE problems}   
\label{app:PDE_problems}

\textbf{(i) Nonlinear convection-type PDEs}

\underline{Burgers' equation (1D + time)}

\begin{equation}
    \frac{\partial u}{\partial t} 
    + u \frac{\partial u}{\partial x} 
    - \gamma \frac{\partial^{2} u}{\partial x^{2}} = 0, \quad x \in [-1,1], \, t \in [0,1].
\end{equation}

a) Varying PDE parameter:

The temporal and spatial resolutions are set to $51 \times 129$ with periodic boundary conditions. We fix the initial condition as $u(x,0) = -\sin(\pi x)$, and vary the viscosity parameter $\gamma$ within the range $[0.001,0.05]$ with an increment of $0.001$ to generate 50 tasks; 16 of them are randomly selected as training tasks, while the remaining 34 are used for testing. The model architecture consists of 4 hidden layers with 128 neurons per layer. Time block decomposition is not employed (a single time block), and the step of the fine-tuning (nonlinear) iterations is set to 6.

b) Varying initial condition:

The temporal and spatial resolutions are set to $101 \times 101$ with periodic boundary conditions. Four distinct viscosity values $\gamma \in \{0.1, 0.01, 0.001, 0.0001\}$ are considered. For each $\gamma$, the initial conditions are sampled from a Gaussian random field $\mathcal{N}(0, 625(-\Delta + 25I)^{-4})$, consistent with \cite{wang2021learning}. We generate 50 training tasks and 1000 test tasks for each viscosity, and experiments are conducted independently for each $\gamma$. The model architecture consists of 4 hidden layers with 128 neurons per layer, and the nonlinear iteration step is set to 4. We evaluate two model variants: unsupervised Newton-PINet (1 time block) and hybrid Newton-PINet (4 time blocks).

\underline{Generalized Korteweg-de Vries (KdV) equation (1D + time)}

\begin{equation}
    \frac{\partial u}{\partial t} 
    + \beta_{1} u \frac{\partial u}{\partial x} 
    + \beta_{2} u^{2} \frac{\partial u}{\partial x} 
    + \beta_{3} u^{3} \frac{\partial u}{\partial x} 
    + \delta \frac{\partial^{3} u}{\partial x^{3}} = 0,  \quad x \in [-1,1], \, t \in [0,1]
\end{equation}

The temporal and spatial resolutions are set to $101 \times 257$ with periodic boundary conditions. We consider three different forms of nonlinear convection terms, corresponding to three combinations of $(\beta_{1}, \beta_{2}, \beta_{3})$. For each combination, 15 tasks are randomly generated by varying $\delta$, leading to a total of 45 tasks. Specifically:  
\[
\text{KdV: } (\beta_{1}, \beta_{2}, \beta_{3}) = (1,0,0), \quad \delta \in [0.032^{2}, 0.12^{2}],
\]
\[
M\text{-KdV: } (\beta_{1}, \beta_{2}, \beta_{3}) = (0,1,0), \quad \delta \in [0.02^{2}, 0.06^{2}],
\]
\[
G\text{-KdV: } (\beta_{1}, \beta_{2}, \beta_{3}) = (0,0,1), \quad \delta \in [0.08^{2}, 0.15^{2}].
\]
Among them, 10 tasks are randomly chosen for training and the remaining 35 tasks for testing. The model architecture consists of 4 hidden layers with 256 neurons per layer. 6 time blocks are employed, and the nonlinear iteration step is set to 6.

\underline{Kuramoto-Sivashinsky (K-S) equation (1D + time)}

\begin{equation}
    \frac{\partial u}{\partial t} 
    + \beta u \frac{\partial u}{\partial x} 
    - \gamma \frac{\partial^{2} u}{\partial x^{2}} 
    + \sigma \frac{\partial^{4} u}{\partial x^{4}} = 0,  \quad x \in [0,2\pi], \, t \in [0,1]
\end{equation}

The simulated data have a temporal and spatial resolution of $251 \times 509$, which is downsampled to $63 \times 129$ for training and evaluation. Periodic boundary conditions are imposed. The fixed coefficients are set as $\beta = 100/16$ and $\gamma = 100/16^{2}$. The parameter $\sigma$ varies within the range $[200/16^{4},\, 300/16^{4}]$, along with varying initial conditions:
\[
u(x,0) = \sum_{j=1}^{J} A_{j} \sin\left(\frac{2\pi l_{j} x}{L} + \phi_{j}\right),
\]
where $J=5$, $L=1$, $A_{j} \in [-0.8,0.8]$, $l_{j} \in \{0,1,2,3,4\}$, and $\phi_{j} \in [-\pi,\pi]$.  
In total, 150 tasks are generated, of which 100 are used for training and 50 for testing. The model architecture consists of 4 hidden layers with 256 neurons per layer. 15 time blocks are employed, and the nonlinear iteration step is set to 4.

\underline{Lid-driven cavity (LDC) equations (2D)}

The 2D steady incompressible Navier-Stokes equations in LDC problem are given by
\begin{equation}
\begin{cases}
    \frac{\partial u}{\partial x} + \frac{\partial v}{\partial y} = 0,\\[2mm]
    u \frac{\partial u}{\partial x} + v \frac{\partial u}{\partial y} + \frac{\partial p}{\partial x} - \frac{1}{Re}\left(\frac{\partial^2 u}{\partial x^2} + \frac{\partial^2 u}{\partial y^2}\right) = 0,\\[1mm]
    u \frac{\partial v}{\partial x} + v \frac{\partial v}{\partial y} + \frac{\partial p}{\partial y} - \frac{1}{Re}\left(\frac{\partial^2 v}{\partial x^2} + \frac{\partial^2 v}{\partial y^2}\right) = 0,
\end{cases}
\end{equation}
where $(x,y) \in [0,1]^2$ and $Re$ is the Reynolds number.

The simulated data are generated on a $200\times200$ grid and then downsampled to $51\times51$ for training and testing. The boundary conditions are
\[
\text{top lid: } u=1, v=0; \quad
\text{other walls: } u=v=0.
\]
The $Re$ ranges from 1 to 1000 (specifically 1, 5, 10, 25, 50, 80, 100, ..., 1000 with a step of 50 from 100 to 1000), yielding 25 flow conditions. We use $Re$ = 1, 50, 200, 400, 600, 800 as the training set and the remaining cases as the test set. The model consists of 4 hidden layers with 128 neurons per layer. With three output variables ($u$, $v$, $p$), the resulting weight dimension of the output layer (i.e., the number of columns in matrix $\mathbf{A}$) is $128 \times 4 \times 3 = 1536$. The nonlinear iteration step is set to~8.

\textbf{(ii) Nonlinear forcing-type PDEs}

\underline{Convection-diffusion-reaction (CDR) equation (1D + time)}
\begin{equation}
    \frac{\partial u}{\partial t} 
    + \alpha \frac{\partial u}{\partial x} 
    - \gamma \frac{\partial^{2} u}{\partial x^{2}}
    - \rho_{1} u + \rho_{2} u^{2} + \rho_{3} u^{3} = 0, 
    \quad x \in [0,1], \, t \in [0,1].
\end{equation}
The simulated data have a temporal and spatial resolution of $201 \times 257$, which are downsampled to $101 \times 129$ for training and evaluation. Periodic boundary conditions are imposed. The parameters are set as $\alpha = 1$, $\gamma \in \{0.005, 0.01, 0.05\}$, $\rho_{1}, \rho_{2} \in \{0,1,3,5\}$, and $\rho_{3} = 5$. The initial conditions are generated from
\[
u(x,0) = \sum_{j=1}^{J} A_{j} \sin(l_{j} x + \phi_{j}),
\]
where $J=5$, $A_{j} \in [0.1,0.5]$, $l_{j} \in \{1,2,3,4\}$, and $\phi_{j} \in [-\pi,\pi]$.  

In total, 200 tasks are generated, with 40 used for training and 160 for testing. The model architecture consists of 4 hidden layers with 128 neurons per layer. 4 time blocks are employed, and the nonlinear iteration step is set to 4.

\underline{Klein-Gordon equation (1D + time)}
\begin{equation}
    \frac{\partial^{2} u}{\partial t^{2}} 
    - \gamma \frac{\partial^{2} u}{\partial x^{2}} 
    + u^{3} = q, 
    \quad x \in [0,1], \, t \in [0,1]
\end{equation}

\begin{equation*}
    \text{ICs:} \quad 
    \frac{\partial u}{\partial t}(x,0) = 0.
\end{equation*}
The solution is defined as 
$u(x,t) = k_{1} x \cos(k_{2} \pi t) + k_{3} (x t)^{3}$, which is used to derive the corresponding  Dirichlet BCs, and source term $q$. The spatio-temporal grid resolution is set to $32 \times 32$ for meta-learning, while a finer grid of $128 \times 128$ is used for testing. The parameters vary within $\gamma, k_{1}, k_{3} \in [0.5,3]$ and $k_{2} \in [1,7]$. A total of 80 tasks are generated, with 16 used for meta-learning and 64 for testing. The model network consists of 4 hidden layers with 128 neurons per layer. 1 time block is employed, and the nonlinear iteration step is set to 4.

\underline{Hyperbolic heat equation (1D + time)}
\begin{equation}
    \frac{\partial u}{\partial t} 
    - \gamma \frac{\partial^{2} u}{\partial x^{2}} 
    + k_{1} \sinh(u) + k_{2} \cosh(u) + k_{3} \tanh(u) = q, 
    \quad x \in [-1,1], \, t \in [0,1].
\label{eq:1D+time hyperbolic heat}
\end{equation}
The solution is defined as 
$u(x,t,k_{1},k_{2},k_{3}) = \sin(k_{1} k_{2} \pi x) \cos(k_{3} \pi x) \, e^{-\pi t^{2}}$, which is used to derive the corresponding ICs, Dirichlet BCs, and source term $q$. The spatio-temporal grid resolution is set to $32 \times 32$ for meta-learning, while a finer grid of $128 \times 128$ is used for testing. The parameters vary within $\gamma \in [0.2,3]$ and $k_{1}, k_{2}, k_{3} \in [0.2,2.5]$. A total of 80 tasks are generated, with 16 used for training and 64 for testing. The model network consists of 2 hidden layers with 450 neurons per layer. 1 time block is employed, and the nonlinear iteration step is set to 4.

\underline{Logarithmic heat equation (1D + time)}
\begin{equation}
    \frac{\partial u}{\partial t} 
    - \gamma \frac{\partial^{2} u}{\partial x^{2}} 
    + k_{1} u \, \log(k_{2} u) 
    + k_{3} e^{k_{4} u} = q, 
    \quad x \in [-1,1], \, t \in [0,1].
\end{equation}
The solution is defined as 
$u(x,t,k_{2},k_{4}) = \big(\sin(k_{2} \pi x) + 1.5\big) e^{-k_{4} \pi x^{2}} e^{-\pi t^{2}}$, which is used to derive the corresponding ICs, Dirichlet BCs, and source term $q$. The spatio-temporal grid resolution is set to $32 \times 32$ for meta-learning, while a finer grid of $128 \times 128$ for testing. The parameters vary within $\gamma, k_{1}, k_{3} \in [0.5,5]$ and $k_{2}, k_{4} \in [0.5,2]$. A total of 80 tasks are generated, with 16 used for meta-learning and 64 for testing. The model network consists of 2 hidden layers with 450 neurons per layer. 1 time block is employed, and the nonlinear iteration step is set to 4.

\underline{Helmholtz equation (2D)}
\begin{equation}
    \frac{\partial^{2} u}{\partial x^{2}} 
    + \frac{\partial^{2} u}{\partial y^{2}} 
    - 100 u + 10 \cosh(u) = q, 
    \quad x, y \in [0,1.5].
\end{equation}
The solution is defined as 
$u(x,y) = 4 \cos(\alpha_{1} \pi x^{2}) \cos(\alpha_{2} \pi y^{2})$, which is used to derive the corresponding Dirichlet BCs, and source term $q$. The spatial grid resolution is set to $32 \times 32$ for meta-learning, while a finer grid of $128 \times 128$ is used for testing. The parameters vary within $\alpha_{1}, \alpha_{2} \in [1,3]$. A total of 80 tasks are generated, with 20 used for meta-learning and 60 for testing. The model network consists of 2 hidden layers with 450 neurons per layer. The nonlinear iteration step is set to 4.

\underline{Parametric diffusion-reaction equation (2D)}
\begin{equation}
    \frac{\partial^{2} u}{\partial x^{2}} 
    + \frac{\partial^{2} u}{\partial y^{2}} 
    + u (1 - u^{2}) = q, 
    \quad x, y \in [-1,1].
\end{equation}
The solution is defined as $u(x,y,a_{1}, a_{2},w_{1}, w_{2}, w_{3}, w_{4}) = a_{1} \tanh(w_{1} x) \tanh(w_{2} y) + a_{2} \sin(w_{3} x) \sin(w_{4} y)$, which is used to derive the corresponding Dirichlet BCs and source term $q$. The spatial grid resolution is set to $32 \times 32$ for meta-learning, while a finer grid of $128 \times 128$ is used for testing. The parameters vary within $a_{1}, a_{2} \in [0.1,1]$ and $w_{1}, w_{2}, w_{3}, w_{4} \in [1,5]$. 17 tasks are generated for meta-learning and 100 for testing. The model network consists of 2 hidden layers with 450 neurons per layer. The nonlinear iteration step is set to 4.

\underline{Parametric diffusion-reaction equation (3D)}
\begin{equation}
    \frac{\partial^{2} u}{\partial x^{2}}
    +\frac{\partial^{2} u}{\partial y^{2}}
    +\frac{\partial^{2} u}{\partial z^{2}}
    +u(1-u^{2})=q,
    \quad x,y,z\in[-1,1].
\label{eq:3D Parametric diffusion-reaction}
\end{equation}
The solution is defined as $u(x,y,z,a_{1},a_{2},w_{1},w_{2},w_{3},w_{4},w_{5},w_{6}) = a_{1} \tanh(w_{1}x) \tanh(w_{2}y) \tanh(w_{3}z) + a_{2} \sin(w_{4}x) \sin(w_{5}y) \sin(w_{6}z)$, which is used to derive the corresponding Dirichlet boundary conditions and source term $q$. The spatial grid resolution is set to $32\times32\times32$ for both meta-training and testing. The PDE parameters are sampled from $a_1,a_2\in[0.1,1]$ and $w_1,w_2,w_3,w_4,w_5,w_6\in[1,5]$. 17 tasks are generated for meta-learning and 100 for testing. The network consists of two hidden layers with 450 neurons per layer, and the number of nonlinear iterations is set to 4.

The meta-learning configurations for each problem are summarized in Table~\ref{tab:Experimental config summary}.

\begin{table}[h]
\setlength{\abovecaptionskip}{8pt}
\setlength{\belowcaptionskip}{10pt}
\caption{Summary of meta-learning parameter configurations on nonlinear PDEs in this study. For time-dependent PDEs, the mesh size is reported as $t \times x$, whereas for time-independent PDEs it is reported as $x \times y\,  (\times z)$.
``Mesh size'' and ``No.~A rows'' refer to settings in the meta-learning stage unless otherwise marked with ``(test)'', which denotes the inference stage.
``Dim.'' denotes the dimension, and ``No.'' stands for ``number of''. ``--'' indicates that the time-blocking strategy is not applicable for time-independent problems.}

\label{tab:Experimental config summary}
\footnotesize 
\setlength{\tabcolsep}{2pt} 
\begin{center}
\begin{tabular}{lccccccc}
 
\multirow{-2}{*}{{Problem}}              & \multirow{-2}{*}{\begin{tabular}[c]{@{}c@{}} Mesh size \\ (test)\end{tabular}} & \multirow{-2}{*}{\begin{tabular}[c]{@{}c@{}}Dim. $n_{\text{pde}} + n_{\text{ic}}+ n_{\text{bc}}$; \\ No. $\mathbf{A}$ rows (test)\end{tabular}}   & \multirow{-2}{*}{\begin{tabular}[c]{@{}c@{}}Dim. $\boldsymbol{w}$; \\ No. $\mathbf{A}$ columns\end{tabular}} & \multirow{-2}{*}{\begin{tabular}[c]{@{}c@{}}No. train/\\ test tasks\end{tabular}}  & \multirow{-2}{*}{\begin{tabular}[c]{@{}c@{}}Nonlinear \\ iteration\end{tabular}} & \multirow{-2}{*}{\begin{tabular}[c]{@{}c@{}}No. \\ time blocks\end{tabular}}     \\  \toprule
 {\begin{tabular}[c]{@{}l@{}}Burgers’ \\  (1D + time)\end{tabular}}                                                                                     & 51×129                                                               &      6810         & 512                      & 16/34                           & 6                                     & 1                                \\ \midrule
 
{\begin{tabular}[c]{@{}l@{}}Generalized KdV \\ (1D + time)\end{tabular}}    & 101×257        &  4402                                                                   & 1024                     & 10/35                                      & 6                                     & 6                                \\ \midrule

{\begin{tabular}[c]{@{}l@{}}Kuramoto-Sivashinsky \\  (1D + time)\end{tabular}}                                              & 63×129      & 649                                                                       & 1024                     & 100/50                                                                                                           & 4                                     & 15                               \\ \midrule

{\begin{tabular}[c]{@{}l@{}}Convection-diffusion- \\  reaction (1D + time)\end{tabular}}                        & 101×129        &   3404                                                                  & 512                      & 40/160                                                                                                            & 4                                     & 4                                \\ \midrule

{\begin{tabular}[c]{@{}l@{}}Klein-Gordon \\ (1D + time)\end{tabular}}                         & \begin{tabular}[c]{@{}c@{}}32×32 \\ (128×128)\end{tabular}                  &      1148 (16892)   & 512                      & 16/64                                                                                                            & 4                                     & 1                                \\ \midrule

{\begin{tabular}[c]{@{}l@{}}Hyperbolic heat \\  (1D + time)\end{tabular}}            & \begin{tabular}[c]{@{}c@{}}32×32\\ (128×128)\end{tabular}                    &    1148 (16892)    & 900                      & 16/64                                                                                                            & 4                                     & 1                                \\ \midrule

{\begin{tabular}[c]{@{}l@{}}Logarithmic heat   \\ (1D + time)\end{tabular}}                          & \begin{tabular}[c]{@{}c@{}}32×32\\ (128×128)\end{tabular}                   &     1148 (16892)    & 900                      & 16/64                                                                                                        & 4                                     & 1                                \\ \midrule

{\begin{tabular}[c]{@{}l@{}}Helmholtz \\  (2D)\end{tabular}}                & \begin{tabular}[c]{@{}c@{}}32×32\\ (128×128)\end{tabular}      &  1148 (16892)                    & 900                      & 20/60                                        & 4                                     & --  \\  \midrule

{\begin{tabular}[c]{@{}l@{}}Parametric diffusion\\-reaction (2D)\end{tabular}}                                                            & \begin{tabular}[c]{@{}c@{}}32×32\\ (128×128)\end{tabular}                     &  1148 (16892)    & 900                      & 17/100                                                                    & 4                                     & --                                \\ \midrule

{\begin{tabular}[c]{@{}l@{}}Lid-driven cavity \\  (2D)\end{tabular}}                                                                          & 51×51                                                                          &7595   & 1536                      & 6/19                                                                                  & 8                                     & --             \\ \midrule 

{\begin{tabular}[c]{@{}l@{}}Parametric diffusion\\-reaction (3D)\end{tabular}}                                                            & 32×32×32                   &  38536    & 900                      & 17/100                                                                    & 4                                     & --                                \\ \bottomrule 
              
\end{tabular}
\end{center}
\end{table}

\subsection{Data generation, error metrics, and computational setup} 
\label{app:Data generation and computational setup}
Datasets for Burgers, generalized KdV, K-S, and CDR problems are generated using the \texttt{Chebfun} package \citep{driscoll2014chebfun} with spectral Fourier discretization and a fourth-order exponential time-differencing Runge-Kutta scheme (ETDRK4) \citep{cox2002exponential}. Datasets for LDC problem are generated using a computational fluid dynamics (CFD) method \citep{chiu2021development}. For the remaining benchmark problems, analytical solutions are available.  

We evaluate the generalization accuracy of the models using both mean squared error (MSE) and relative $L^2$ error. In addition, for the 1D nonlinear diffusion-reaction problem, we adopt the relative MSE metric, consistent with the baseline study~\citep{boudec2025learning} for comparison (see Table~\ref{tab:Reaction_diffusion_model_compare} and Figure~\ref{fig:ICLR_DR-Visualization plot}).
\begin{equation}
\mathrm{MSE} = \frac{1}{n} \sum_{s=1}^n \big( u^{\text{label}}_s - u_s \big)^2 ,
\end{equation}
\begin{equation}
\mathrm{Relative}~L^2 = 
\frac{\left\lVert u^{\text{label}} - u \right\rVert_2}
{\left\lVert u^{\text{label}} \right\rVert_2} ,
\end{equation}
\begin{equation}
\mathrm{Relative~MSE} = 
\frac{\sum_{s=1}^n \big( u^{\text{label}}_s - u_s \big)^2}
{\sum_{s=1}^n \big( u^{\text{label}}_s \big)^2} ,
\end{equation}
where $u^{\text{label}}$ and $u$ denote the ground-truth and predicted solutions for a given PDE instance, and $n$ is the total number of collocation points, including those for the PDE residual, IC/BC constraints.

In this study, all models are implemented in the JAX framework~\citep{bradbury2018jax}. For the KdV, K-S, CDR, and LDC problems, model training is parallelized across four Tesla V100 GPUs; all other models are trained on a single Tesla V100 GPU. Inference for all models is performed on a single Tesla V100 GPU.

\subsection{Summary of model performance on nonlinear PDE problems}
\label{app:Summary_of_model_performance}

Table~\ref{tab:Experimental-summary} in the main text summarizes the model performance on each problem. For each problem, besides the \textbf{Newton-PINet} model proposed in this study, two additional baseline models were tested: one that uses the Picard method, referred to as \textbf{PINet}, and a purely data-driven \textbf{DNN} baseline which incorporates PDE parameters as inputs. Note that the tables do not include comparisons with other benchmark models on problems such as the Burgers' equation with varying initial conditions; the detailed results can be found in Table~\ref{tab:Burgers-Compar}, Table~\ref{tab:Reaction_diffusion_model_compare}, and Table~\ref{tab:Baldwin-PINN-Meta-PINN-compare}. 

As shown, Newton-PINet achieves superior test accuracy across nearly all nonlinear PDEs compared to DNN and PINet, except for the Klein-Gordon equation. The Klein-Gordon equation involves second-order temporal dynamics, making the nonlinear iterations more sensitive to the quality of the initial guess. Initializing all time steps with the initial field $u(t=0)$ may still be relatively far from the true solution, leading to a slight degradation in convergence for Newton-PINet. However, the difference in prediction accuracy is negligible (MSE: $1.88\times10^{-7}$ vs. $2.36\times10^{-7}$). Notably, for hyperbolic and logarithmic heat problems, the Picard approach employed by PINet fails to converge (leading to NaN values during meta-learning), whereas Newton-PINet remains stable and achieves accurate generalization.

\begin{figure}[t]
\begin{center}

\includegraphics[width=1\linewidth]{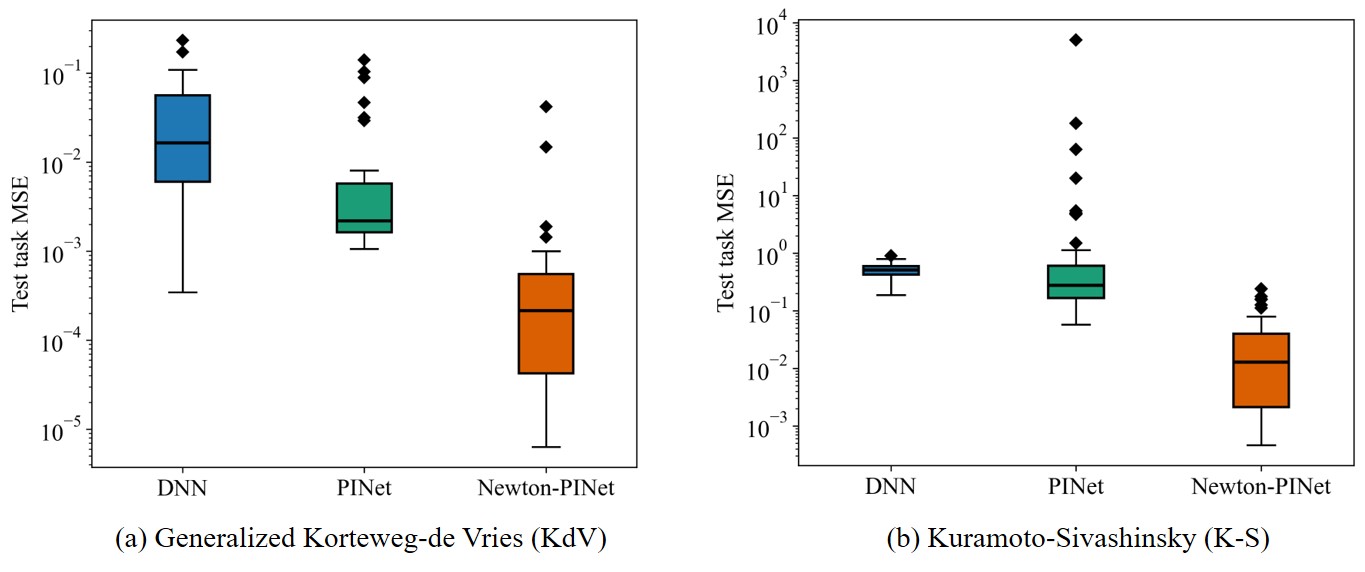}
\caption{\label{fig:Test MSE results of KDV and KS}MSE distributions across all test tasks for different models on the generalized KdV and K-S problems.}
\end{center}
\end{figure}

\begin{figure}[h]
\begin{center}
\includegraphics[width=1\linewidth]{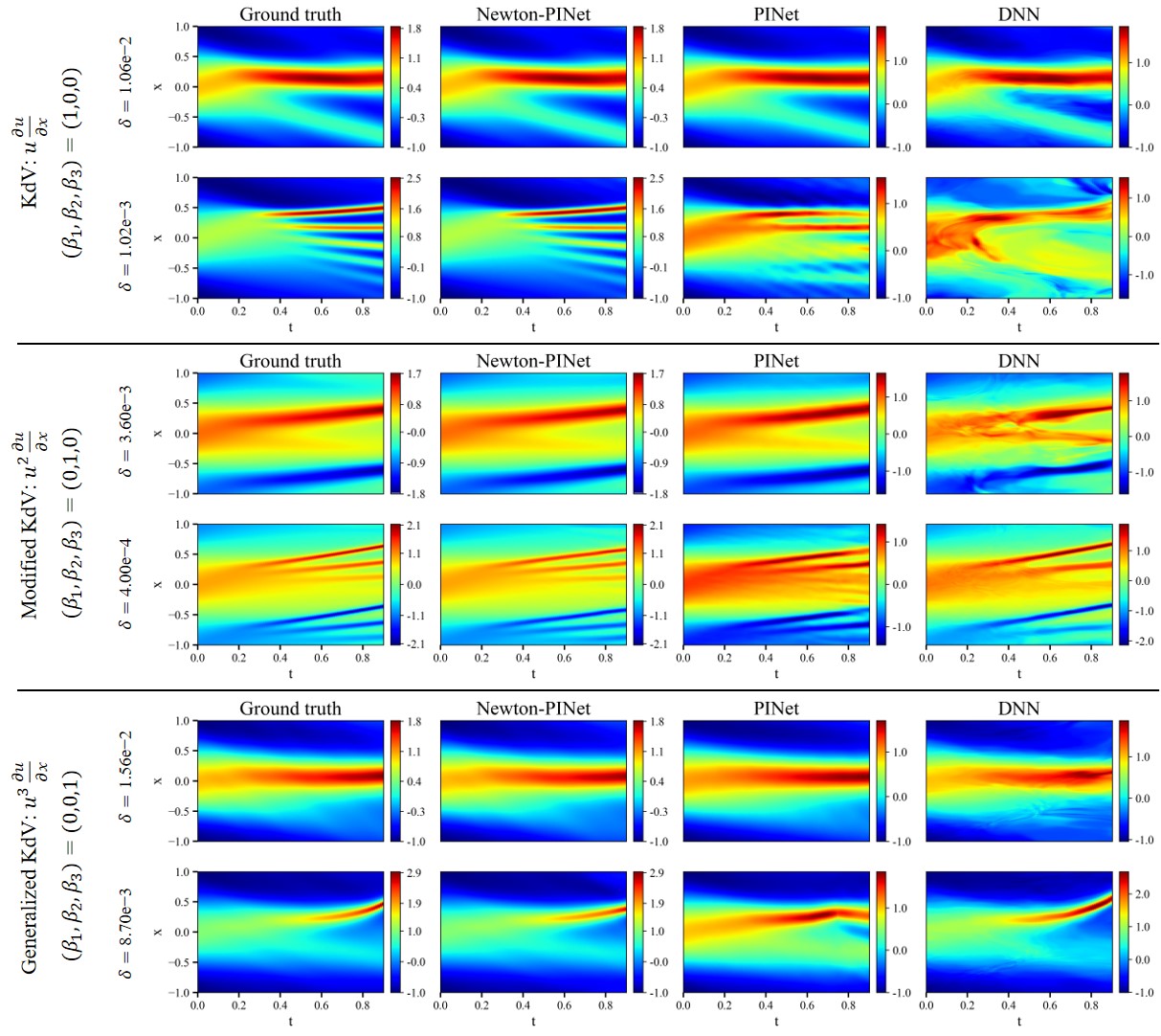}
\caption{\label{fig:KDV-Visualization plot}Prediction results of Newton-PINet for the generalized KdV problem.}
\end{center}
\end{figure}

\begin{figure}[h]
\begin{center}
    \includegraphics[width=1\linewidth]{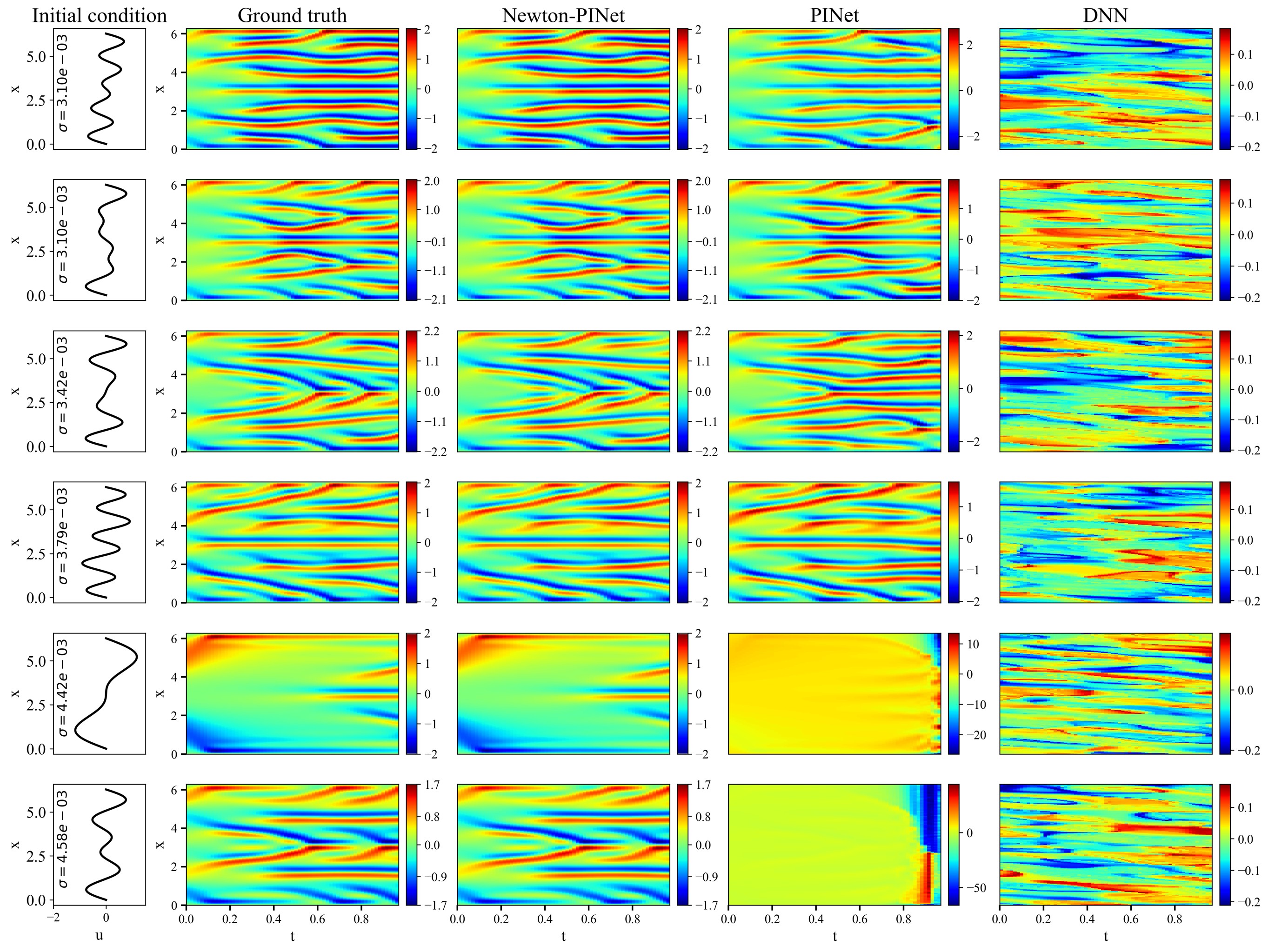}
    \caption{\label{fig:KS-Visualization plot}Prediction results of Newton-PINet for the K-S problem.}
\end{center}
\end{figure}

\begin{figure}[h]
\begin{center}
    \includegraphics[width=1\linewidth]{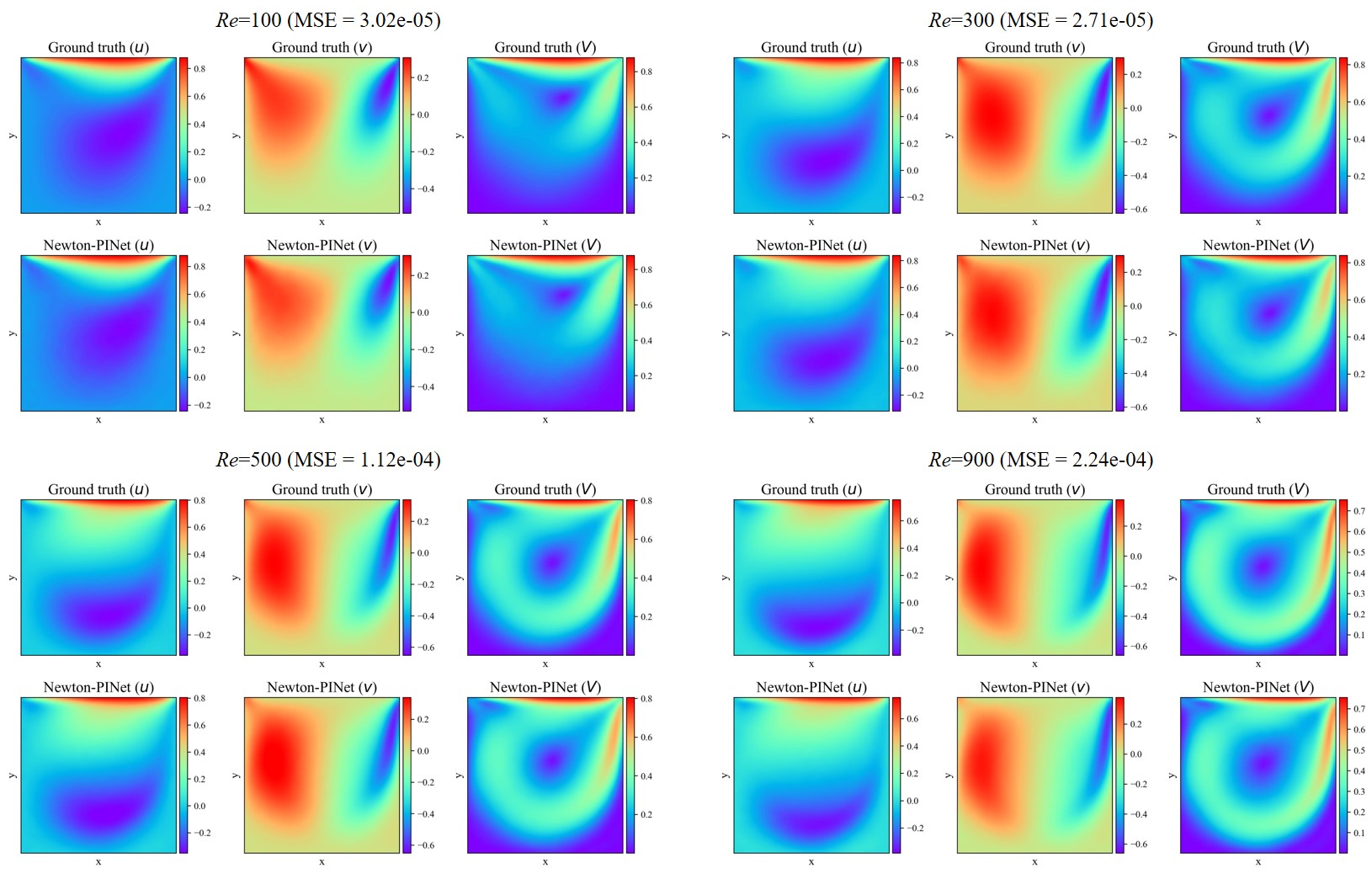}
    \caption{\label{fig:LDC-Re25-100-500-900}
    Prediction result of Newton-PINet for the 2D lid-driven cavity problem.
    Velocity magnitude is computed as $V = \sqrt{u^2 + v^2}$.
    }
\end{center}
\end{figure}

\paragraph{Prediction results of generalized KdV, K-S, and LDC problems:} 
Due to the coexistence of nonlinear convection terms and high-order dispersion operators, solving the generalized KdV and K-S equations is particularly challenging for PINN-based models. To mitigate these difficulties, we adopt a larger number of temporal domain partitions to stabilize training and improve convergence (6 time blocks for the generalized KdV and 15 time blocks for the K-S equations). As shown in Table~\ref{tab:Experimental-summary} and Figure~\ref{fig:Test MSE results of KDV and KS}, Newton-PINet achieves test errors of $\mathrm{MSE}=1.94\times 10^{-3}$ on the generalized KdV, $\mathrm{MSE}=3.45\times 10^{-2}$ on the K-S, and $\mathrm{MSE}=1.67\times 10^{-4}$ on the LDC problem, outperforming PINet and DNN under the same configuration. Figures~\ref{fig:KDV-Visualization plot}, ~\ref{fig:KS-Visualization plot}, ~\ref{fig:LDC-Re25-100-500-900} show the prediction results of the Newton-PINet model on test tasks for the generalized KdV, K-S, and LDC problems, respectively.

\begin{figure}[h]
\begin{center}
    \includegraphics[width=1\linewidth]{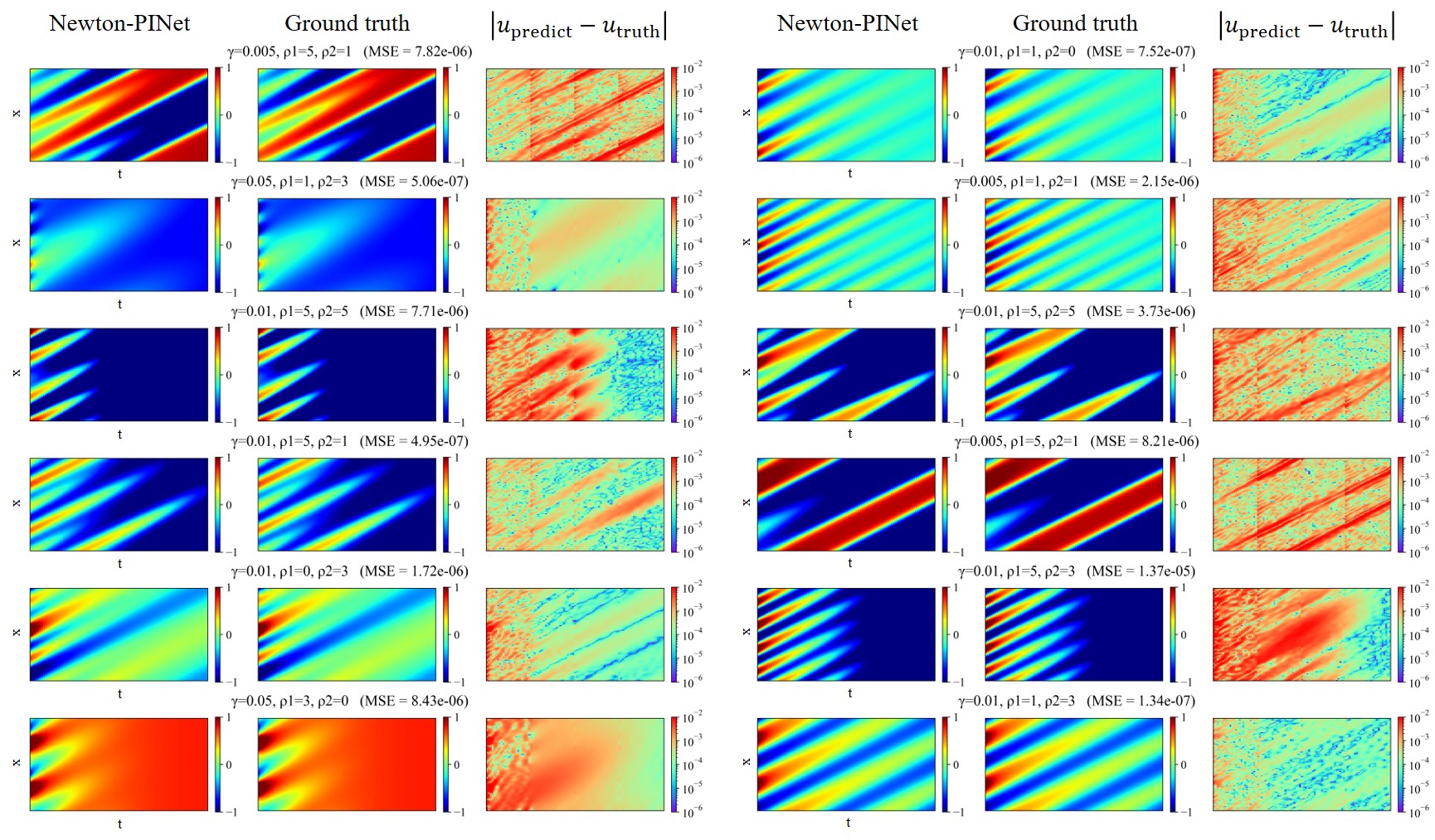}
    \caption{\label{fig:CDR-Visualization plot}Prediction results of Newton-PINet for the CDR problem.}
\end{center}
\end{figure}

\begin{figure}[h]
\begin{center}
    \includegraphics[width=1\linewidth]{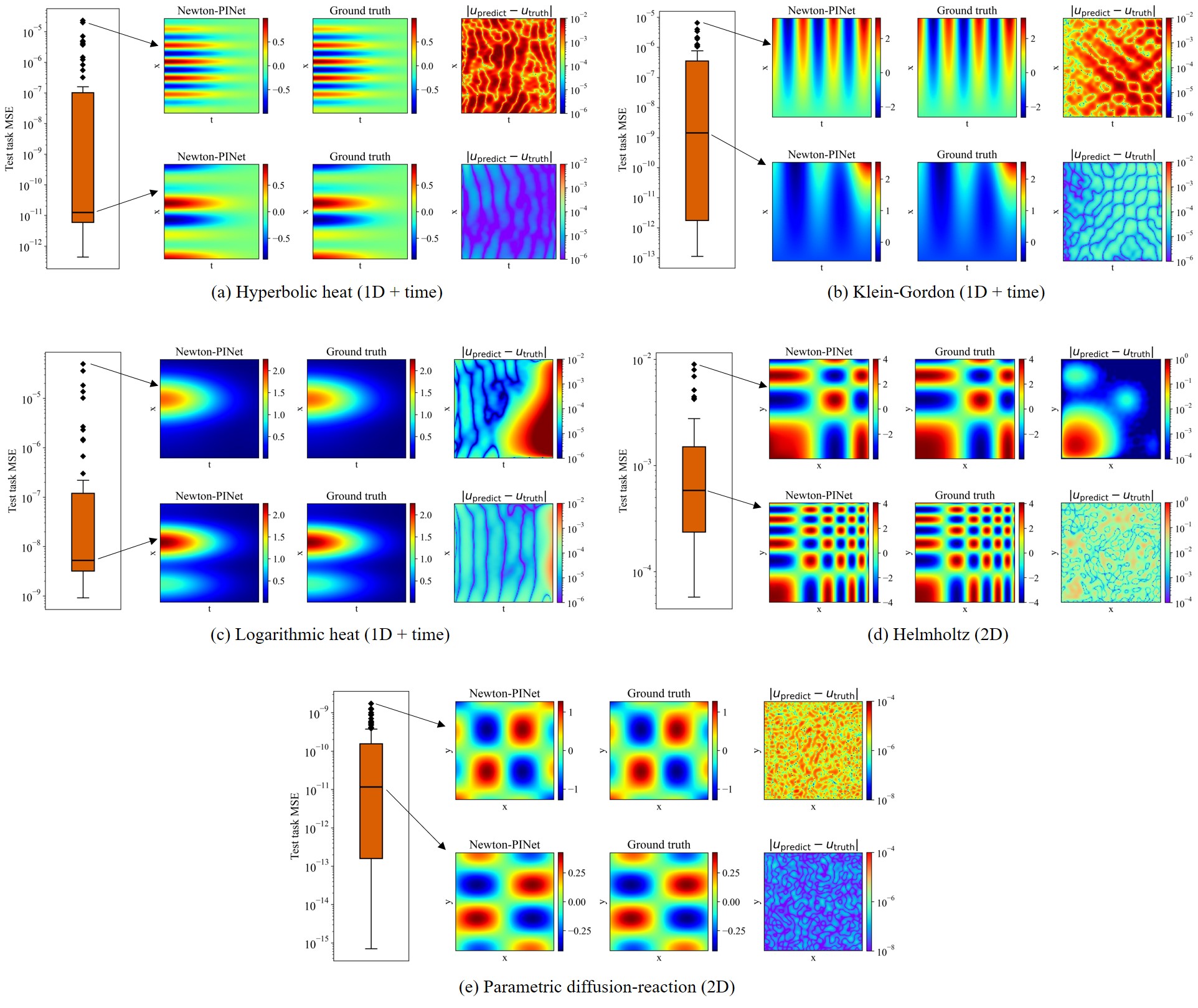}
    \caption{\label{fig:KG-Heat-DR-Visualization plot}Prediction results of Newton-PINet for the other 5 nonlinear PDE problems.}
\end{center}
\end{figure}

\begin{figure}[h]
    \centering
    \includegraphics[width=0.8\linewidth]{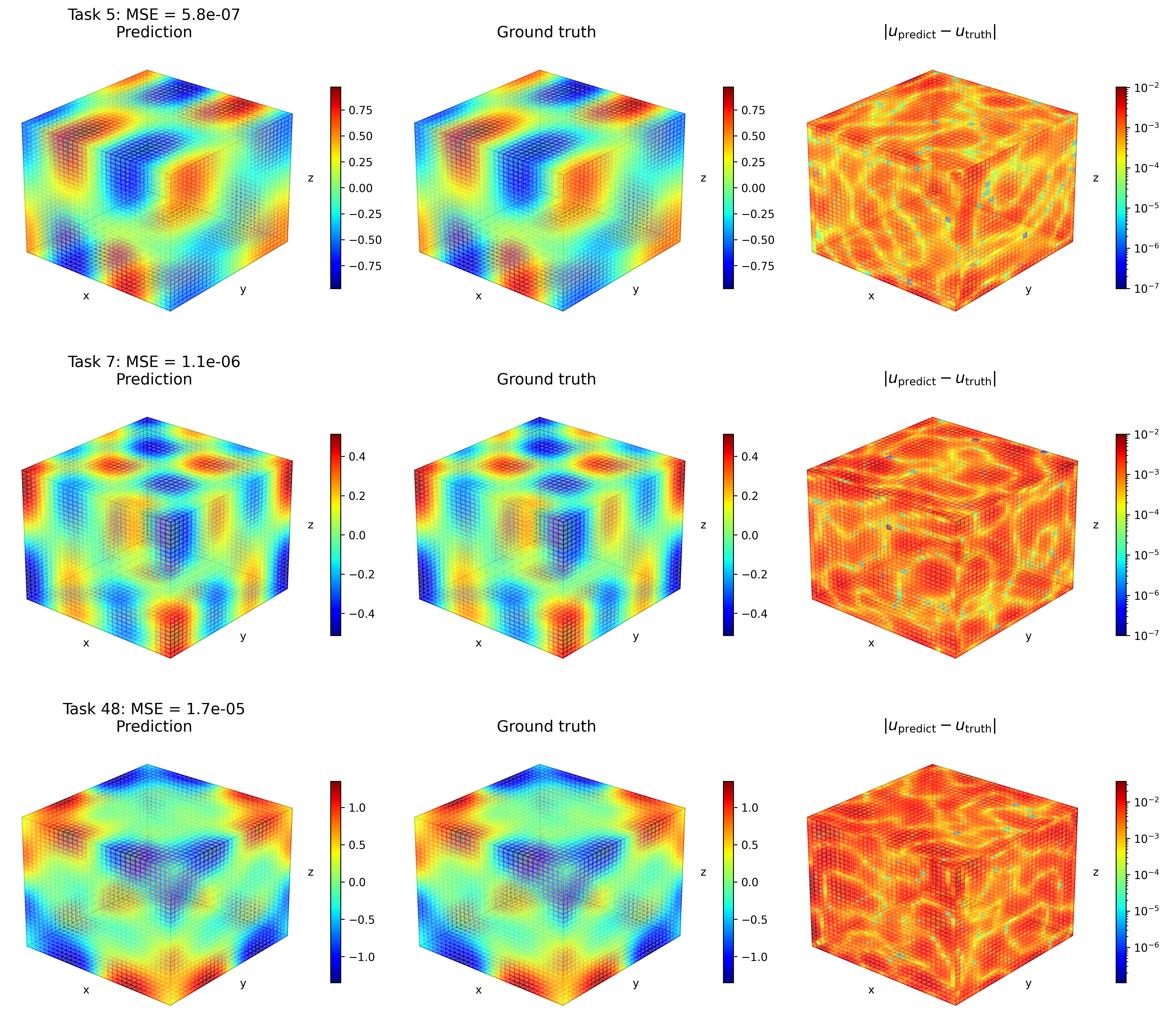}
    \caption{\label{fig:3D parametric diffusion-reaction}Prediction results of Newton-PINet for the 3D parametric diffusion-reaction problem.}
\end{figure}

\paragraph{Prediction results of nonlinear forcing-type PDEs:} 
Figures~\ref{fig:CDR-Visualization plot}, ~\ref{fig:KG-Heat-DR-Visualization plot}, ~\ref{fig:3D parametric diffusion-reaction} show the prediction results of Newton-PINet on some test cases for the convection-diffusion-reaction (CDR) and other nonlinear forcing-type PDEs described in Appendix~\ref{app:PDE_problems}.

\paragraph{Model generalization to unseen boundary conditions:}
We first investigate the generalization capability of Newton-PINet to unseen boundary condition (BC) based on the 1D+time hyperbolic heat problem defined in Eq.~(\ref{eq:1D+time hyperbolic heat}). The model is meta-trained on 16 tasks with \emph{Neumann} boundary conditions and subsequently evaluated on 64 unseen tasks with \emph{Dirichlet} and mixed BC. As shown in Table~\ref{tab:heat_hyperbolic_BC_generalization}, the Newton-PINet model can effectively generalize to these out-of-distribution boundary conditions that were not encountered during meta-training.

\begin{table*}[htbp]

    \caption{Model generalization to unseen boundary condition (BC) for the 1D+time hyperbolic heat problem.} 

    \label{tab:heat_hyperbolic_BC_generalization} 

    \begin{center} 

    \setlength{\tabcolsep}{8pt} 
    \begin{tabular}{lcc}
    & Dirichlet BC & Mixed BC \\
    \toprule
    Inference time (s) & 0.067 & 0.071 \\
    MSE & $(1.42 \pm 4.54) \times 10^{-5}$ & $(1.49 \pm 4.48) \times 10^{-5}$ \\
    Relative $L^2$ & $(5.11 \pm 11.30) \times 10^{-3}$ & $(5.47 \pm 11.30) \times 10^{-3}$ \\
    \bottomrule
    \end{tabular}
    \end{center} 

\end{table*}

\paragraph{Model generalization to unseen irregular geometries:}
We further evaluate the generalization capability of Newton-PINet on unseen irregular geometries using two PDE problems, including the 1D+time hyperbolic heat equation and the 3D parametric diffusion-reaction equation.

\textbf{(1) 1D+time hyperbolic heat problem}

For the hyperbolic heat equation defined in Eq.~(\ref{eq:1D+time hyperbolic heat}), the model is meta-trained on regular geometry with 16 different PDE parameter configurations and evaluated on 64 unseen test cases with irregular curved geometries and simultaneously varying PDE parameters. During testing, the spatial domain is replaced by a time-dependent curved geometry with symmetric moving boundaries:
\begin{equation}
x_l(t)=-1-A\sin(2\pi f t), \qquad
x_u(t)=1+A\sin(2\pi f t),
\end{equation}
where the deformation amplitude and frequency are randomly sampled from $A\in[0.1,0.5]$ and $f\in[1.0,5.5]$, respectively.

\begin{figure}[h]
\centering
\includegraphics[width=0.8\linewidth]{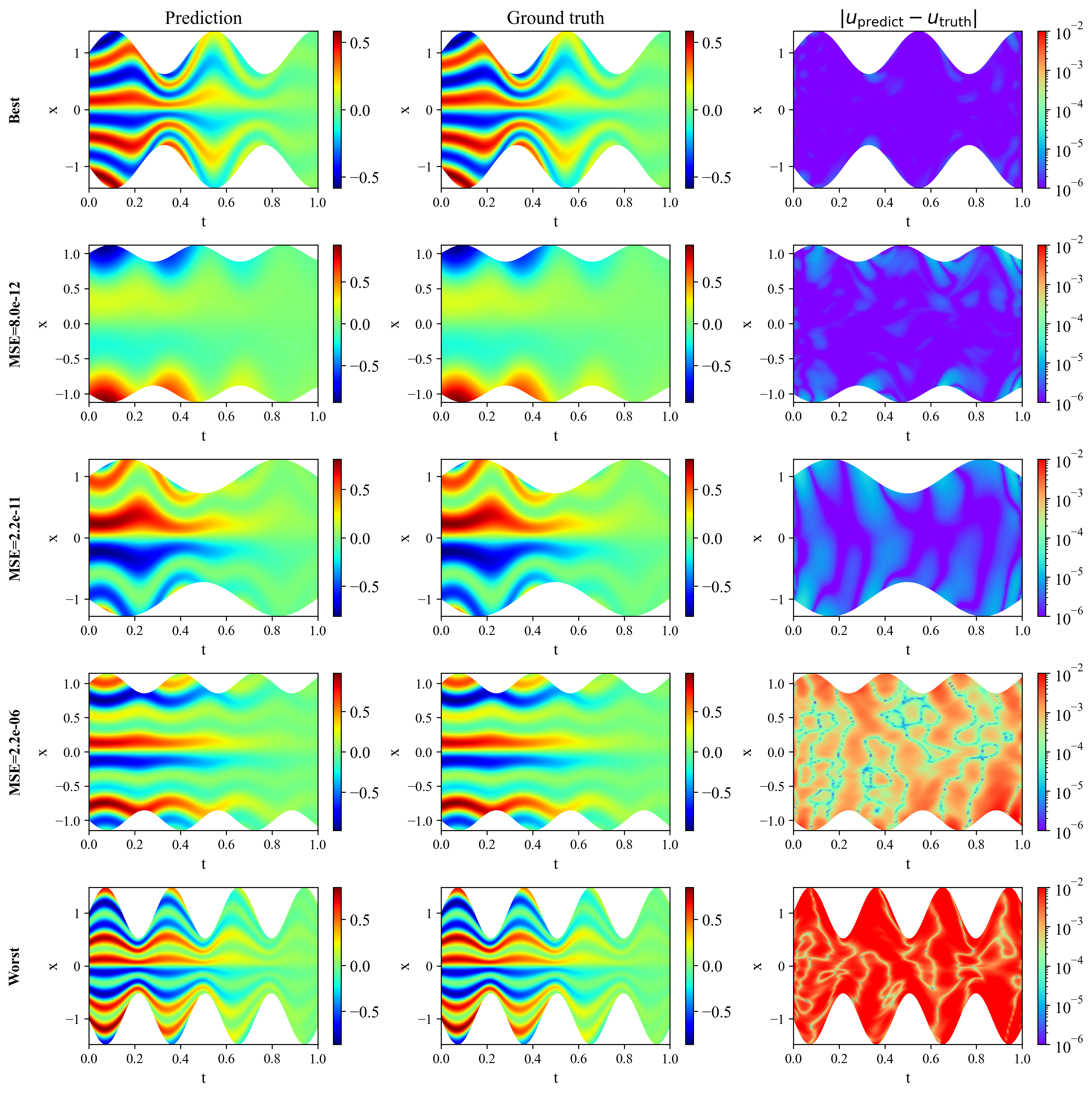}
\caption{\label{fig:Hyperbolic Heat problem across varying domain shapes}Prediction results of Newton-PINet for the hyperbolic heat problem across varying domain shapes and PDE parameters, with meta-training performed on regular geometry.}
\end{figure}

The adaptation is performed solely through physics-informed least-squares fine-tuning of the output layer, without retraining the network parameters. The prediction results are shown in Table~\ref{tab:heat_hyperbolic_shape_param} and Figure~\ref{fig:Hyperbolic Heat problem across varying domain shapes}.

\begin{table*}[htbp]

    \caption{Generalization to unseen irregular geometries and parameters for the 1D+time hyperbolic heat problem.}
    \label{tab:heat_hyperbolic_shape_param}
    \begin{center}
    \setlength{\tabcolsep}{8pt}
    \begin{tabular}{lc}
    Metric & Value \\ \toprule
    Inference time (s) & 0.074 \\
    MSE & $(1.50 \pm 3.23) \times 10^{-5}$ \\
    Relative $L^2$ & $(6.12 \pm 11.40) \times 10^{-3}$ \\ \bottomrule
    \end{tabular}
    \end{center}
    
\end{table*}

\textbf{(2) 3D parametric diffusion-reaction problem}

We further examine the generalization capability of Newton-PINet using the 3D parametric diffusion-reaction problem (Eq.~(\ref{eq:3D Parametric diffusion-reaction})). The model is meta-trained on regular geometry with 17 different PDE parameter configurations and evaluated on 100 unseen test cases with irregular domains of diverse shapes and simultaneously varying PDE parameters. The irregular geometries are generated through a spherical-domain perturbation procedure. Specifically, sampling points are defined by radius $r\in[0.05,1.0]$, azimuth $\theta\in[0,2\pi)$, and polar angle $\phi\in[0.05,\pi-0.05]$ to avoid polar singularities. A perturbation scale is then defined as
\begin{equation}
\mathrm{scale}
=1.0+A\cos(k_\theta\theta+\varphi_\theta)
\sin(k_\phi\phi+\varphi_\phi),
\end{equation}
where $A\in[0.1,0.3]$ controls the perturbation amplitude, $k_\theta,k_\phi\in{2,3,4,5}$ determine the spatial frequencies, and $\varphi_\theta,\varphi_\phi$ are randomly sampled phases. The perturbed points are mapped to Cartesian coordinates according to
\begin{equation}
    \left\{
    \begin{aligned}
    x &= r\cdot\mathrm{scale}\cdot\sin(\phi)\cos(\theta),\\
    y &= r\cdot\mathrm{scale}\cdot\sin(\phi)\sin(\theta),\\
    z &= r\cdot\mathrm{scale}\cdot\cos(\phi).
    \end{aligned}
    \right.
\end{equation}
The outermost shell ($r=1.0$) of the $32\times32\times32$ grid is extracted to impose the Dirichlet boundary conditions. This procedure preserves the topology of the original domain while introducing diverse irregular boundary shapes.

\begin{figure}[h]
\centering
\includegraphics[width=0.8\linewidth]{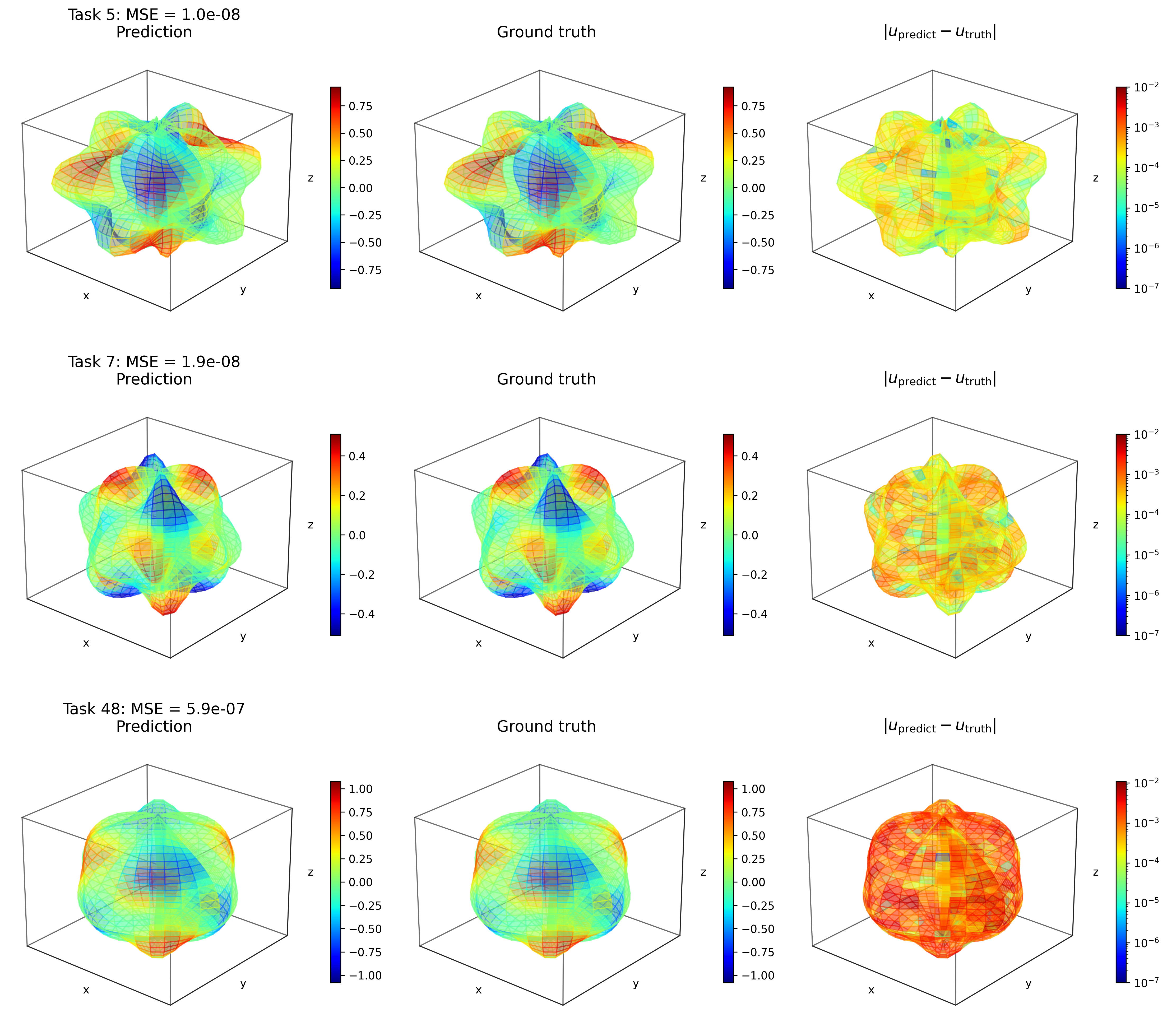}
\caption{\label{fig:3D parametric diffusion-reaction test}Prediction results of Newton-PINet for the 3D parametric diffusion-reaction equation across varying domain shapes and PDE parameters, with meta-training performed on regular geometry.}
\end{figure}

The test results are summarized in Table~\ref{tab:heat_hyperbolic_shape_param_3d}, with representative predictions shown in Figure~\ref{fig:3D parametric diffusion-reaction test}. 

\begin{table*}[htbp]

    \caption{Generalization to unseen irregular geometries and parameters for the 3D parametric diffusion-reaction problem.}
    \label{tab:heat_hyperbolic_shape_param_3d}
    \begin{center}
    \setlength{\tabcolsep}{8pt}
    \begin{tabular}{lc}
    Metric & Value \\ \toprule
    Inference time (s) & 0.104 \\
    MSE & $(5.56 \pm 7.39) \times 10^{-6}$ \\
    Relative $L^2$ & $(9.21 \pm 4.09) \times 10^{-3}$ \\ \bottomrule
    \end{tabular}
    \end{center}
\end{table*}

Overall, these results demonstrate that Newton-PINet can generalize to unseen boundary conditions and irregular domain geometries while maintaining high prediction accuracy and computational efficiency. Notably, the adaptation to these out-of-distribution settings requires only physics-informed fine-tuning of the output layer, without retraining the network parameters.

\section{Comprehensive comparison of model performance on nonlinear benchmark problems} 
\label{app:summary of model comparations}

\subsection{Model comparison under varying initial conditions on Burgers' problem} 
\label{app:Model comparison under varying initial conditions}

We evaluate the model on four settings with $\gamma \in \{0.1,0.01,0.001,0.0001\}$ in Burgers' problem, where, for each $\gamma$, the initial conditions are sampled from a Gaussian random field $\mathcal{N}(0,625(-\Delta+25I)^{-4})$ to generate 50 training and 1000 test tasks. Experiments are conducted independently for each $\gamma$. We compare (i) \textbf{unsupervised Newton-PINet} (1 time block) and (ii) \textbf{hybrid Newton-PINet} (4 time blocks) against popular neural operator baselines in terms of training time and test error. The \textbf{unsupervised} operator considered is Physics-Only DeepONet (PO-DeepONet)~\citep{wang2021learning}. The \textbf{supervised} operators include DeepONet~\citep{lu2021learning}, Transformer-DeepONet with Trunk net enhanced by Fourier coefficients (T-DeepONet-TF)~\citep{wei2025efficient}, and FNO~\citep{li2020fourier}, while the \textbf{hybrid} operator is PINO (a physics-informed FNO)~\citep{li2024physics}. Note that PINO uses a training loss that combines both data-driven and PDE-based physics losses. The results are summarized in Table~\ref{tab:Burgers-Compar}. Compared to PO-DeepONet, unsupervised Newton-PINet reduces training time from 9.25h to 291s while improving test accuracy across all $\gamma$ experiments (in a data-absent scenario). Compared to supervised and hybrid neural operators, including the state-of-the-art T-DeepONet-TF and PINO, Newton-PINet consistently achieves superior test accuracy under $\gamma=0.1, 0.01, 0.0001$, with significantly reduced training cost (10x less time). Our model consistently achieves higher accuracy while requiring the least training time, whether fully unsupervised (without data) or with a few labeled samples.

\begin{table*}[h]
    \caption{Model comparison on the 1D Burgers' problem. The initial conditions for generating all the training and test data are drawn from $\mathrm{GRF}\sim\mathcal{N}(0,625(-\Delta+25I)^{-4})$. The lowest errors are highlighted in bold. “--” denotes results not reported in the references.}
    \label{tab:Burgers-Compar}
    \begin{center}
    \begin{small}
    \setlength{\tabcolsep}{2pt} 
    \begin{tabular}{llc cccc}
        & \multicolumn{1}{l}{\multirow[b]{2}{*}{Model}}  
        & \multirow[b]{2}{*}{\begin{tabular}[b]{@{}c@{}}No. training\\ tasks\end{tabular}} 

        & \multicolumn{4}{c}{\begin{tabular}[b]{@{}c@{}}Test relative $L^2$ error (training time)\end{tabular}} \\ \cmidrule{4-7}
        \multicolumn{4}{l}{}  $\gamma=0.1$ & $\gamma=0.01$ & $\gamma=0.001$ & $\gamma=0.0001$ \\ \toprule
    \multirow{4}{*}{Unsupervised} 
    & \multirow{2}{*}{\begin{tabular}[c]{@{}l@{}}PO-DeepONet\\ \scriptsize{\citep{wang2021learning}}\end{tabular}}      
    & \multirow{2}{*}{1000} 
    &   \multirow{2}{*}{--}                   & 1.38e-2 & 2.16e-1 & 2.48e-1 \\ 
    &                                   &            
    &                      & (9.25h)  &   --      &   --      \\ \cmidrule{2-7}
    
    & \multirow{2}{*}{\begin{tabular}[c]{@{}l@{}}\textbf{Newton-PINet}\\ \scriptsize{(Ours)}\end{tabular}} 
    & \multirow{2}{*}{50} 
    & 1.33e-3 & 3.51e-3 & 1.37e-1 & 2.01e-1 \\ 
    &                                   &             
    & (291s)  & (291s)  & (291s)  & (291s)  \\  \midrule \midrule
    
    \multirow{5}{*}{Supervised}   
    & \multirow{2}{*}{\begin{tabular}[c]{@{}l@{}}DeepONet\\ \scriptsize{\citep{wei2025efficient}}\end{tabular}}         
    & \multirow{2}{*}{1000} 
    &    \multirow{2}{*}{--}                  & 1.17e-2 & 2.30e-1 & 2.88e-1 \\ 
    &                                   &            
    &                      & (2800s)  & (2620s) & (2660s) \\ \cmidrule{2-7}
    
    & \multirow{2}{*}{\begin{tabular}[c]{@{}l@{}}T-DeepONet-TF\\ \scriptsize{\citep{wei2025efficient}}\end{tabular}}      
    & \multirow{2}{*}{1000} 
    &   \multirow{2}{*}{--}                   & 2.08e-3 & \textbf{9.81e-3} & 1.18e-1 \\ 
    &                                   &             
    &                      & (3041s) & (2569s) & (3466s) \\ \cmidrule{2-7}
    
    & FNO~\scriptsize{\citep{li2020fourier}}    
    & 1000 
    & 1.39e-2 & --  &  -- & --  \\ \midrule
    
    \multirow{4}{*}{Hybrid}       
    & \multirow{2}{*}{\begin{tabular}[c]{@{}l@{}}PINO\\ \scriptsize{\citep{li2024physics}}\end{tabular}}             
    & \multirow{2}{*}{1000}  
    &    \multirow{2}{*}{--}                  & 3.80e-3 & \multirow{2}{*}{--}  & \multirow{2}{*}{--}  \\ 
    &                                   &             
    &                      & (1200s)  &   &   \\ \cmidrule{2-7}
    
    & \multirow{2}{*}{\begin{tabular}[c]{@{}l@{}}\textbf{Newton-PINet}\\ \scriptsize{(Ours)}\end{tabular}} 
    & \multirow{2}{*}{50} 
    & \textbf{1.13e-3} & \textbf{5.46e-4} & 4.66e-2 & \textbf{9.15e-2} \\ 
    &                                   &             
    & (133s) & (133s) & (133s) & (133s) \\ \bottomrule
    \end{tabular}
    \end{small}
    \end{center}
    \vskip -0.1in
\end{table*}

\subsection{Model comparison under varying PDE parameters on a nonlinear reaction-diffusion problem} 
\label{app:Model comparison under varying PDE parameters on a nonlinear reaction-diffusion problem}

We further evaluate our method on a widely studied 1D nonlinear reaction-diffusion problem $\frac{\partial u}{\partial t} - \gamma \frac{\partial^2 u}{\partial x^2} - \rho u(1-u)= 0$~\citep{krishnapriyan2021characterizing}. The benchmark configuration and baseline model results are all taken from~\citep{boudec2025learning}. The benchmark uses a fixed Gaussian initial condition, with PDE parameters sampled over $\gamma \in [1,5]$ and $\rho \in [-5,5]$. 
The baseline models are grouped into two categories: (i) \textbf{unsupervised}, including instance-wise PINNs trained with L-BFGS (PINNs+L-BFGS) or Adam+L-BFGS (PINNs-multi-opt), parametric PINNs (PPINNs)~\citep{boudec2025learning} and P$^2$INNs~\citep{cho2024parameterized} that incorporate PDE parameters as inputs, and Physics-Only DeepONet (PO-DeepONet)~\citep{wang2021learning}; and (ii) \textbf{hybrid}, which combine supervised and physics-informed approaches, such as PI-DeepONet~\citep{goswami2023physics}, PINO~\citep{li2024physics}, and the physics-informed neural solver (PI-neural-solver)~\citep{boudec2025learning}. These baselines were trained with 800 tasks and tested on 200 unseen tasks. It can be observed that, among the baseline models, purely physics-based unsupervised training tends to struggle with highly nonlinear problems, whereas hybrid approaches leverage large amounts of labeled data to improve representation learning, thereby enhancing test accuracy. In contrast, our hybrid \textbf{Newton-PINet} \textit{achieves state-of-the-art performance using only 50 training tasks}, i.e., \emph{16$\times$ fewer tasks}, and is evaluated on the same test set using a 4-time-block decomposition. For a fair comparison, both models are trained and evaluated on an NVIDIA A6000 GPU. As shown in Table~\ref{tab:Reaction_diffusion_model_compare}, Newton-PINet reaches a test MSE roughly three orders of magnitude lower than the best-performing baseline (PI-neural-solver) while requiring only 301 seconds of total meta-training wall-clock time, compared with several hours for the baselines. These results highlight the \textit{high data efficiency and computational efficiency} of Newton-PINet. Model prediction results are provided in Appendix Figure~\ref{fig:ICLR_DR-Visualization plot}.

\begin{table*}[htbp]
\caption{Model comparison on the 1D nonlinear reaction-diffusion problem. Baseline results are from \cite{boudec2025learning}. Time is reported in hours (h), minutes (m), and seconds (s). Best and second-best are bold and underlined, respectively.}
\label{tab:Reaction_diffusion_model_compare}
\begin{center}
\begin{small}
\setlength{\tabcolsep}{3pt} 
\begin{tabular}{llc ccc}
                              &\multicolumn{1}{l}{Model}                     & \begin{tabular}[c]{@{}c@{}}No. training\\ tasks\end{tabular} & \begin{tabular}[c]{@{}c@{}}Test relative\\ MSE\end{tabular} & \begin{tabular}[c]{@{}c@{}}Training\\time\end{tabular} & \begin{tabular}[c]{@{}c@{}}Inference\\time\end{tabular} \\ \toprule

\multirow{5}{*}{Unsupervised} & PINNs+L-BFGS        & \multirow{5}{*}{\begin{tabular}[c]{@{}c@{}}800 \end{tabular}} & 6.13e-1                              & --                                 & 369s                              \\
                              & PINNs-multi-opt     &                                         & 7.57e-1                              & --                                 & 16.5s                             \\
                              & PPINNs              &                                         & 3.94e-1                              & 4h15m                          & 0.291s                            \\
                              & P$^2$INNs              &                                         & 5.69e-1                              & 11h                              & 0.676s                            \\
                              & PO-DeepONet         &                                         & 4.10e-1                              & 3h30m                          & 0.438s                            \\ \midrule
\multirow{4}{*}{Hybrid}       & PI-DeepONet         & \multirow{3}{*}{\begin{tabular}[c]{@{}c@{}}800 \end{tabular}} & 7.90e-2                              & 3h30m                          & 0.443s                            \\
                              & PINO                &                                         & 4.21e-4                              & \underline{1h10m}               & 0.519s                            \\
                              & PI-neural-solver    &                                         & \underline{2.91e-4}                  & 4h30m                          & \underline{0.284s}                \\
                              & \textbf{Newton-PINet} (Ours) & \begin{tabular}[c]{@{}c@{}}50 \end{tabular}                       & \textbf{1.72e-7}                     & \textbf{301s}                   & \textbf{0.080s }                  \\ \bottomrule
\end{tabular}
\end{small}
\end{center}
\vskip -0.1in
\end{table*}

\begin{figure}[h]
\begin{center}
    \includegraphics[width=1\linewidth]{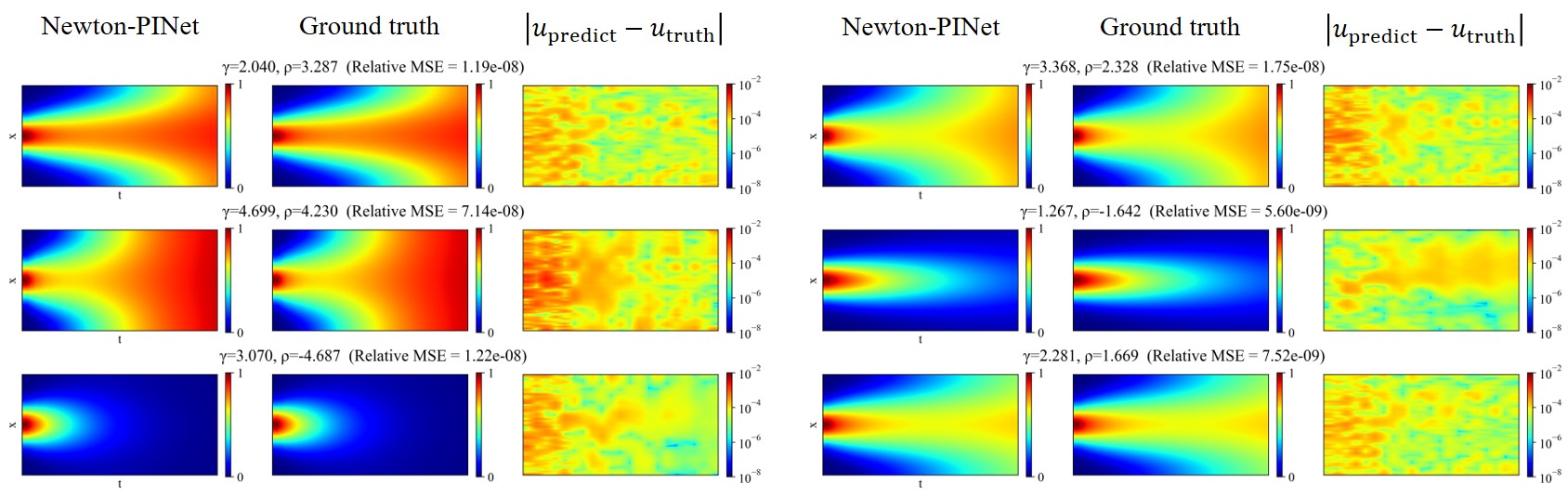}
    \caption{\label{fig:ICLR_DR-Visualization plot}Prediction results of Newton-PINet for the 1D nonlinear reaction-diffusion benchmark problem.}
\end{center}
\end{figure}

\subsection{Model comparison under different GPU configurations} 
\label{app:Model comparison under different GPU configurations}

Following the problem setup in Appendix~\ref{app:Model comparison under varying PDE parameters on a nonlinear reaction-diffusion problem}, Table~\ref{tab:differ-GPU-Reaction_diffusion_model_compare} further reports the model performance under different GPU configurations. It can be observed that Newton-PINet consistently outperforms PI-neural-solver on both A6000 and V100 GPUs. 

In addition, a comparison of Newton-PINet on A6000 and V100 GPUs shows that training time differs across the two GPUs, while inference time remains comparable. This is likely because GPU configurations mainly affect the efficiency of gradient-based training, whereas our gradient-free Tikhonov-based adaptation is much less sensitive to such differences.

\begin{table*}[htbp]
\caption{Comparison of model performance under different GPU configurations.}
\label{tab:differ-GPU-Reaction_diffusion_model_compare}
\begin{center}
\setlength{\tabcolsep}{3pt} 
\begin{tabular}{llc ccc}
                              &\multicolumn{1}{l}{Model}                     & \begin{tabular}[c]{@{}c@{}}No. training\\tasks\end{tabular} & \begin{tabular}[c]{@{}c@{}}Test relative\\ MSE\end{tabular} & \begin{tabular}[c]{@{}c@{}}Training\\time\end{tabular} & \begin{tabular}[c]{@{}c@{}}Inference\\time\end{tabular} \\ \toprule

                              & PI-neural-solver (A6000-GPU)    &  \begin{tabular}[c]{@{}c@{}}800 \end{tabular}                                                              & 2.91e-4                 & 4h30m                          & 0.284s               \\
                              & \textbf{Newton-PINet} (A6000-GPU) & \begin{tabular}[c]{@{}c@{}}50 \end{tabular}                       & 1.72e-7                     & 301s                  & 0.080s                  \\ 
                              & \textbf{Newton-PINet} (V100-GPU) & \begin{tabular}[c]{@{}c@{}}50 \end{tabular}                       & 1.71e-7                     & 119s                   & 0.084s                  \\ \bottomrule
\end{tabular}

\end{center}

\end{table*}

\subsection{Model comparison with other meta-learning PINN methods} 
\label{app:Model comparison with other meta-learning PINN methods}

We further compare the performance of Newton-PINet with state-of-the-art gradient-based meta-learning PINNs \citep{penwarden2023metalearning} and Baldwinian-PINNs \citep{wong2026evolutionary}. For a fair comparison, we train an unsupervised Newton-PINet with a meta-learning loss of LSE and a hybrid Newton-PINet with a meta-learning loss of MSE, considering a single time-block setting. Table~\ref{tab:Baldwin-PINN-Meta-PINN-compare} summarizes the generalization performance and computational cost of Newton-PINet relative to these meta-learned PINN models. Compared to unsupervised gradient-based meta-PINNs, the unsupervised Newton-PINet consistently achieves lower test errors and shorter inference times. For instance, on a parametric diffusion-reaction problem, Newton-PINet exhibits approximately a 179× improvement in generalization accuracy while requiring 7000× less computational time. Against gradient-free Baldwinian-PINN, which employs Picard to handle PDE nonlinear terms, Newton-PINet which employs the Newton linearization method reduces test MSE by 2$\sim$5 orders of magnitude across the benchmark problems and shortens single-task fine-tuning time by roughly one order of magnitude. 

We also note that, in some cases, the unsupervised Newton-PINet achieves slightly higher accuracy than the hybrid Newton-PINet. This occurs for relatively simple nonlinear PDEs, where using an LSE-based meta-learning loss facilitates better convergence and generalization than MSE. Conversely, for more complex nonlinear PDEs, such as the KdV and K-S equations, adopting an MSE-based meta-learning loss leads to much better performance.

\begin{table}[t]
\caption{Model comparison on nonlinear benchmark problems. Baselines are reported from~\cite{penwarden2023metalearning,wong2026evolutionary}. “--” denotes results not reported in the references.}
\footnotesize 
\label{tab:Baldwin-PINN-Meta-PINN-compare}
\begin{center}
\begin{threeparttable}
\begin{tabular}{lllcccc}
\multicolumn{3}{l}{\multirow[b]{3}{*}{Meta-learning PINNs}} & \multicolumn{4}{c}{Nonlinear benchmark problems (number of test tasks)} \\  
\cmidrule(lr){4-7}
\multicolumn{3}{l}{} & \begin{tabular}[c]{@{}c@{}}Heat\\ (64)\end{tabular} 
& \begin{tabular}[c]{@{}c@{}}Allen-Cahn\\ (32)\end{tabular} 
& \begin{tabular}[c]{@{}c@{}}Diffusion\\ -reaction (64)\end{tabular} 
& \begin{tabular}[c]{@{}c@{}}Parametric diffusion\\ -reaction (100)\end{tabular} \\  
\toprule

\multirow{14}{*}{\rotatebox{90}{Unsupervised}} 
& \multirow{2}{*}{Random}     & Error\textsuperscript{1} & 0.0052 & 0.015 & 0.011 & 0.0022 \\
&                             & Time\textsuperscript{2}       & 156    & 496   & 1073  & 612    \\ \cmidrule(lr){2-7}

& \multirow{2}{*}{MAML}       & Error\textsuperscript{1} & 0.0045 & --     & --     & --      \\
&                             & Time\textsuperscript{2}       & 188    & --     & --     & --      \\ \cmidrule(lr){2-7}

& \multirow{2}{*}{Center}     & Error\textsuperscript{1} & 0.0045 & 0.012 & 0.0095 & 0.0018 \\
&                             & Time\textsuperscript{2}       & 96     & 201   & 426    & 494    \\ \cmidrule(lr){2-7}

& \multirow{2}{*}{Multitask}  & Error\textsuperscript{1} & 0.0048 & 0.012 & 0.009  & 0.0017 \\
&                             & Time\textsuperscript{2}       & 49     & 120   & 243    & 431    \\ \cmidrule(lr){2-7}

& \multirow{2}{*}{LMC}        & Error\textsuperscript{1} & 0.0049 & 0.011 & 0.0091 & 0.0015 \\
&                             & Time\textsuperscript{2}       & 59     & 120   & 302    & 428    \\ \cmidrule(lr){2-7}

& \multirow{2}{*}{RBF}        & Error\textsuperscript{1} & 0.0044 & 0.012 & 0.0081 & 0.0017 \\
&                             & Time\textsuperscript{2}       & 35     & 68    & 280    & 375    \\ \cmidrule(lr){2-7}

& \multirow{2}{*}{Polynomial} & Error\textsuperscript{1} & 0.0046 & 0.012 & 0.0085 & 0.0018 \\
&                             & Time\textsuperscript{2}       & 38     & 44    & 249    & 496    \\ \cmidrule(lr){2-7}

& \multirow{2}{*}{\begin{tabular}[c]{@{}c@{}}\textbf{Newton-PINet}\\(Ours)\end{tabular} } 
                              & Rel. $L^2$ & 4.0e-6 & 4.2e-7 & 1.4e-4 & 9.5e-6 \\
&                             & Time\textsuperscript{3}       & 0.06   & 0.044  & 0.047  & 0.049  \\ \midrule

\multirow{6}{*}{\rotatebox{90}{Hybrid}} 
& \multirow{3}{*}{Baldwinian-PINN} 
                              & Rel. $L^2$ & 6.0e-4 & 9.5e-4 & 1.3e-4 & 1.9e-4 \\
&                             & MSE                           & 1.3e-7 & 2.0e-7 & 1.2e-8 & 1.6e-8 \\
&                             & Time\textsuperscript{3}       & 0.87   & 1.19   & 0.58   & 0.6    \\ \cmidrule(lr){2-7}

& \multirow{3}{*}{\begin{tabular}[c]{@{}c@{}}\textbf{Newton-PINet}\\(Ours)\end{tabular} }
                              & Rel. $L^2$ & 7.3e-6 & 3.7e-6 & 1.4e-4 & 1.3e-5 \\
&                             & MSE                           & 1.7e-11 & 2.5e-12 & 6.8e-10 & 1.6e-10 \\
&                             & Time\textsuperscript{3}       & 0.062  & 0.044  & 0.047  & 0.047  \\ 
\bottomrule
\end{tabular}

\vspace{2mm}
\begin{tablenotes}
\footnotesize
\item[1] Relative $L^2$ errors obtained after 500 iterations of fine-tuning using ADAM followed by L-BFGS, as described in~\cite{penwarden2023metalearning}.  
\item[2] Only the L-BFGS optimization time (s) is reported in~\cite{penwarden2023metalearning}.  
\item[3] Mean inference time per test task (s), on a Tesla V100 GPU.  
\end{tablenotes}

\end{threeparttable}
\end{center}
\end{table}

\subsection{Model comparison with other neural operators on the newly introduced datasets} 
\label{app:Model comparison with neural operators on the newly introduced datasets}

In the above experiments, we validated the performance of our model on several previously reported challenging datasets. We next further evaluate our method on the newly introduced and more complex PDE benchmarks in this study, including the generalized KdV, Kuramoto-Sivashinsky (K-S), convection-diffusion-reaction (CDR), and the lid-driven cavity (LDC) nonlinear problems. Specifically, we compare the proposed hybrid Newton-PINet with several state-of-the-art neural operator methods, including DeepONet~\citep{lu2021learning}, Physics-Only DeepONet (PO-DeepONet)~\citep{wang2021learning}, Multi-task DeepONet (MT-DeepONet)~\citep{kumar2025synergistic}, FNO~\citep{li2020fourier}, PINO~\citep{li2024physics}, and Tucker-FNO~\citep{zhou2026tucker}. The hyperparameter settings of these baseline models are summarized in Table~\ref{tab:baseline model hyperparameters}. For each problem, all baseline models and our method are trained and evaluated using the same training and test datasets under identical NVIDIA V100 GPU configurations. The comparison results are presented in Table~\ref{tab:new-data model_comparison}. It can be observed that our model maintains a moderate parameter scale while consistently achieving the lowest test error across all considered benchmarks. Moreover, for the generalized KdV, CDR, and LDC problems, Newton-PINet requires substantially less training time than the competing methods. For the K-S problem, however, the optimization remains relatively challenging for our model, leading to a comparatively longer training time. Overall, on these newly introduced datasets, existing state-of-the-art neural operator methods exhibit relatively poor generalization performance under limited training data settings, primarily because they lack physics-based fine-tuning mechanisms and their generalization capability heavily relies on large amounts of training data. In contrast, the proposed Newton-PINet is able to maintain high generalization accuracy even in data-sparse regimes.

\begin{table*}[t]
\centering
\caption{Summary of baseline model architectures and hyperparameter configurations.}
\label{tab:baseline model hyperparameters}
\resizebox{\textwidth}{!}{
\begin{tabular}{@{}llcccl@{}}
\toprule
Model & Network architecture & Activation & Initialization & Optimizer / LR & Special components \\ \midrule

\multirow{2}{*}{DeepONet} 
& Branch: 4 layers $\times$ 100 units & \multirow{2}{*}{tanh} & \multirow{2}{*}{He uniform} & Adam, Cosine decay & \multirow{2}{*}{Learnable global bias} \\
& Trunk: 4 layers $\times$ 100 units & & & $10^{-3} \to 10^{-6}$ & \\ \midrule

\multirow{2}{*}{PO-DeepONet} 
& Branch: 4 layers $\times$ 100 units & \multirow{2}{*}{tanh} & \multirow{2}{*}{Glorot normal} & Adam, Cosine decay & \multirow{2}{*}{AD-based physics residual} \\
& Trunk: 4 layers $\times$ 100 units & & & $10^{-3} \to 10^{-6}$ & \\ \midrule

\multirow{2}{*}{MT-DeepONet} 
& Branch: [100, 100] $\to$ 100 (latent) & ReLU & \multirow{2}{*}{Glorot normal} & Adam, Cosine decay & Layer normalization \\
& Trunk: [100, 100, 100] $\to$ 100 (latent) & Leaky ReLU & & $10^{-3} \to 10^{-6}$ & No activation on final layer \\ \midrule

\multirow{3}{*}{FNO} 
& \multirow{3}{*}{\begin{tabular}[c]{@{}l@{}}4 spectral conv blocks\\ width 32, modes 12 $\times$ 12\\ Projection: 32 $\to$ 128 $\to$ 1\end{tabular}} & \multirow{3}{*}{GELU} & \multirow{3}{*}{Uniform} & Adam, Cosine decay & Frequency truncation \\
& & & & $10^{-3} \to 10^{-6}$ & Complex multiplication \\
& & & & & Residual connections \\ \midrule

\multirow{3}{*}{PINO} 
& \multirow{3}{*}{\begin{tabular}[c]{@{}l@{}}4 spectral conv blocks\\ width 32, modes 12 $\times$ 12\\ Projection: 32 $\to$ 128 $\to$ 1\end{tabular}} & \multirow{3}{*}{GELU} & \multirow{3}{*}{Uniform} & Adam, Cosine decay & Spectral derivative \\
& & & & $10^{-3} \to 10^{-6}$ & Data + physics loss \\
& & & & & Residual connections \\ \midrule

\multirow{3}{*}{Tucker-FNO} 
& \multirow{3}{*}{\begin{tabular}[c]{@{}l@{}}4 Tucker-spectral blocks\\ width 32, modes 12, rank 32\\ Projection: 32 $\to$ 128 $\to$ 1\end{tabular}} & \multirow{3}{*}{GELU} & Normal / & Adam, Cosine decay & Tucker core tensor contraction \\
& & & Complex normal & $10^{-3} \to 10^{-6}$ & Residual connections \\
& & & & & \\ \bottomrule
\end{tabular}
}
\end{table*}

\begin{table*}[htbp]
\caption{Model comparison on the newly introduced benchmarks. Best and second-best results are highlighted in bold and underlined, respectively.}
\label{tab:new-data model_comparison}
\begin{center}
\begin{small}
\setlength{\tabcolsep}{4pt}
\begin{tabular}{llcccc}
\toprule
Problem & Model & \begin{tabular}[c]{@{}c@{}}No. model\\ parameters\end{tabular} & \begin{tabular}[c]{@{}c@{}}No. training / \\ test tasks\end{tabular} & \begin{tabular}[c]{@{}c@{}}Training\\ time (s)\end{tabular} & \begin{tabular}[c]{@{}c@{}}Test\\ MSE\end{tabular} \\
\midrule

\multirow[t]{7}{*}{Generalized KdV}
& DeepONet     & 87101   & \multirow{7}{*}{10 / 35} & 642  & 8.24e-2 \\
& PO-DeepONet  & 87101   &                         & 4437 & 5.01e-1 \\
& MT-DeepONet  & 77400   &                         & \underline{588} & 9.13e-2 \\
& FNO          & 2376321 &                         & 639  & \underline{6.12e-2} \\
& PINO         & 2376321 &                         & 657  & 9.63e-2 \\
& Tucker-FNO   & 1067137 &                         & 954  & 9.12e-2 \\
& \textbf{Newton-PINet} (Ours) & 198146 &                 & \textbf{453} & \textbf{1.94e-3} \\
\midrule

\multirow[t]{7}{*}{Kuramoto-Sivashinsky}
& DeepONet     & 74001   & \multirow{7}{*}{100 / 50} & \underline{916}  & 5.07 \\
& PO-DeepONet  & 74001   &                          & 2334 & 5.16e-1 \\
& MT-DeepONet  & 64300   &                          & \textbf{901} & 5.36 \\
& FNO          & 2372129 &                          & 934  & \underline{3.89e-1} \\
& PINO         & 2372129 &                          & 1438 & 4.39e-1 \\
& Tucker-FNO   & 1062945 &                          & 1726  & 4.26e-1 \\
& \textbf{Newton-PINet} (Ours) & 198146  &                          & 1014 & \textbf{3.45e-2} \\
\midrule

\multirow[t]{7}{*}{\vspace{1pt}\begin{tabular}[t]{@{}l@{}}Convection-diffusion\\reaction\end{tabular}}
& DeepONet     & 74401   & \multirow{7}{*}{40 / 160} & 1190 & 7.03e-1 \\
& PO-DeepONet  & 74401   &                          & 2587 & 2.18e-1 \\
& MT-DeepONet  & 64700   &                          & 1078 & 1.87e-1 \\
& FNO          & 2372257 &                          & 1572 & 1.55e-1 \\
& PINO         & 2372257 &                          & \underline{609} & \underline{1.42e-1} \\
& Tucker-FNO   & 1063073 &                          & 1179 & 2.03e-1 \\
& \textbf{Newton-PINet} (Ours) & 49922 &                  & \textbf{203} & \textbf{4.32e-5} \\
\midrule

\multirow[t]{7}{*}{Lid-driven cavity}
& DeepONet     & 61403   & \multirow{7}{*}{6 / 19} & \underline{660} & 1.17e-1 \\
& PO-DeepONet  & 61403   &                       & 2825 & 1.34 \\
& MT-DeepONet  & 57443   &                       & 681  & 8.44e-2 \\
& FNO          & 2368259 &                       & 1158 & \underline{9.98e-4} \\
& PINO         & 2368259 &                       & 1305 & 1.06e-1 \\
& Tucker-FNO   & 1059075 &                       & 1353 & 3.17 \\
& \textbf{Newton-PINet} (Ours) & 49922 &                & \textbf{290} & \textbf{1.67e-4} \\
\bottomrule

\end{tabular}
\end{small}
\end{center}
\vskip -0.1in
\end{table*}

\subsection{Runtime comparison with classical numerical solvers} 
\label{app:Runtime comparison with classical numerical solvers}

The goal is not to replace classical solvers for solving a single PDE instance, but to provide an efficient framework for \textit{offline meta-training followed by online real-time adaptation} for parameterized PDE families. Unlike classical methods, which solve each new PDE configuration from scratch, the pretrained Newton-PINet can directly adapt to new configurations through fast inference while achieving comparable accuracy to numerical solvers. 

For example, for the Burgers' equation, a standard Gauss-Newton solver requires approximately 23\,s for solving a new instance from scratch, whereas Newton-PINet completes adaptation within only 0.035\,s. For the CDR problem, the \texttt{Chebfun} spectral solver (\textsc{Matlab}) requires approximately $8\,\mathrm{s}$, compared with $0.1\,\mathrm{s}$ for Newton-PINet. For the 2D lid-driven cavity flow, a CFD solver requires more than 30\,s (excluding meshing), while Newton-PINet achieves prediction within 0.4\,s with an MSE of only $1.67\times10^{-4}$ compared with the CFD solution.

\section{Ablation studies}  
\label{app:Ablation_studies}

\subsection{Robustness of the meta-learned Tikhonov regularization parameter} 
\label{app:Robustness of the meta-learned Tikhonov regularization parameter}

In our method, the Tikhonov regularization parameter ($\lambda_{\mathrm{reg}}$) is not manually tuned but meta-learned, enabling the model to identify a stable and task-agnostic regularization level across all training tasks. Once learned, $\lambda_{\mathrm{reg}}$ remains fixed during test-time adaptation.

To evaluate its stability, using the Burgers' problem as an example, we additionally perform per-task optimization by initializing $\lambda_{\mathrm{reg}}$ from the meta-learned value and further updating it for each test task with 300 epochs of gradient-based updates. As shown in Table~\ref{tab:reg-param-ablation}, the meta-learned $\lambda_{\mathrm{reg}}$ remains close to the per-task optimized values across a wide range of test tasks, and their generalization performance differs only marginally. This indicates that the meta-learned hyperparameter exhibits strong stability and low sensitivity during test-time adaptation.

\begin{table}[h]
\caption{Sensitivity of the meta-learned $\lambda_{\mathrm{reg}}$ to task adaptation in test inference. For meta-learning, $\lambda_{\mathrm{reg}}$ is learned during meta-training and kept fixed during testing. For Per-task optimization, $\lambda_{\mathrm{reg}}$ is initialized from the meta-learned model parameters and then optimized for each test task through gradient-based updates.}
\label{tab:reg-param-ablation}
\begin{center}
\footnotesize 
\renewcommand{\arraystretch}{1.2} 
\setlength{\tabcolsep}{3pt}       
\begin{tabular}{llcccccc}

\multirow[b]{2}{*}{Method} & \multirow{2}{*}{} & \multicolumn{6}{c}{$\gamma$ (test tasks of Burgers’ problem)} \\
\cmidrule(lr){3-8}
                        &                   & 0.001 & 0.003 & 0.004 & 0.006 & … & 0.047  \\
\toprule
\multirow{2}{*}{Meta-learning} 
    & $\lambda_{\mathrm{reg}}$ & \multicolumn{6}{c}{7.57e-4} \\
    & MSE                      & 3.30e-5 & 2.07e-6 & 1.44e-6 & 4.59e-7 & … & 7.54e-8  \\
 
\midrule
\multirow{2}{*}{\begin{tabular}[c]{@{}l@{}}Per-task \\ optimization\end{tabular}}
    & $\lambda_{\mathrm{reg}}$ & 2.07e-5 & 4.34e-4 & 4.81e-4 & 3.26e-4 & … & 9.79e-5  \\
    & MSE                      & 3.21e-5 & 2.01e-6 & 1.44e-6 & 4.69e-7 & … & 7.07e-8  \\
\bottomrule
\end{tabular}
\end{center}
\end{table}

\subsection{Effect of initial guess in nonlinear iterations on model accuracy} 
\label{app:Effect of initial guess in nonlinear iterations}

We conduct empirical ablations to evaluate how the initial field guess ($u_{\mathrm{guess}}$) in nonlinear iterations influences model accuracy. Let's quantify the distance between $u_{\mathrm{guess}}$ and the ground truth solution ($u_{\mathrm{label}}$) using the $\ell_2$ norm ($\|u_{\mathrm{guess}} - u_{\text{label}}\|_2$). To assess sensitivity to initialization, we test several representative choices for $u_{\mathrm{guess}}$: setting $u_{\mathrm{guess}}$ at all time steps equal to the initial field ($u_{t=0}$), all zeros field, all ones field, and normal random field. 

\begin{figure}[h]
\begin{center}
    \includegraphics[width=1\linewidth]{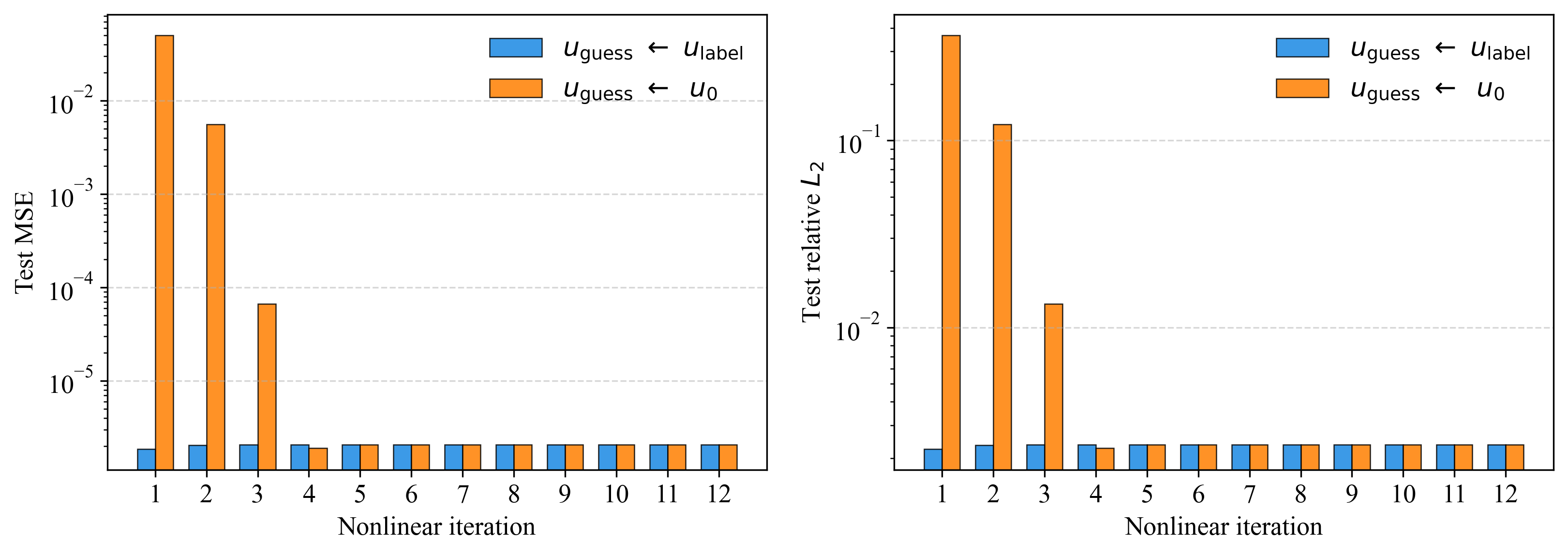}
    \caption{\label{fig:Effect of different initial guesses}Convergence comparison of single-task generalization starting from the true field versus $u_0$ as initial guesses.}
\end{center}
\end{figure}

As shown in Table~\ref{tab:initial_guess_ablation}, for the time-dependent Burgers’ problem, using the $u_{\mathrm{label}}$ as $u_{\mathrm{guess}}$ leads to the lowest error, representing the convergence limit of our method. The $u_{t=0}$ strategy keeps $u_{\mathrm{guess}}$ close to the $u_{\mathrm{label}}$ (lowest non-zero distance) and achieves the best accuracy among the evaluated initialization choices. As shown in the Figure~\ref{fig:Effect of different initial guesses}, the converged error starting from $u_{t=0}$ is comparable to that obtained when using $u_{\mathrm{label}}$ as the initial guess. For the more chaotic K-S system, attempting to solve a single time-block with the initialization of $u_0$ proved very challenging. This is because a simple initial condition can evolve into highly complex structures over time, causing this $u_{\mathrm{guess}}$ to deviate from $u_{\mathrm{label}}$. To address this, we employed a 15-block temporal decomposition strategy. The block-wise inference generally makes the $u_{\mathrm{guess}}$ for each sub-block closer to the $u_{\mathrm{label}}$, thereby improving the convergence of the least-squares solve. As Table~\ref{tab:initial_guess_ablation} shows, the test errors under different initial guesses are comparable. This indicates that the temporal block decomposition mitigates the sensitivity of the least-squares solution to the choice of $u_{\mathrm{guess}}$. We further tested the time-independent lid-driven cavity (LDC) problem and found that zero initialization achieves the lowest test error.

Our experiments show that for the strongly nonlinear/chaotic K-S equations, even a relatively small initialization distance (10–20) can lead to significantly worse convergence, whereas tasks like the Burgers’ equation and LDC problems are much less sensitive. Crucially, our empirical observations unveil a few strategies to mitigate this sensitivity to initialization. Setting $u_{\mathrm{guess}}$ at all time steps equal to the initial field ($u_{t=0}$) can help provide high accuracy for time-dependent problems, while a temporal block decomposition strategy may be employed for increasingly more complex nonlinear time-dependent systems like the K-S problem.

\begin{table}[h]
\caption{Effect of different initial guesses for the nonlinear iterations on test errors.}
\label{tab:initial_guess_ablation}

\begin{center}
\footnotesize 
\renewcommand{\arraystretch}{1.2}
\begin{tabular}{llccccc}

\multirow[b]{2}{*}{Problem} 
& \multirow[b]{2}{*}{Metrics} 
    & \multicolumn{5}{c}{Initial guess for the nonlinear iteration ($u_{\mathrm{guess}}$)} \\
\cmidrule(lr){3-7}
                          &                              
    & Ground truth & All $u_{t=0}$ (ours) & All zeros & All ones & Normal random \\
\toprule
\multirow{3}{*}{Burgers'} 
    & Distance       & 0 & 26.4 & 45.9 & 93.2 & 93.6 \\
    & MSE       & 1.24e-6 & 1.24e-6 & 1.24e-6 & 1.11e-04 & 4.84e-3 \\
    & Relative $L^2$ & 1.04e-3 & 1.04e-3 & 1.04e-3 & 9.61e-03 & 4.96e-2 \\
\midrule

\multirow{3}{*}{K-S}     
    & Distance       & 0 & 3.3 & 13.3 & 23.9 & 24.4 \\
    & MSE       & 2.84e-2 & 2.83e-2 & 2.91e-2 & 2.91e-2 & 3.08e-2 \\
    & Relative $L^2$ & 1.76e-1 & 1.76e-1 & 1.79e-1 & 1.79e-1 & 1.86e-1 \\
\midrule
\multirow{3}{*}{LDC}  
    & Distance       & 0 & -- & 10.5 & 51.7 & 51.3 \\
    & MSE       & 1.58e-4 & --      & 1.67e-4 & 1.76e-3 & 1.86e-4 \\
    & Relative $L^2$ & 5.54e-2 & --      & 5.61e-2 & 1.32e-1 & 5.69e-2 \\
\bottomrule
\end{tabular}
\end{center}
\end{table}

\subsection{Effect of the number of columns in \textbf{A} on inference time} 
\label{app:Effect of the number of columns in A on model inference time}

The scalability of the Tikhonov solution in Newton-PINet is determined by the matrix $\mathbf{A}$. Its rows grow with the total number of PDE, IC, and BC residual samples ($n_{\mathrm{pde}},n_{\mathrm{ic}},n_{\mathrm{bc}}$). Its column size is determined by the product of the output dimension $D_o$ and the pre-final feature dimension (i.e., the total number of neurons in the $L\!-\!1$ hidden layers with each of width $N_n$ due to skip connections). This results in $\mathbf{A}\in \mathbb{R}^{(n_{\mathrm{pde}}+n_{\mathrm{ic}}+n_{\mathrm{bc}}) \times D_o N_n(L-1)}$.

However, the Tikhonov update only involves $\mathbf{A}^T\mathbf{A} \in \mathbb{R}^{D_o N_n(L-1)\times D_o N_n(L-1)}$. We conduct experiments on the 2D steady Navier-Stokes equations for the lid-driven cavity (LDC) problem to evaluate how inference time varies with $D_o$. For this LDC case, $N_n = 128$ and $L = 5$ are fixed, $D_o = 3$ corresponds to predicting $(u, v, p)$, while $D_o = 2$ and $D_o = 1$ correspond to predicting $(u, v)$ with known pressure $p$, and $u$ with known $(p, v)$, respectively. Table~\ref{tab:Inference time versus Do} reports the inference time for different $D_o$. 

We further examine the cost increase due to the column size by varying $N_n$ (Table~\ref{tab:Inference time versus Nn}). The results show that increasing either $D_o$ or $N_n$ enlarges $\mathbf{A}$, thus increasing the cost of the Tikhonov update and task inference time.

\begin{table}[h]
\caption{Inference time versus $D_o$, with $N_n = 128$ and $L = 5$ fixed.}
\label{tab:Inference time versus Do}
\begin{center}
\begin{tabular}{lccc}
    & $D_o=1$  & $2$                     & $3$                     \\  \toprule
Number of \textbf{A} columns  & 512  & 1024 & 1536 \\              
Inference time (s)                                                     & 0.25                    & 0.31                  & 0.40     \\   \bottomrule          
\end{tabular}
\end{center}
\end{table}

\begin{table}[h]
\caption{Inference time versus $N_n$, with $D_o = 3$ and $L = 5$ fixed.}
\label{tab:Inference time versus Nn}
\begin{center}
\begin{tabular}{lcccc}
                    & $N_n=64$ & 128  & 256  & 512  \\  \toprule
Number of \textbf{A} columns & 768   & 1536 & 3072 & 6144 \\
Inference time (s)  & 0.29  & 0.40  & 0.76 & 2.11 \\   \bottomrule     
\end{tabular}
\end{center}
\end{table}

\subsection{Effect of time-blocking strategies on test error} 
\label{app:different time-blocking strategies}

We evaluate the effect of different numbers of time blocks on the average test error taking the convection-diffusion-reaction (CDR) problem as an example. As shown in Table~\ref{tab:Ablation study for different time-blocking strategies}, without temporal decomposition (i.e., using a single time block), the model achieves fast inference but suffers from large errors. This is because the initial guess in the nonlinear iteration may deviate significantly from the true solution, causing poor least-squares accuracy. To address this issue, we apply a temporal domain decomposition strategy: starting from the known initial condition, the final prediction of each block serves as the initial guess for the next. This ensures that each block starts more closely to the true dynamics, enabling reliable long-horizon prediction. As shown in the table, increasing the number of blocks can substantially reduce error, though at the cost of longer inference time. For this problem, a 4-block configuration is adopted to balance the accuracy and efficiency.

\begin{table}[h]
\caption{Ablation study for different time-blocking strategies.}
\label{tab:Ablation study for different time-blocking strategies}
\footnotesize 
\label{tab:time_block_ablation}
\begin{center}
\renewcommand{\arraystretch}{1.2}
\begin{tabular}{lccccccc}
\toprule
\multirow{2}{*}{Metrics} & \multicolumn{7}{c}{Number of blocks} \\
\cmidrule(lr){2-8}
 & 1 & 2 & \textbf{4} & 6 & 10 & 15 & 20 \\
\midrule
Test MSE & 3.56e-3 & 4.46e-3 & \textbf{9.57e-6} & 2.93e-6 & 9.07e-7 & 1.05e-6 & 5.78e-7 \\
Test relative $L^2$ & 2.23e-1 & 3.96e-2 & \textbf{3.56e-3} & 1.80e-3 & 1.13e-3 & 1.12e-3 & 1.08e-3 \\
\begin{tabular}[c]{@{}l@{}}Inference time\\ per task (s)\end{tabular} & 3.74e-2 & 6.82e-2 & \textbf{1.01e-1} & 4.04e-1 & 6.77e-1 & 1.26 & 1.66 \\
\bottomrule
\end{tabular}
\end{center}
\end{table}

\subsection{Effect of initial condition weight on drift at temporal block boundaries} 
\label{app:Effect of initial condition weight parameter on test errors}

\begin{figure}[h]
\begin{center}
    \includegraphics[width=1\linewidth]{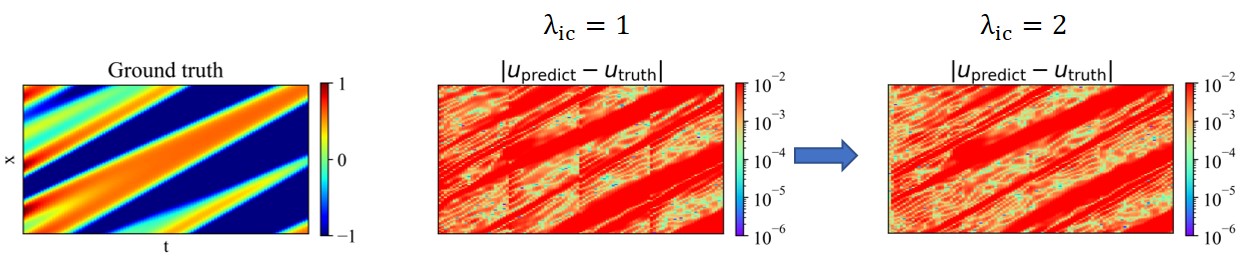}
    \caption{\label{fig:IC weight effect on BC errors} Mitigation of drift at block boundaries via adjustment of the initial condition (IC) weight in the Tikhonov regularization (4-time-block for CDR inference).}
\end{center}
\end{figure}

As can be seen in Figure~\ref{fig:CDR-meta-learning and test results}, the CDR prediction results exhibit noticeable drift at temporal block boundaries. Upon investigation, the drift at boundaries is actually due to an under-weighted enforcement of the initial condition (IC) relative to the PDE and boundary condition (BC) in Tikhonov regularization during the meta-learning and test time. After the ablation study, as shown in Table~\ref{tab:lambda_ic_ablation}, we found that increasing the IC weight ($\lambda_{\mathrm{ic}}$) from 1 to 2 can reduce the test error. Figure~\ref{fig:IC weight effect on BC errors} illustrates that this IC weight adjustment can effectively mitigate the drift at block boundaries.

\begin{table}[h]
\caption{Effect of initial condition (IC) weight in the Tikhonov regularization on test errors.}
\label{tab:lambda_ic_ablation}
\begin{center}
\renewcommand{\arraystretch}{1.2}
\begin{tabular}{lccccc}

\multirow[b]{2}{*}{Errors}

 & \multicolumn{5}{c}{Weight parameter for initial condition ($\lambda_{\mathrm{ic}}$)} \\
\cmidrule(lr){2-6}
 & 1 & 1.5 & 2 & 5 & 10 \\ \toprule
Test MSE         & 9.57e-6 & 6.56e-6 & 5.75e-6 & 5.77e-6 & 6.46e-6 \\
Test relative $L^2$ & 3.56e-3 & 2.89e-3 & 2.67e-3 & 2.67e-3 & 2.80e-3 \\
\bottomrule
\end{tabular}
\end{center}
\end{table}

\subsection{Model performance under other types of boundary conditions} 
\label{app:Effect of boundary condition types}

As shown in Table~\ref{tab:bc_ablation}, we test the model performance for Dirichlet, Neumann, and mixed BCs on the hyperbolic heat and the Klein-Gordon problems. As shown in Table~\ref{tab:discontinuous_bc}, we also evaluate the model for discontinuous BCs on the 2D lid-driven cavity Navier-Stokes problem, where the velocity field exhibits corner discontinuities at the moving lid. These experiments demonstrate that the proposed framework is both theoretically and practically applicable to a broad range of BC types, beyond periodic and Dirichlet settings.

\begin{table}[h]
\caption{Model performance under different boundary conditions.}
\label{tab:bc_ablation}
\begin{center}
\footnotesize 
\renewcommand{\arraystretch}{1.2} 
\setlength{\tabcolsep}{3pt}       
\begin{tabular}{lccccccccc}

\multirow[b]{2}{*}{Problem} 
    & \multicolumn{3}{c}{Dirichlet} 
    & \multicolumn{3}{c}{Neumann} 
    & \multicolumn{3}{c}{Mixed} \\
\cmidrule(lr){2-4} \cmidrule(lr){5-7} \cmidrule(lr){8-10}
    & Time\textsuperscript{1} & MSE       & Rel. $L^2$ 
    & Time\textsuperscript{1} & MSE      & Rel. $L^2$ 
    & Time\textsuperscript{1} & MSE      & Rel. $L^2$ \\
\toprule
Heat
    & 0.067 & 1.69e-6 & 1.76e-3 
    & 0.071 & 1.80e-5 & 5.84e-3 
    & 0.086 & 5.35e-5 & 1.04e-2 \\
Klein-Gordon 
    & 0.064 & 2.36e-7 & 3.86e-4 
    & 0.084 & 2.34e-6 & 1.16e-3 
    & 0.082 & 4.05e-6 & 1.13e-3 \\
\bottomrule
\end{tabular}
\begin{tablenotes}
\footnotesize
\item[1] 1. Inference time per test task (s)
\end{tablenotes}
\end{center}
\end{table}

\begin{table}[h]
\caption{Model performance for 2D lid-driven cavity (LDC) under discontinuous boundary conditions.}
\label{tab:discontinuous_bc}
\centering
\renewcommand{\arraystretch}{1.2}
\begin{tabular}{lccc}

Problem & Inference time per test task (s) & Test MSE & Test Relative $L^2$ \\ 
\toprule
2D LDC & 0.40 & 2.30e-4 & 6.04e-2 \\
\bottomrule
\end{tabular}
\end{table}

\subsection{Ablation studies on model architecture}  
\label{app:Ablation_studies_on_model_architecture}

Newton-PINet employs a skip-connected multi-layer architecture with Tikhonov regularization, Sinusoidal feature embeddings at the input, and $\sin(\cdot)$ activations in all hidden layers. In this section, we perform ablation studies to examine the impact of these design choices and other hyperparameters. All test errors in the following tables are aggregated from 5 individual runs.  

\paragraph{Effect of network depth:}  
We first study the impact of network depth using the Burgers' and CDR problems, with and without skip connections (Table~\ref{tab:skip_ablation_network_depths}). For comparison, Sinusoidal feature embeddings are always included at the input, and all hidden layers adopt $\sin(\cdot)$ activations. The number of neurons per layer is set to 512, 256, 170, and 128 for depths of 1$\sim$4, respectively, so that the last-layer width remains around 512 when skip connections are included. The results show that shallow networks (1 layer) fail to achieve satisfactory accuracy in both problems. Increasing depth without skip connections often deteriorates generalization. In contrast, with skip connections, the model consistently maintains high accuracy across all depths, confirming the stabilizing effect of skip connections and their robustness to depth variations. This property greatly simplifies hyperparameter tuning compared to standard PINNs.  

\begin{table}[h]
\caption{Test relative $L^2$ errors of model with (w/) and without (w/o) skip connections under different network depths.}
\label{tab:skip_ablation_network_depths}
\begin{center}
\renewcommand{\arraystretch}{1.2} 
\begin{tabular}{llcccc}
\multirow[b]{2}{*}{Problem} & \multirow[b]{2}{*}{\shortstack[l]{Skip \\ connections}} & \multicolumn{4}{c}{Network depth} \\ \cmidrule{3-6}
                         &                                   & 1 layer & 2 layers & 3 layers & 4 layers \\ \toprule
\multirow{4}{*}{Burgers'} & \multirow{2}{*}{w/o}          
                         & 1.04e-2 & 3.75e-4 & 4.77e-5 & 5.02e-5 \\
                         & & {\scriptsize $\pm$1.22e-2} & {\scriptsize $\pm$1.50e-3} & {\scriptsize $\pm$1.20e-4} & {\scriptsize $\pm$1.05e-4} \\
                         & \multirow{2}{*}{w/}             
                         & 1.04e-2 & 1.43e-5 & 1.93e-6 & 1.79e-6 \\
                         & & {\scriptsize $\pm$1.22e-2} & {\scriptsize $\pm$7.21e-5} & {\scriptsize $\pm$9.24e-6} & {\scriptsize $\pm$7.98e-6} \\ \midrule
\multirow{4}{*}{CDR}     & \multirow{2}{*}{w/o}          
                         & 4.70e-3 & 7.51e-5 & 6.80e-4 & 2.12e-3 \\
                         & & {\scriptsize $\pm$1.62e-3} & {\scriptsize $\pm$7.46e-5} & {\scriptsize $\pm$7.30e-4} & {\scriptsize $\pm$1.97e-3} \\
                         & \multirow{2}{*}{w/}             
                         & 4.70e-3 & 5.43e-5 & 5.64e-5 & 4.32e-5 \\
                         & & {\scriptsize $\pm$1.62e-3} & {\scriptsize $\pm$4.58e-5} & {\scriptsize $\pm$5.88e-5} & {\scriptsize $\pm$4.18e-5} \\ \bottomrule
\end{tabular}
\end{center}
\end{table}

\paragraph{Effect of network width:}  
Next, we fix the depth to 4 layers and compare the effect of different per-layer widths, with and without skip connections (Table~\ref{tab:width_ablation}). Across both Burgers' and CDR problems, incorporating skip connections consistently improves accuracy under almost all width settings. An exception occurs for the case where all hidden layers have 256 neurons, where skip connections slightly degrade performance—likely due to the substantially increased final-layer width, which makes the Tikhonov regularization computation numerically more challenging. Overall, skip connections improve robustness to width variations.  

\begin{table}[h]
\setlength{\tabcolsep}{4pt} 
\caption{Test relative $L^2$ errors with (w/) and without (w/o) skip connections under different network widths (from top to bottom, corresponding to layer 1$\sim$4).}
\label{tab:width_ablation}
\begin{center}
\renewcommand{\arraystretch}{1.2} 
\begin{tabular}{llccccccc}
\multirow[b]{4}{*}{Problem} & \multirow[b]{4}{*}{\shortstack[l]{Skip \\ connections}} & \multicolumn{7}{c}{Number of nodes in each layer} \\ \cmidrule{3-9}
                         &                                                        & \shortstack{256\\256\\256\\256} & \shortstack{128\\128\\128\\128} & \shortstack{64\\64\\64\\64} & \shortstack{32\\32\\32\\32} & \shortstack{16\\16\\16\\16} & \shortstack{16\\32\\64\\128} & \shortstack{128\\64\\32\\16} \\ \toprule
\multirow{4}{*}{Burgers'} & \multirow{2}{*}{w/o}                                                    & 9.13e-6  & 5.02e-5  & 1.01e-3  & 1.24e-3  & 1.77e-3  & 4.87e-4  & 1.14e-3  \\
                         &                                                        & {\scriptsize $\pm$1.87e-5} & {\scriptsize $\pm$1.05e-4} & {\scriptsize $\pm$2.06e-3} & {\scriptsize $\pm$2.12e-3} & {\scriptsize $\pm$2.38e-3} & {\scriptsize $\pm$8.03e-4} & {\scriptsize $\pm$1.95e-3} \\
                         & \multirow{2}{*}{w/}                                                     & 2.19e-6  & 1.79e-6  & 4.12e-6  & 8.14e-5  & 2.03e-3  & 3.17e-4  & 5.41e-6  \\
                         &                                                        & {\scriptsize $\pm$7.43e-6} & {\scriptsize $\pm$7.98e-6} & {\scriptsize $\pm$1.44e-5} & {\scriptsize $\pm$1.11e-4} & {\scriptsize $\pm$2.78e-3} & {\scriptsize $\pm$5.66e-4} & {\scriptsize $\pm$1.66e-5} \\ \midrule
\multirow{4}{*}{CDR}     & \multirow{2}{*}{w/o}                                                   & 3.88e-4  & 2.12e-3  & 4.80e-3  & 5.98e-2  & 1.05e-1  & 1.34e-3  & 1.24e-1  \\
                         &                                                        & {\scriptsize $\pm$3.44e-4} & {\scriptsize $\pm$1.97e-3} & {\scriptsize $\pm$3.79e-3} & {\scriptsize $\pm$8.14e-2} & {\scriptsize $\pm$6.29e-2} & {\scriptsize $\pm$8.42e-4} & {\scriptsize $\pm$7.24e-2} \\
                         & \multirow{2}{*}{w/}                                                     & 1.10e-3  & 4.32e-5  & 2.25e-4  & 7.73e-4  & 6.48e-3  & 2.02e-4  & 4.55e-4  \\
                         &                                                        & {\scriptsize $\pm$2.20e-3} & {\scriptsize $\pm$4.18e-5} & {\scriptsize $\pm$2.14e-4} & {\scriptsize $\pm$4.77e-4} & {\scriptsize $\pm$8.32e-3} & {\scriptsize $\pm$1.82e-4} & {\scriptsize $\pm$3.47e-4} \\  \bottomrule
\end{tabular}
\end{center}
\end{table}

\paragraph{Effect of mesh resolution:}  
We then examine performance under different mesh resolutions, with consistent resolutions applied during meta-training and testing (Table~\ref{tab:mesh_ablation}). With skip connections, the model exhibits stronger adaptability to resolution changes, particularly in the CDR problem. Notably, different resolutions alter the size of the matrices, which can influence numerical stability. Nevertheless, skip connections help preserve accuracy under these settings.  

\begin{table}[h]
\caption{Test relative $L^2$ errors with (w/) and without (w/o) skip connections under different mesh resolutions}
\label{tab:mesh_ablation}
\begin{center}
\renewcommand{\arraystretch}{1.2} 
\begin{tabular}{llccccc}
\multirow[b]{2}{*}{Problem} & \multirow[b]{2}{*}{\shortstack[l]{Skip \\ connections}} & \multicolumn{5}{c}{Mesh resolution ($t, x$)} \\ \cmidrule{3-7}
                         &                                                        & 11$\times$29 & 13$\times$33 & 17$\times$43 & 26$\times$65 & 51$\times$129 \\  \toprule
\multirow{4}{*}{Burgers'} & \multirow{2}{*}{w/o}                                   & 4.17e-5  & 7.19e-6  & 2.23e-5  & 1.27e-5  & 5.02e-5  \\
                         &                                                        & {\scriptsize $\pm$4.57e-5} & {\scriptsize $\pm$1.52e-5} & {\scriptsize $\pm$2.30e-5} & {\scriptsize $\pm$1.74e-5} & {\scriptsize $\pm$1.05e-4} \\
                         & \multirow{2}{*}{w/}                                    & 2.57e-4  & 5.54e-6  & 2.80e-5  & 1.49e-6  & 1.79e-6  \\
                         &                                                        & {\scriptsize $\pm$7.24e-4} & {\scriptsize $\pm$1.62e-5} & {\scriptsize $\pm$4.61e-6} & {\scriptsize $\pm$7.00e-6} & {\scriptsize $\pm$7.98e-6} \\ \cmidrule{3-7}
\multirow{5}{*}{CDR}     & \multirow{1}{*}{\shortstack[l]{}}                     & 11$\times$29 & 21$\times$49 & 41$\times$69 & 71$\times$99 & 101$\times$129 \\ \midrule 
                         & \multirow{2}{*}{w/o}                                   & 6.34e-2  & 3.08e-3  & 2.26e-3  & 1.51e-3  & 2.12e-3  \\
                         &                                                        & {\scriptsize $\pm$9.09e-2} & {\scriptsize $\pm$2.42e-3} & {\scriptsize $\pm$1.78e-3} & {\scriptsize $\pm$1.16e-3} & {\scriptsize $\pm$1.97e-3} \\
                         & \multirow{2}{*}{w/}                                    & 3.72e-2  & 3.27e-3  & 4.03e-5  & 3.36e-5  & 4.32e-5  \\
                         &                                                        & {\scriptsize $\pm$3.85e-2} & {\scriptsize $\pm$4.55e-3} & {\scriptsize $\pm$1.44e-5} & {\scriptsize $\pm$3.13e-5} & {\scriptsize $\pm$4.18e-5} \\  \bottomrule
\end{tabular}
\end{center}
\end{table}

\paragraph{Effect of activation functions and Sinusoidal embeddings:}  
We further compare activation functions under varying depths, with skip connections included (Table~\ref{tab:activation_ablation}). For $\tanh(\cdot)$ and Gaussian activations, Sinusoidal embeddings are not used, while for $\sin(\cdot)$ activations they are applied at the input. The combination of Sinusoidal embeddings with $\sin(\cdot)$ consistently outperforms other activations across different depths, highlighting the effectiveness of this design choice for nonlinear PDE learning.  

\begin{table}[h]
\caption{Test relative $L^2$ errors of models with different activation functions under varying network depths}
\label{tab:activation_ablation}
\begin{center}
\begin{tabular}{llcccc}

\multirow[b]{2}{*}{Problem} & \multirow[b]{2}{*}{\shortstack[l]{Activation\\ function}}  & \multicolumn{4}{c}{Network depth} \\
\cmidrule{3-6}
        &                     & 1 layer & 2 layers & 3 layers & 4 layers \\
\toprule
\multirow{6}{*}{Burgers'} 
        & \multirow{2}{*}{\shortstack{tanh}}      
                                & 1.07e-2 & 1.30e-4 & 6.39e-4 & 7.13e-5 \\
        &                     & {\scriptsize $\pm$1.30e-2} & {\scriptsize $\pm$2.42e-4} & {\scriptsize $\pm$2.54e-3} & {\scriptsize $\pm$1.60e-4} \\
        & \multirow{2}{*}{\shortstack{Gaussian}} 
                                & 1.14e-2 & 4.64e-3 & 8.27e-6 & 3.28e-6 \\
        &                     & {\scriptsize $\pm$1.31e-2} & {\scriptsize $\pm$8.95e-3} & {\scriptsize $\pm$3.06e-5} & {\scriptsize $\pm$1.26e-5} \\
        & \multirow{2}{*}{\shortstack{sin}}     
                                & 1.04e-2 & 1.43e-5 & 1.93e-6 & 1.79e-6 \\
        &                     & {\scriptsize $\pm$1.22e-2} & {\scriptsize $\pm$7.21e-5} & {\scriptsize $\pm$9.24e-6} & {\scriptsize $\pm$7.98e-6} \\
\midrule
\multirow{6}{*}{CDR} 
        & \multirow{2}{*}{\shortstack{tanh}}      
                                & 3.94e-2 & 1.84e-3 & 6.09e-3 & 4.16e-2 \\
        &                     & {\scriptsize $\pm$1.33e-2} & {\scriptsize $\pm$2.18e-3} & {\scriptsize $\pm$1.16e-2} & {\scriptsize $\pm$8.27e-2} \\
        & \multirow{2}{*}{\shortstack{Gaussian}} 
                                & 3.60e-2 & 5.62e-3 & 6.62e-4 & 2.57e-4 \\
        &                     & {\scriptsize $\pm$2.11e-2} & {\scriptsize $\pm$5.25e-3} & {\scriptsize $\pm$7.16e-4} & {\scriptsize $\pm$2.01e-4} \\
        & \multirow{2}{*}{\shortstack{sin}}     
                                & 4.70e-3 & 5.43e-5 & 5.64e-5 & 4.32e-5 \\
        &                     & {\scriptsize $\pm$1.62e-3} & {\scriptsize $\pm$4.58e-5} & {\scriptsize $\pm$5.88e-5} & {\scriptsize $\pm$4.18e-5} \\
\bottomrule
\end{tabular}
\end{center}
\end{table}

\paragraph{Effect of weight initialization and learning rate:} During meta-learning, we update the hidden-layer weights ($\boldsymbol{\tilde{w}}$) and other learning parameters via gradient descent. Here, we investigate the impact of weight initialization schemes and learning rate ($\eta$) settings on the model generalization (Table~\ref{tab:weight_lr_ablation}). Results indicate that He initialization together with a smaller initial learning rate leads to the most stable and accurate generalization performance, whereas larger learning rates often cause unstable training or accuracy degradation.  

\begin{table}[h]
\caption{Test relative $L^2$ errors under different weight initialization methods and initial learning rates}
\label{tab:weight_lr_ablation}
\begin{center}
\begin{tabular}{llcccc}
\multirow[b]{2}{*}{Problem} & \multirow[b]{2}{*}{\shortstack[l]{Weight\\initialization}} & \multicolumn{3}{c}{Initial learning rate} \\
\cmidrule(lr){3-5}
        &                       & 0.0005 & 0.005 & 0.05 \\
\toprule
\multirow{4}{*}{Burgers'} 
        & \multirow{2}{*}{Xavier} 
                                & 7.57e-5 & 2.83e-6 & 2.09e-3 \\
        &                       & {\scriptsize $\pm$1.62e-4} & {\scriptsize $\pm$9.33e-6} & {\scriptsize $\pm$5.54e-3} \\
        & \multirow{2}{*}{He}  
                                & 1.57e-5 & 1.79e-6 & 2.69e-3 \\
        &                       & {\scriptsize $\pm$7.89e-5} & {\scriptsize $\pm$7.98e-6} & {\scriptsize $\pm$4.93e-3} \\
\midrule
\multirow{4}{*}{CDR}     
        & \multirow{2}{*}{Xavier} 
                                & 2.62e-2 & 5.79e-4 & 1.11e-3 \\
        &                       & {\scriptsize $\pm$2.17e-2} & {\scriptsize $\pm$4.69e-4} & {\scriptsize $\pm$2.20e-3} \\
        & \multirow{2}{*}{He}   
                                & 2.78e-4 & 4.32e-5 & 1.68e-2 \\
        &                       & {\scriptsize $\pm$3.27e-4} & {\scriptsize $\pm$4.18e-5} & {\scriptsize $\pm$3.20e-2} \\
\bottomrule
\end{tabular}
\end{center}
\end{table}

\paragraph{In summary:} Newton-PINet benefits significantly from skip connections, which enhance robustness to depth, width, and mesh resolution, thereby reducing the need for extensive hyperparameter tuning. Sinusoidal feature embeddings with $\sin(\cdot)$ activations further boost accuracy across depths, and careful initialization and learning rate selection improve stability. Collectively, these results explain why Newton-PINet maintains reliable generalization across diverse nonlinear PDE problems.  

\section{Practical scaling discussion of Newton-PINet}
\label{app:Practical scaling discussion of Newton-PINet}

\subsection{Effectiveness of Tikhonov regularization} The scalability of the Tikhonov solution is determined by the matrix $\mathbf{A}$. Its rows grow with the total number of PDE, IC, and BC residual samples. Its column size is determined by the product of the pre-final feature dimension and the output dimension ($D_o$). The pre-final feature dimension equals the total number of neurons in the $L-1$ hidden layers (each of width $N_n$) due to the skip connections. This results in $\mathbf{A}\in \mathbb{R}^{(n_{\mathrm{pde}}+n_{\mathrm{ic}}+n_{\mathrm{bc}}) \times D_o N_n(L-1)}$.

For 2D/3D or multi-physics systems (e.g., Navier-Stokes), the increase in the number of PDE equations requires more residual sampling points, while a higher output dimensionality enlarges the number of columns, making $\mathbf{A}$ larger and more challenging to solve, particularly during Newton iterations for nonlinear problems.

Our strategy is to moderately control the pre-final feature dimension, for example, by decreasing the number of neurons in each layer, thereby reducing the number of columns of $\mathbf{A}$ and ensuring that the least-squares problem remains reasonably overdetermined. Since the Tikhonov update $\boldsymbol{w}=(\lambda_{\text{reg}} \mathbf{I} + \mathbf{A}^{T}\mathbf{A})^{-1}\mathbf{A}^{T}\mathbf{b}$ involves only $\mathbf{A}^T\mathbf{A}\in\mathbb{R}^{ D_o N_n(L-1)\times  D_o N_n(L-1)}$, its computational cost depends primarily on the feature dimension, as opposed to the potentially very large number of residual samples in high-dimensional or multiphysics domains. This keeps the Tikhonov solver tractable as long as the feature dimension is properly controlled. In our experiments, the Newton-PINet achieves high accuracy without requiring a large pre-final feature dimension (i.e., no. features $\ll$ no. collocation points). Appendix~\ref{app:Effect of the number of columns in A on model inference time} investigates the effect of the column size of $\mathbf{A}$ on task inference time. The results show that increasing either the output dimension or the number of network width enlarges $\mathbf{A}$, thereby increasing the cost of the Tikhonov update and the task inference time.

Although the computational cost of the Tikhonov update can be effectively controlled, the cost of constructing $\mathbf{A}$ may become considerable and dominate the overall inference time when the number of residual samples is large (e.g., in high-dimensional systems). Nevertheless, even in such cases, the proposed gradient-free adaptation can be more efficient than gradient-based fine-tuning in PINNs.

\subsection{Effectiveness for solving parametric PDEs}
Our model is well suited for solving parametric PDEs, as the number of residual samples need not be large and tasks can be efficiently mini-batched. The parameters define different tasks and can be large but they are not directly used as network inputs. Instead, our model learns shared representations across tasks. During training, we avoid sampling residual points across parameter inputs as in P$^2$INNs~\citep{cho2024parameterized}, and instead construct $\mathbf{A}$ directly from spatiotemporal points for least-squares adaptation. So, the number of residual samples depends only on the spatiotemporal dimension, not the parameter dimension, which avoids unnecessary growth in the size of $\mathbf{A}$. Table~\ref{tab:Baldwin-PINN-Meta-PINN-compare} demonstrates that the model achieves superior performance on a parametric PDE problem compared to other meta-learning PINN methods.

\subsection{Grid-free nature and resolution adaptability} 
Similar to PINNs and neural operators, our model is grid-free: no mesh construction is required; collocation points can be sampled arbitrarily, and the trained model can give predictions at arbitrary locations, even on finer grids. For example, in the five benchmark cases (Tables~\ref{tab:Experimental-summary} and~\ref{tab:Experimental config summary}), trained on a coarse $32 \times 32$ grid, the model achieves accurate zero-shot generalization on a $128 \times 128$ grid.

\subsection{Time-blocking strategy} 
In addition to the matrix size, the scalability of our model for nonlinear problems also depends on the quality of the initial guess for the iterative updates. Similar to the theoretical behavior of the Newton method, if the initial guess deviates too far from the true solution, more Newton iterations are required to achieve reasonable convergence. This challenge becomes particularly severe for time-dependent chaotic systems, such as the Kuramoto-Sivashinsky (K-S) equation, where even simple initial conditions can quickly evolve into highly complex spatiotemporal patterns. Consequently, it is difficult to obtain an initial guess close to the solution. We can adopt the temporal domain decomposition strategy: starting from the known initial condition, the final prediction of a previous block is used as the initial guess for the next block. This ensures that each block starts more closely to the true dynamics, improving the stability and accuracy of the least-squares updates without using large Newton iterations, and enabling reliable long-horizon prediction.

For complex time-dependent PDEs, temporal domain decomposition is a practical and effective strategy across both AI-based and classical methods. For PINNs methods, training PINNs over the entire spatiotemporal domain in a single-shot usually fails. As noted in prior work~\citep{wang2023expert}, gradient-based training tends to fit later-time dynamics before resolving initial conditions, which can lead to premature convergence (and incorrect solutions), especially for chaotic systems such as K-S. Consequently, many recent PINNs adopt temporal decomposition, typically at the expenses of high computational cost. Separately, many neural PDE solvers (e.g., neural operators)~\citep{hou2026learning,han2022predicting} also rely on autoregressive (one-step) inference for long-time prediction. This strategy is consistent with classical numerical methods, where solutions are advanced sequentially in time to ensure stability. 

\subsection{Low-rank strategy}
Although the task adaptation step is formulated as a least-squares problem, the matrix $\mathbf{A}$ in our method is not sparse in the traditional numerical PDE solver sense. Because the features are generated by a multilayer neural representation with skip connections, each row of $\mathbf{A}$ contains dense network features rather than the typical discretization-induced sparsity and structure. As a consequence, the rank and conditioning of our least-squares system is determined jointly by (i) the density and distribution of collocation points, (ii) the Tikhonov parameter, and more importantly (iii) the quantity and expressiveness of hidden-layer weights that shape the feature space. \textbf{In this instance, the practical scaling for large collocation sets may derive more from the fact that one can still control the cost of the solve through meta-learning a better (yet minimal) set of pre-final layer features.} For this reason, low-rank or iterative solvers designed for large sparse PDE matrices (e.g., LU decomposition or Krylov solvers) may not offer clear advantages in our setting. We experimented with SVD-based solvers, LU, and Cholesky factorizations; none achieved better accuracy-stability trade-offs than the Tikhonov formulation used in our method. A more comprehensive investigation of scalable low-rank or iterative solvers tailored to dense neural-feature matrices is an interesting direction for future work.


\end{document}